# Can SGD Learn Recurrent Neural Networks with Provable Generalization?

Zeyuan Allen-Zhu
zeyuan@csail.mit.edu
Microsoft Research AI

Yuanzhi Li
yuanzhil@stanford.edu
Stanford University

**Abstract**

Recurrent Neural Networks (RNNs) are among the most popular models in sequential data analysis. Yet, in the foundational PAC learning language, what concept class can it learn? Moreover, how can the same recurrent unit simultaneously learn functions from different input tokens to different output tokens, without affecting each other? Existing generalization bounds for RNN scale exponentially with the input length, significantly limiting their practical implications.

In this paper, we show using the vanilla stochastic gradient descent (SGD), RNN can actually learn some notable concept class *efficiently*, meaning that both time and sample complexity scale *polynomially* in the input length (or almost polynomially, depending on the concept). This concept class at least includes functions where each output token is generated from inputs of earlier tokens using a smooth two-layer neural network.

## 1 Introduction

Recurrent neural networks (RNNs) is one of the most popular models in sequential data analysis [25]. When processing an input sequence, RNNs repeatedly and sequentially apply the same operation to each input token. The recurrent structure of RNNs allows it to capture the dependencies among different tokens inside each sequence, which is empirically shown to be effective in many applications such as natural language processing [28], speech recognition [12] and so on.

The recurrent structure in RNNs shows great power in practice, however, it also imposes great challenge in theory. Until now, RNNs remains to be one of the least theoretical understood models in deep learning. Many fundamental open questions are still largely unsolved in RNNs, including

1. (Optimization). When can RNNs be trained efficiently?
2. (Generalization). When do the results learned by RNNs generalize to test data?

Question 1 is technically challenging due to the notorious question of vanishing/exploding gradients, and the non-convexity of the training objective induced by *non-linear* activation functions.

Question 2 requires even deeper understanding of RNNs. For example, in natural language processing, "*Juventus* beats Bacerlona" and "Bacerlona beats *Juventus*" have completely different

*V1 appears on this date and V2 polishes writing. We would like to thank Yingyu Liang and Zhao Song for very helpful conversations. Part of the work was done when Yuanzhi Li was visiting Microsoft Research Redmond.

meanings. How can the same operation in RNN encode a *different rule* for "Juventus" at token 1 vs. "Juventus" at token 3, instead of merely memorizing each training example?

There have been some recent progress towards obtaining more principled understandings of these questions.

On the optimization side, Hardt, Ma, and Recht [13] show that over-parameterization can help in the training process of a linear dynamic system, which is a special case of RNNs with *linear* activation functions. Allen-Zhu, Li, and Song [2] show that over-parameterization also helps in training RNNs with ReLU activations. This latter result gives no generalization guarantee.

On the generalization side, our understanding to RNN is even more limited. The VC-dimension bounds [10] and [17] polynomially depend on the size of the network, and either only apply to linear (or threshold) networks or to networks with one dimension input. Technically speaking, these bounds can be modified to the total number of parameters in the network if only classification error is considered. However, a bound scaling with the total number of parameters usually cannot be applied to modern neural networks, which are heavily over-parameterized. Others [9, 31] (or the earlier work [14]) establish sample complexity bounds that exponentially grow in the input length. In particular, they depend on the operator norm of the recurrent unit, that we denote by $\beta$. If $\beta > 1$, their bounds scale *exponentially* with input length. Since most applications do not regularize $\beta$ and allow $\beta > 1$ for a richer expressibility,[1] their bounds are still insufficient.

Indeed, bridging the gap between optimization (question 1) and generalization (question 2) can be quite challenging in neural networks. The case of RNN is particularly so due to the (potentially) exponential blowup in input length.

- Generalization $\nrightarrow$ Optimization. One could imagine adding a strong regularizer to ensure $\beta \leq 1$ for generalization purpose; however, it is unclear how an optimization algorithm such as stochastic gradient descent (SGD) finds a network that *both* minimizes training loss and maintains $\beta \leq 1$. One could also use a very small network so the number of parameters is limited; however, it is not clear how SGD finds a small network with small training loss.
- Optimization $\nrightarrow$ Generalization. One could try to train RNNs without any regularization; however, it is then quite possible that the number of parameters need to be large and $\beta > 1$ after the training. This is so both in practice (since "memory implies larger spectral radius" [24]) and in theory [2]. All known generalization bounds fail to apply in this regime.

In this paper, we give arguably the first theoretical analysis of RNNs that captures optimization and generalization *simultaneously.* Given any set of input sequences, as long as the outputs are (approximately) realizable by some smooth function in a certain concept class, then after training a vanilla RNN with ReLU activations, SGD provably finds a solution that has both small training and generalization error. Our result allows $\beta$ to be *larger* than 1 by a constant, but is still *efficient*: meaning that the iteration complexity of the SGD, the sample complexity, and the time complexity scale only *polynomially* (or almost polynomially) with the length of the input.

## 2 Notations

We denote by $\|\cdot\|_2$ (or sometimes $\|\cdot\|$) the Euclidean norm of vectors, and by $\|\cdot\|_2$ the spectral norm of matrices. We denote by $\|\cdot\|_\infty$ the infinite norm of vectors, $\|\cdot\|_0$ the sparsity of vectors or diagonal matrices, and $\|\cdot\|_F$ the Frobenius norm of matrices. Given matrix $W$, we denote by $W_k$

[1] For instance, if $W \in \mathbb{R}^{m\times m}$ is the recurrent weight matrix, and is followed with an ReLU activation $\sigma$. Under standard random initialization $\mathcal{N}(0, \frac{2}{m})$, the combined operator $\sigma(Wx) \colon \mathbb{R}^m \to \mathbb{R}^m$ has operator norm $\sqrt{2}$ with high probability. If instead one uses $\mathcal{N}(0, \frac{\sqrt{2}}{m})$, then $\beta$ becomes 1 but gradients will vanish exponentially fast in $L$.

or $w_k$ the $k$-th row vector of $W$. We denote the row $\ell_p$ norm for $W \in \mathbb{R}^{m\times d}$ as

$$\|W\|_{2,p} \stackrel{\text{def}}{=} \left(\sum_{i\in[m]} \|w_i\|_2^p\right)^{1/p} . \tag{2.1}$$

By definition, $\|W\|_{2,2} = \|W\|_F$. We use $\mathcal{N}(\mu, \sigma)$ to denote Gaussian distribution with mean $\mu$ and variance $\sigma$; or $\mathcal{N}(\mu, \Sigma)$ to denote Gaussian vector with mean $\mu$ and covariance $\Sigma$. We use $x = y \pm z$ to denote that $x \in [y - z, y + z]$. We use $\mathbb{1}_{event}$ to denote the indicator function of whether *event* is true. We denote by $\mathbf{e}_k$ the $k$-th standard basis vector. We use $\sigma(\cdot)$ to denote the ReLU function $\sigma(x) = \max\{x, 0\} = \mathbb{1}_{x\geq 0} \cdot x$. Given univariate function $f : \mathbb{R} \to \mathbb{R}$, we also use $f$ to denote the same function over vectors: $f(x) = (f(x_1), \dots, f(x_m))$ if $x \in \mathbb{R}^m$.

Given vectors $v_1, \dots, v_n \in \mathbb{R}^m$, we define $U = \mathsf{GS}(v_1, \dots, v_n)$ as their Gram-Schmidt orthonormalization. Namely, $U = [\widehat{v}_1, \dots, \widehat{v}_n] \in \mathbb{R}^{m\times n}$ where

$$\widehat{v}_1 = \frac{v_1}{\|v_1\|} \quad \text{and} \quad \text{for } i \geq 2: \quad \widehat{v}_i = \frac{\prod_{j=1}^{i-1}(I - \widehat{v}_j\widehat{v}_j^\top)v_i}{\|\prod_{j=1}^{i-1}(I - \widehat{v}_j\widehat{v}_j^\top)v_i\|} .$$

Note that in the occasion that $\prod_{j=1}^{i-1}(I - \widehat{v}_j\widehat{v}_j^\top)v_i$ is the zero vector, we let $\widehat{v}_i$ be an arbitrary unit vector that is orthogonal to $\widehat{v}_1, \dots, \widehat{v}_{i-1}$.

We say a function $f : \mathbb{R}^d \to \mathbb{R}$ is $L$-Lipscthiz continuous if $|f(x) - f(y)| \leq L\|x - y\|_2$; and say it is is $L$-smooth if its gradient is $L$-Lipscthiz continuous, that is $\|\nabla f(x) - \nabla f(y)\|_2 \leq L\|x - y\|_2$.

**Function complexity.** The following notions from [1] measure the complexity of any infinite-order smooth function $\phi : \mathbb{R} \to \mathbb{R}$. Suppose $\phi(z) = \sum_{i=0}^{\infty} c_i z^i$ is its Taylor expansion. Given non-negative $R$,

$$\mathfrak{C}_\varepsilon(\phi, R) \stackrel{\text{def}}{=} \sum_{i=0}^{\infty} \left((C^* R)^i + \left(\frac{\sqrt{\log(1/\varepsilon)}}{\sqrt{i}} C^* R\right)^i\right)|c_i|$$

$$\mathfrak{C}_\mathfrak{s}(\phi, R) \stackrel{\text{def}}{=} C^* \sum_{i=0}^{\infty} (i + 1)^{1.75} R^i |c_i|$$

where $C^*$ is a sufficiently large constant (e.g., $10^4$). It holds $\mathfrak{C}_\mathfrak{s}(\phi, R) \leq \mathfrak{C}_\varepsilon(\phi, R) \leq \mathfrak{C}_\mathfrak{s}(\phi, O(R)) \times \mathsf{poly}(1/\varepsilon)$, and for $\sin z, e^z$ or low degree polynomials, they only differ by $o(1/\varepsilon)$. [1]

**Example 2.1.** If $\phi(z) = z^d$ for constant $d$ then $\mathfrak{C}_\mathfrak{s}(\phi, R) \leq O(R^d)$, $\mathfrak{C}_\varepsilon(\phi, R) \leq O(R^d \mathsf{polylog}(\frac{1}{\varepsilon}))$. For functions such as $\phi(z) = e^z - 1, \sin z$, sigmoid$(z)$ or $\tanh(z)$, it suffices to consider $\varepsilon$-approximations of them so we can truncate their Taylor expansions to degree $O(\log(1/\varepsilon))$. This gives $\mathfrak{C}_\mathfrak{s}(\phi, R), \mathfrak{C}_\varepsilon(\phi, R) \leq (1/\varepsilon)^{O(\log R)}$.

# 3 Problem Formulation

The data are generated from an unknown distribution $\mathcal{D}$ over $(x^\star, y^\star) \in (\mathbb{R}^{d_x})^{(L-2)} \times \mathcal{Y}^{(L-2)}$. Each input sequence $x^\star$ consists of $x_2^\star, \dots, x_{L-1}^\star \in \mathbb{R}^{d_x}$ with $\|x_\ell^\star\| = 1$ and $[x_\ell^\star]_{d_x} = \frac{1}{2}$ without loss of generality.[2] Each label sequence $y^\star$ consists of $y_3^\star, \dots, y_L^\star \in \mathcal{Y}$. The training dataset $\mathcal{Z} = \{((x^\star)^{(i)}, (y^\star)^{(i)})\}_{i\in[N]}$ is given as $N$ i.i.d. samples from $\mathcal{D}$. When $(x^\star, y^\star)$ is generated from $\mathcal{D}$, we call $x^\star$ the *true* input sequence and $y^\star$ the true label.

**Definition 3.1.** *Without loss of generality (see Remark 3.4), for each true input $x^\star$, we transform it into an* actual input *sequence $x_1, x_2, \dots, x_L \in \mathbb{R}^{d_x+1}$ as follows.*

$$x_1 = (0^{d_x}, 1) \quad \textit{and} \quad x_\ell = (\varepsilon_x x_\ell^\star, 0) \quad \textit{for} \quad \ell = 2, 3, \dots, L - 1 \quad \textit{and} \quad x_L = (0^{d_x}, \varepsilon_x)$$

[2] This is without loss of generality, since $\frac{1}{2}$ can always be padded to the last coordinate, and $\|x_\ell^\star\|_2 = 1$ can always be ensured from $\|x_\ell^\star\|_2 \leq 1$ by padding $\sqrt{1 - \|x_\ell^\star\|_2^2}$ to the second-last coordinate. We make this assumption to simplify our notations: for instance, $(x_\ell^\star)_{d_x} = \frac{1}{2}$ allows us to focus only on networks in the concept class without bias.

*where $\varepsilon_x \in (0,1)$ is a parameter to be chosen later. We then feed this actual sequence $x$ into RNN.*

In this way we have ensured that the actual input sequence is *normalized*:

**Definition 3.2.** *We say the sequence $x_1, \dots, x_L \in \mathbb{R}^{d_x+1}$ is normalized if*

$$\|x_1\| = 1 \quad \text{and} \quad \|x_\ell\| = \varepsilon_x \quad \text{for all } \ell = 2, 3, \dots, L.$$

## 3.1 Our Learner Network: Elman RNN

To present the simplest result, we focus on the classical Elman RNN with ReLU activation. Let $W \in \mathbb{R}^{m\times m}$, $A \in \mathbb{R}^{m\times(d_x+1)}$, and $B \in \mathbb{R}^{d\times m}$ be the weight matrices.

**Definition 3.3.** *Our Elman RNN can be described as follows. On input $x_1, \dots, x_L \in \mathbb{R}^{d_x+1}$,*

$$h_0 = 0 \in \mathbb{R}^m \qquad g_\ell = W \cdot h_{\ell-1} + Ax_\ell \in \mathbb{R}^m$$
$$y_\ell = B \cdot h_\ell \in \mathbb{R}^d \qquad h_\ell = \sigma(W \cdot h_{\ell-1} + Ax_\ell) \in \mathbb{R}^m$$

*We say that $W, A, B$ are at random initialization, if the entries of $W$ and $A$ are i.i.d. generated from $\mathcal{N}(0, \frac{2}{m})$, and the entries of $B$ are i.i.d. generated from $\mathcal{N}(0, \frac{1}{d})$.*

For simplicity, in this paper we only update $W$ and let $A$ and $B$ be at their random initialization. Thus, we write $F_\ell(x^\star; W) = y_\ell = Bh_\ell$ as the output of the $\ell$-th layer.

Our goal is to use $y_3, \dots, y_L \in \mathbb{R}^d$ to fit the true label $y_3^\star, \dots, y_L^\star \in \mathcal{Y}$ using some *loss* function $G \colon \mathbb{R}^d \times \mathcal{Y} \to \mathbb{R}$. In this paper we assume, for every $y^\star \in \mathcal{Y}$, $G(0^d, y^\star) \in [-1, 1]$ is bounded, and $G(\cdot, y^\star)$ is convex and 1-Lipschitz continuous in its first variable. This includes for instance the cross-entropy loss and $\ell_2$-regression loss (for $y^\star$ being bounded).

*Remark* 3.4. Since we only update $W$, the label sequence $y_3^\star, \dots, y_L^\star$ is off from the input sequence $x_2^\star, \dots, x_{L-1}^\star$ by one. The last $x_L$ can be made zero, but we keep it normalized for notational simplicity. The first $x_1$ gives a random seed fed into the RNN (one can equivalently put it into $h_0$). We have scaled down the input signals by $\varepsilon_x$, which can be equivalently thought as scaling down $A$.

## 3.2 Concept Class

Let $\{\Phi_{i\to j,r,s} \colon \mathbb{R} \to \mathbb{R}\}_{i,j\in[L],r\in[p],s\in[d]}$ be infinite-order differentiable functions, and $\{w^*_{i\to j,r,s} \in \mathbb{R}^{d_x}\}_{i,j\in[L],r\in[p],s\in[d]}$ be unit vectors. Then, for every $j = 3, 4, \dots, L$, we consider *target functions* $F_j^* \colon \mathbb{R}^{d_x} \to \mathbb{R}^d$ where $F_j^* = \big(F_{j,1}^*, \dots, F_{j,d}^*\big)$ can be written as

$$F_{j,s}^*(x^\star) \stackrel{\text{def}}{=} \sum_{i=2}^{j-1} \sum_{r\in[p]} \Phi_{i\to j,r,s}(\langle w^*_{i\to j,r,s}, x_i^\star\rangle) \in \mathbb{R} \ . \tag{3.1}$$

For proof simplicity, we assume $\Phi_{i\to j,r,s}(0) = 0$. We also use

$$\mathfrak{C}_\varepsilon(\Phi, R) = \max_{i,j,r,s}\{\mathfrak{C}_\varepsilon(\Phi_{i\to j,r,s}, R)\} \quad \text{and} \quad \mathfrak{C}_\mathfrak{s}(\Phi, R) = \max_{i,j,r,s}\{\mathfrak{C}_\mathfrak{s}(\Phi_{i\to j,r,s}, R)\}$$

to denote the complexity of $F^*$.

**Agnostic PAC-learning language.** Our concept class consists of all functions $F^*$ in the form of (3.1) with complexity bounded by threshold $C$ and parameter $p$ bounded by threshold $p_0$. Let $\mathsf{OPT}$ be the population risk achieved by the *best* target function in this concept class. Then, our goal is to learn this concept class with population risk $\mathsf{OPT} + \varepsilon$ using sample and time complexity

**Algorithm 1** SGD for RNNs

**Input:** Data set $\mathcal{Z}$, step size $\eta$.
1: $W_0 = 0$.
2: **for** $t = 1, 2, \dots T$ **do**
3: Randomly sample $(x^\star, y^\star)$ from the data set $\mathcal{Z}$
4: Update: $W_t = W_{t-1} - \eta \nabla_{W'} \mathsf{Obj}(x^\star, y^\star; W_{t-1})$.
5: **end for**

*polynomial* in $C$, $p_0$ and $1/\varepsilon$. In the remainder of this paper, to simplify notations, we do not explicitly define this concept class parameterized by $C$ and $p$. Instead, we equivalently state our theorem with respect to any (unknown) target function $F^*$ with specific parameters $C$ and $p$.

**Example 3.5.** Our concept class is general enough and contains functions where the output at each token is generated from inputs of previous tokens using any two-layer neural network. Indeed, one can verify that our general form (3.1) includes functions of the following:

$$F_j^*(x^\star) = \sum_{i=2}^{j-1} A_{j-i}^* \phi(W_{j-i}^* x_i^\star) \ .$$

# 4 Our Result: RNN Provably Learns the Concept Class

Suppose the distribution $\mathcal{D}$ is generated by some (unknown) target function $F^*$ of the form (3.1) in the concept class with population risk $\mathsf{OPT}$, namely,

$$\mathbb{E}_{(x^\star, y^\star)\sim\mathcal{D}} \left[\sum_{j=3}^{L} G\left(F_j^*(x^\star), y_j^\star\right)\right] \leq \mathsf{OPT} \ ,$$

and suppose we are given training dataset $\mathcal{Z}$ consisting of $N$ i.i.d. samples from $\mathcal{D}$. We consider the following stochastic training objective

$$\mathsf{Obj}(W') \stackrel{\text{def}}{=} \mathop{\mathbb{E}}_{(x^\star, y^\star)\sim\mathcal{Z}} \left[\mathsf{Obj}(x^\star, y^\star; W')\right]$$

$$\text{where} \quad \mathsf{Obj}(x^\star, y^\star; W') \stackrel{\text{def}}{=} \sum_{j=3}^{L} G\left(\lambda F_j(x^\star; W + W'), y_j^\star\right)$$

Above, $W \in \mathbb{R}^{m\times m}$ is random initialization, $W' \in \mathbb{R}^{m\times m}$ is the additional shift, and $\lambda \in (0,1)$ is a constant scaling factor on the network output.[3] We consider the vanilla stochastic gradient descent (SGD) algorithm with step size $\eta$, see Algorithm 1.[4]

**Theorem 1.** *For every* $0 < \varepsilon < \widetilde{O}\big(\frac{1}{\mathsf{poly}(L,d)\cdot p \cdot \mathfrak{C}_{\mathfrak{s}}(\Phi, O(\sqrt{L}))}\big)$*, define complexity* $C = \mathfrak{C}_\varepsilon(\Phi, \sqrt{L})$ *and* $\lambda = \widetilde{\Theta}\big(\frac{\varepsilon}{L^2 d}\big)$*, if the number of neurons* $m \geq \mathsf{poly}(C, \varepsilon^{-1})$ *and the number of samples is* $N = |\mathcal{Z}| \geq \mathsf{poly}(C, \varepsilon^{-1}, \log m)$*, then SGD with* $\eta = \widetilde{\Theta}\big(\frac{1}{\varepsilon L^2 d^2 m}\big)$ *and*

$$T = \widetilde{\Theta}\Big(\frac{p^2 C^2 \mathsf{poly}(L,d)}{\varepsilon^2}\Big)$$

[3] Equivalently, one can scale matrix $B$ by factor $\lambda$. For notational simplicity, we split the matrix into $W + W'$ but this does not change the algorithm since gradient with respect to $W + W'$ is the same with respect to $W'$.

[4] Strictly speaking, $\mathsf{Obj}(x^\star, y^\star; W')$ does not have gradient everywhere due to the non-differentiability of ReLU. Throughout the paper, $\nabla$ is used to denote the value computed by setting $\nabla\sigma(x) = \mathbb{1}[x \geq 0]$, which is also what is used in practical auto-differentiation softwares.

*satisfies that, with probability at least* $1 - e^{-\Omega(\rho^2)}$ *over the random initialization*

$$\mathbb{E}_{sgd}\left[\frac{1}{T}\sum_{t=0}^{T-1}\mathbb{E}_{(x^\star,y^\star)\sim\mathcal{D}}\left[\mathsf{Obj}(x^\star,y^\star;W+W_t)\right]\right] \le \mathsf{OPT}+\varepsilon \ .$$

Above, $\mathbb{E}_{sgd}$ takes expectation with respect to the randomness of SGD. Since SGD takes only one example per iteration, the sample complexity $N$ is also bounded by $T$.

## 4.1 Our Contribution, Interpretation, and Discussion

**Sample complexity.** Our sample complexity only scales with $\log(m)$, making the result applicable to *over-parameterized* RNNs that have $m \gg N$. Following Example 2.1, if $\phi(z)$ is constant degree polynomial we have $C = \mathsf{poly}(L, \log\varepsilon^{-1})$ so Theorem 1 says that RNN learns such concept class

$$\text{with size } m = \frac{\mathsf{poly}(L,d,p)}{\mathsf{poly}(\varepsilon)} \text{ and sample complexity } \min\{N,T\} = \frac{p^2\mathsf{poly}(L,d,\log m)}{\varepsilon^2}$$

If $\phi(z)$ is a function with good Taylor truncation, such as $e^z-1, \sin z$, $\mathrm{sigmoid}(z)$ or $\tanh(z)$, then $C = L^{O(\log(1/\varepsilon))}$ is almost polynomial.

**Non-linear measurements.** Our result shows that vanilla RNNs can efficiently learn a weighted average of non-linear measurements of the input. As we argued in Example 3.5, this at least includes functions where the output at each token is generated from inputs of previous tokens using any two-layer neural networks. Average of non-linear measurements can be quite powerful, achieving the state-of-the-art performance in some sequential applications such as sentence embedding [4] and many others [23], and acts as the base of attention mechanism in RNNs [5].

**Adapt to tokens.** In the target function, $\Phi_{i\to j,r,s}$ can be different at each token, meaning that they can *adapt* to the position of the input tokens. We emphasize that the positions of the tokens (namely, the values $i, j$) are not *directly fed* into the network, rather it is discovered through sequentially reading the input. As one can see from our proofs, the ability of adapting to the tokens comes from the inhomogeneity in hidden layers $h_\ell$: even when $x_\ell = x_{\ell'}$ for different tokens $\ell' \neq \ell$, there is still big difference between $h_\ell$ and $h_{\ell'}$. Albeit the *same* operator is applied to $x_\ell$ and $x_{\ell'}$, RNNs can still use this crucial inhomogeneity to learn different functions at different tokens.

In our result, the function $\Phi_{i\to j,r,s}$ only adapts with the positions of the input tokens, but in many applications, we would like the function to adapt with the values of the past tokens $x_1^\star, \dots, x_{i-1}^\star$ as well. We believe a study on other models (such as LSTM [15]) can potentially settle these questions.

**Long term memory.** It is commonly believed that vanilla RNNs cannot capture *long term* dependencies in the input. This does not contradict our result. Our complexity parameter $\mathfrak{C}_\varepsilon(\Phi, \sqrt{L})$ of the learning process in Theorem 1 indeed *suffers* from $L$, the length of the input sequence. This is due to the fact that vanilla RNN, the hidden neurons $h_\ell$ will incorporate more and more noise as the time horizon $\ell$ increases, making the new signal $Ax_\ell$ less and less significant.

**Comparison to feed-forward networks.** Recently there are many interesting results on analyzing the learning process of feed-forward neural networks [7, 8, 11, 16, 18–20, 26, 27, 29, 30, 32]. Most of them either assume that the input is structured (e.g. Gaussian or separable) or only consider linear networks. Allen-Zhu, Li, and Liang [1] show a result in the same flavor as this paper but for two and three-layer networks. Since RNNs apply the same unit repeatedly to each input token in a sequence, our analysis is significantly different from [1] and creates lots of difficulties in the analysis.

### 4.2 Conclusion

We show RNN can actually learn some notable concept class *efficiently*, using simple SGD method with sample complexity polynomial or almost-polynomial in input length. This concept class at least includes functions where each output token is generated from inputs of earlier tokens using a smooth neural network. To the best of our knowledge, this is the first proof that some non-trivial concept class is *efficiently* learnable by RNN. Our sample complexity is almost independent of $m$, making the result applicable to over-parameterized settings. On a separate note, our proof explains why the same recurrent unit is capable of learning various functions from different input tokens to different output tokens.

# Proof Sketch

Our proof of Theorem 1 divides into four conceptual steps.

1. We obtain first-order approximation of how much the outputs of the RNN change if we move from $W$ to $W+W'$. This change (up to small error) is a linear function in $W'$. (See Section 6).

   *(This step can be derived from prior work [2] without much difficulty.)*

2. We construct some (unknown) matrix $W^* \in \mathbb{R}^{m\times m}$ so that this "linear function", when evaluated on $W^*$, approximately gives the target $F^*$ in the concept class (see Section 5).

   *(This step is the most interesting part of this paper.)*

3. We argue that the SGD method moves in a direction nearly as good as $W^*$ and thus efficiently decreases the training objective (see Section 7).

   *(This is a routine analysis of SGD in the non-convex setting given Steps 1&2.)*

4. We use the first-order linear approximation to derive a Rademacher complexity bound that does not grow exponentially in $L$ (see Section 8). By feeding the output of SGD into this Rademacher complexity, we finish the proof of Theorem 1 (see Section 9).

   *(This is a one-paged proof given the Steps 1&2&3.)*

Although our proofs are technical, to help the readers, we write 7 pages of sketch proofs for Steps 1 through 4. This can be found in Section 5 through 9. Our final proofs reply on many other technical properties of RNN that may be of independent interests: such as properties of RNN at random initialization (which we include in Section B and C), and properties of RNN stability (which we include in Section D, E, F). Some of these properties are simple modifications from prior work, but some are *completely new* and require new proof techniques (namely, Section C, D and E). We introduce some notations for analysis purpose.

**Definition 4.1.** *For each $\ell \in [L]$, let $D_\ell \in \mathbb{R}^{m\times m}$ be the diagonal matrix where*

$$(D_\ell)_{k,k} = \mathbb{1}_{(W\cdot h_{\ell-1}+Ax_\ell)_k\geq 0} = \mathbb{1}_{(g_\ell)_k\geq 0} \ .$$

*As a result, we can write $h_\ell = D_\ell W h_{\ell-1}$. For each $1\leq \ell \leq a \leq L$, we define*

$$\mathsf{Back}_{\ell\to a} = B D_a W \cdots D_{\ell+1} W \in \mathbb{R}^{d\times m}.$$

*with the understanding that $\mathsf{Back}_{\ell\to\ell} = B \in \mathbb{R}^{d\times m}$.*

Throughout the proofs, to simplify notations when specifying polynomial factors, we introduce

$$\rho \stackrel{\text{def}}{=} 100Ld\log m \quad \text{and} \quad \varrho \stackrel{\text{def}}{=} \frac{100Ldp\cdot \mathfrak{C}_\varepsilon(\Phi,\sqrt{L})\cdot \log m}{\varepsilon}$$

We assume $m \geq \mathsf{poly}(\varrho)$ for some sufficiently large polynomial factor.

# 5 Existence of Good Network Through Backward

One of our main contributions is to show the existence of some "good linear network" to approximate any target function. Let us explain what this means. Suppose $W, A, B$ are at random initialization. We consider a linear function over $W^{*} \in \mathbb{R}^{m\times m}$:

$$f_{j'} \stackrel{\text{def}}{=} \textstyle\sum_{i'=2}^{j'} \mathsf{Back}_{i'\to j'} D_{i'} W^{*} h_{i'-1} \ . \tag{5.1}$$

As we shall see later, in first-order approximation, this linear function captures how much the output of the RNN changes at token $j'$, if one we move $W$ to $W+W'$. The goal in this section is to construct some $W^{*} \in \mathbb{R}^{m\times m}$ satisfying that, for *any* true input $x^\star$ in the support of $\mathcal{D}$, if we define the actual input $x$ according to $x^\star$ (see Definition 3.1), then with high probability,

$$\forall s' \in [d] \qquad f_{j',s'} \approx F^{*}_{j',s'}(x^\star) = \textstyle\sum_{i=2}^{j'-1}\sum_{r\in[p]} \Phi_{i\to j',r,s'}(\langle w^{*}_{i\to j',r,s'}, x^\star_i\rangle) \tag{5.2}$$

Furthermore, $W^{*}$ is appropriately bounded in Frobenius norm. In our sketched proof below, it shall become clear how this *same* matrix $W^{*}$ can simultaneously represent functions $\Phi_{i\to j'}$ that come from different input tokens $i$. Since SGD can be shown to descend in a direction "comparable" to $W^{*}$, it converges to a matrix $W$ with similar guarantees.

## 5.1 Indicator to Function

In order to show (5.2), we first show a variant of the "indicator to function" lemma from [1].

**Lemma 5.1** (indicator to function). *For every smooth function $\Phi\colon \mathbb{R}\to\mathbb{R}$, every unit vector $w^* \in \mathbb{R}^{d_x}$ with $w^*_{d_x} = 0$, every constant $\sigma \geq 0.1$, every constant $\gamma > 1$, every constant $\varepsilon_e \in \big(0, \frac{1}{\mathfrak{C}_{\mathfrak{s}}(\Phi, O(\sigma))}\big)$, there exists*

$$C' = \mathfrak{C}_{\varepsilon_e}(\Phi,\sigma) \text{ and a function } H\colon \mathbb{R}\to[-C',C'],$$

*such that for every fixed unit vectors $x^\star \in \mathbb{R}^{d_x}$ with $x^\star_{d_x} = \frac{1}{2}$,*

*(a)* $\left|\mathbb{E}_{a\sim\mathcal{N}(0,\mathbf{I}), n\sim\mathcal{N}(0,\sigma^2)}\left[\mathbb{1}_{\langle a,x^\star\rangle+n\geq 0} H(a)\right] - \Phi(\langle w^*, x^\star\rangle)\right| \leq \varepsilon_e$ *(on target)*

*(b)* $\left|\mathbb{E}_{a\sim\mathcal{N}(0,\mathbf{I}), n\sim\mathcal{N}(0,\sigma^2)}\left[\mathbb{1}_{\langle a,x^\star\rangle+\gamma n\geq 0} H(a)\right] - \Phi(0)\right| \leq \varepsilon_e + O\big(\frac{C'\log(\gamma\sigma)}{\gamma\sigma}\big)$ *(off target)*

Above, Lemma 5.1a says that we can use a bounded function $\mathbb{1}_{\langle a,x^\star\rangle+n\geq 0}H(a)$ to fit a target function $\Phi(\langle w^*, x^\star\rangle)$, and Lemma 5.1b says that if the magnitude of $n$ is large then this function is close to being constant. For such reason, we can view $n$ as "*noise*." While the proof of 5.1a is from prior work [1], our new property 5.1b is completely new and it requires some technical challenge to simultaneously guarantee 5.1a and 5.1b. The proof is in Appendix G.1

## 5.2 Fitting a Single Function

We now try to apply Lemma 5.1 to approximate a single function $\Phi_{i\to j,r,s}(\langle w^*_{i\to j,r,s}, x^\star_i\rangle)$. For this purpose, let us consider two (normalized) input sequences. The first (null) sequence $x^{(0)}$ is given as

$$x^{(0)}_1 = (0^{d_x}, 1) \quad \text{and} \quad x^{(0)}_\ell = (0^{d_x}, \varepsilon_x) \text{ for } \ell = 2,3,\dots,L$$

The second sequence $x$ is generated from an input $x^\star$ in the support of $\mathcal{D}$ (recall Definition 3.1). Let

- $h_\ell, D_\ell, \mathsf{Back}_{i\to j}$ be defined with respect to $W, A, B$ and input sequence $x$, and
- $h_\ell^{(0)}, D_\ell^{(0)}, \mathsf{Back}_{i\to j}^{(0)}$ be defined with respect to $W, A, B$ and input sequence $x^{(0)}$

We remark that $h_\ell^{(0)}$ has the good property that it does not depend $x^\star$ but somehow stays "close enough" to the true $h_\ell$ (see Appendix D for a full description).

**Lemma 5.2** (fit single function). *For every $2 \le i < j \le L$, $r \in [p], s \in [d]$ and every constant $\varepsilon_e \in \big(0, \frac{1}{\mathfrak{C}_\mathfrak{s}(\Phi_{i\to j,r,s}, O(\sqrt{L}))}\big)$, there exists $C' = \mathfrak{C}_{\varepsilon_e}(\Phi_{i\to j,r,s}, \sqrt{L})$ so that, for every*

$$\varepsilon_x \in \big(0, \tfrac{1}{\rho^4 C'}\big) \quad \text{and} \quad \varepsilon_c = \tfrac{\varepsilon_e \varepsilon_x}{4C'} \ ,$$

*there exists a function $H_{i\to j,r,s} \colon \mathbb{R} \to \big[-\frac{4(C')^2}{\varepsilon_e\varepsilon_x}, \frac{4(C')^2}{\varepsilon_e\varepsilon_x}\big]$, such that, let*

- *$x$ be a fixed input sequence defined by some $x^\star$ in the support of $\mathcal{D}$ (see Definition 3.1),*
- *$W, A$ be at random initialization,*
- *$h_\ell$ be generated by $W$,$A$,$x$ and $h_\ell^{(0)}$ be generated by $W$,$A$,$x^{(0)}$, and*
- *$\widetilde{w}_k, \widetilde{a}_k \sim \mathcal{N}\big(0, \frac{2\mathbf{I}}{m}\big)$ be freshly new random vectors,*

*with probability at least $1 - e^{-\Omega(\rho^2)}$ over $W$ and $A$,*

*(a) (on target)*

$$\left| \mathop{\mathbb{E}}_{\widetilde{w}_k, \widetilde{a}_k} \left[ \mathbb{1}_{|\langle \widetilde{w}_k, h_{i-1}^{(0)}\rangle| \le \frac{\varepsilon_c}{\sqrt{m}}} \mathbb{1}_{\langle \widetilde{w}_k, h_{i-1}\rangle + \langle \widetilde{a}_k, x_i\rangle \ge 0} H_{i\to j,r,s}(\widetilde{a}_k) \right] - \Phi_{i\to j,r,s}(\langle w^*_{i\to j,r,s}, x_i^\star\rangle) \right| \le \varepsilon_e$$

*(b) (off target), for every $i' \ne i$*

$$\left| \mathop{\mathbb{E}}_{\widetilde{w}_k, \widetilde{a}_k} \left[ \mathbb{1}_{|\langle \widetilde{w}_k, h_{i-1}^{(0)}\rangle| \le \frac{\varepsilon_c}{\sqrt{m}}} \mathbb{1}_{\langle \widetilde{w}_k, h_{i'-1}\rangle + \langle \widetilde{a}_k, x_{i'}\rangle \ge 0} H_{i\to j,r,s}(\widetilde{a}_k) \right] \right| \le \varepsilon_e$$

Lemma 5.2 implies there is a quantity $\mathbb{1}_{|\langle \widetilde{w}_k, h_{i-1}^{(0)}\rangle| \le \frac{\varepsilon_c}{\sqrt{m}}} H_{i\to j,r,s}(\widetilde{a}_k)$ that *only depends* on the target function and the random initialization (namely, $\widetilde{w}_k, \widetilde{a}_k$) such that,

- when multiplying $\mathbb{1}_{\langle \widetilde{w}_k, h_{i-1}\rangle + \langle \widetilde{a}_k, x_i\rangle \ge 0}$ gives the target $\Phi_{i\to j,r,s}(\langle w^*_{i\to j,r,s}, x_i^\star\rangle$, but
- when multiplying $\mathbb{1}_{\langle \widetilde{w}_k, h_{i'-1}\rangle + \langle \widetilde{a}_k, x_{i'}\rangle \ge 0}$ gives near zero.

The full proof is in Appendix G.2 but we sketch why Lemma 5.2 can be derived from Lemma 5.1.

*Sketch proof of Lemma 5.2.* Let us focus on indicator $\mathbb{1}_{\langle \widetilde{w}_k, h_{i'-1}\rangle + \langle \widetilde{a}_k, x_{i'}\rangle \ge 0}$:

- $\langle \widetilde{a}_k, x_{i'}\rangle$ is distributed like $\mathcal{N}(0, \frac{2\varepsilon_x^2}{m})$ because $\langle \widetilde{a}_k, x_{i'}\rangle = \big\langle(\widetilde{a}_k, (\varepsilon_x x^\star_{i'}, 0)\big\rangle$; but
- $\langle \widetilde{w}_k, h_{i'-1}\rangle$ is roughly $\mathcal{N}(0, \frac{2}{m})$ because $\|h_{i'-1}\| \approx 1$ by random init. (see Lemma B.1a).

Thus, if we treat $\langle \widetilde{w}_k, h_{i'-1}\rangle$ as the "noise $n$" in Lemma 5.1 it can be $\frac{1}{\varepsilon_x}$ times larger than $\langle \widetilde{a}_k, x_{i'}\rangle$.

To show Lemma 5.2a, we only need to focus on $|\langle \widetilde{w}_k, h_{i'-1}^{(0)}\rangle| \le \frac{\varepsilon_c}{\sqrt{m}}$ because $i = i'$. Since $h^{(0)}$ can be shown close to $h$ (see Lemma D.1), this is almost equivalent to $|\langle \widetilde{w}_k, h_{i'-1}\rangle| \le \frac{\varepsilon_c}{\sqrt{m}}$. Conditioning on this happens, the "noise $n$" must be small so we can apply Lemma 5.1a.

To show Lemma 5.2a, we can show when $i' \ne i$, the indicator on $|\langle \widetilde{w}_k, h_{i-1}\rangle| \le \frac{\varepsilon_c}{\sqrt{m}}$ gives *little* information about the true noise $\langle \widetilde{w}_k, h_{i'-1}\rangle$. This is so because $h_{i-1}$ and $h_{i'-1}$ are somewhat uncorrelated (details in Lemma B.1k). As a result, the "noise $n$" is still large and thus Lemma 5.1b applies with $\Phi_{i\to j,r,s}(0) = 0$. $\square$

### 5.3 Fitting the Target Function

We are now ready to design $W^* \in \mathbb{R}^{m\times m}$ using Lemma 5.2.

**Definition 5.3.** *Suppose $\varepsilon_e \in \big(0, \frac{1}{\mathfrak{C}_\mathfrak{s}(\Phi, O(\sqrt{L}))}\big)$, $C' = \mathfrak{C}_{\varepsilon_e}(\Phi, \sqrt{L})$, $\varepsilon_x \in \big(0, \frac{1}{\rho^4 C'}\big)$, we choose*

$$\varepsilon_c \stackrel{\text{def}}{=} \frac{\varepsilon_e \varepsilon_x}{4C'} \quad , \quad C \stackrel{\text{def}}{=} \frac{4(C')^2}{\varepsilon_e\varepsilon_x} \quad , \quad C_{i\to j,s} \stackrel{\text{def}}{=} \frac{1}{m}\left\|\mathbf{e}_s^\top \mathsf{Back}_{i\to j}^{(0)}\right\|_2^2 \|h_{i-1}^{(0)}\|^2 \ .$$

*We construct $W^* \in \mathbb{R}^{m\times m}$ by defining its $k$-th row vector as follows:*

$$w_k^* \stackrel{\text{def}}{=} \sum_{i=2}^{L-1}\sum_{j=i+1}^{L}\sum_{r\in[p],s\in[d]} \frac{1}{mC_{i\to j,s}}\left[\mathbf{e}_s^\top \mathsf{Back}_{i\to j}^{(0)}\right]_k \mathbb{1}_{|\langle w_k, h_{i-1}^{(0)}\rangle| \le \frac{\varepsilon_c}{\sqrt{m}}} H_{i\to j,r,s}(a_k) h_{i-1}^{(0)}$$

$$\text{where} \quad C_{i\to j,s} \stackrel{\text{def}}{=} \frac{1}{m}\left\|\mathbf{e}_s^\top \mathsf{Back}_{i\to j}^{(0)}\right\|_2^2 \|h_{i-1}^{(0)}\|^2$$

*Above, functions $H_{i\to j,r,s}\colon \mathbb{R} \to \big[-C, C\big]$ come from Lemma 5.2.*

The following lemma that says $f_{j',s'}$ is close to the target function $F^*_{j',s'}$.

**Lemma 5.4** (existence through backward)**.** *The construction of $W^*$ in Definition 5.3 satisfies the following. For every normalized input sequence $x$ generated from $x^\star$ in the support of $\mathcal{D}$, we have with probability at least $1 - e^{-\Omega(\rho^2)}$ over $W, A, B$, it holds for every $3 \le j' \le L$ and $s' \in [d]$*

$$\begin{aligned} f_{j',s'} &\stackrel{\text{def}}{=} \sum_{i'=2}^{j'} \mathbf{e}_{s'}^\top \mathsf{Back}_{i'\to j'} D_{i'} W^* h_{i'-1} \\ &= \sum_{i=2}^{j'-1}\sum_{r\in[p]} \Phi_{i\to j',r,s'}(\langle w^*_{i\to j',r,s'}, x_i^\star\rangle) \pm \Big(p\rho^{11}\cdot O(\varepsilon_e + \mathfrak{C}_\mathfrak{s}(\Phi,1)\varepsilon_x^{1/3} + Cm^{-0.05})\Big) \ . \end{aligned}$$

*Proof sketch of Lemma 5.4.* Using definition of $f_{j',s'}$ in (5.1) and $W^*$, one can write down

$$\begin{aligned} f_{j',s'} = \sum_{i',j',j}\sum_{r\in[p],s\in[d]}\sum_{k\in[m]} \bigg(&\frac{1}{mC_{i\to j',s}}\left[\mathbf{e}_{s'}^\top \mathsf{Back}_{i'\to j'}\right]_k \left[\mathbf{e}_s^\top \mathsf{Back}_{i\to j}^{(0)}\right]_k \\ &\times \mathbb{1}_{|\langle w_k, h_{i-1}^{(0)}\rangle| \le \frac{\varepsilon_c}{\sqrt{m}}} \mathbb{1}_{[g_{i'}]_k \ge 0} H_{i\to j,r,s}(a_k)\langle h_{i'-1}, h_{i-1}^{(0)}\rangle\bigg) \end{aligned} \tag{5.3}$$

Now,

- The summands in (5.3) with $i \neq i'$ are negligible owing to Lemma 5.2b.
- The summands in (5.3) with $i = i'$ but $j \neq j'$ are negligible, after proving that $\mathsf{Back}_{i\to j}$ and $\mathsf{Back}_{i\to j'}$ are very uncorrelated (details in Lemma C.1).
- The summands in (5.3) with $s \neq s'$ are negligible using the randomness of $B$.
- One can also prove $\mathsf{Back}_{i'\to j'} \approx \mathsf{Back}_{i'\to j'}^{(0)}$ and $h_{i'-1} \approx h_{i'-1}^{(0)}$ (details in Lemma D.1).

Together,

$$f_{j',s'} \approx \sum_{i'=2}^{j'-1}\sum_{r\in[p]}\sum_{k\in[m]} \left(\frac{1}{mC_{i'\to j',s'}}\Big(\left[\mathbf{e}_{s'}^\top \mathsf{Back}_{i'\to j'}^{(0)}\right]_k\Big)^2 \mathbb{1}_{|\langle w_k, h_{i'-1}^{(0)}\rangle| \le \frac{\varepsilon_c}{\sqrt{m}}} \mathbb{1}_{[g_{i'}]_k \ge 0} H_{i'\to j,r,s'}(a_k)\|h_{i'-1}^{(0)}\|^2\right)$$

Applying Lemma 5.2a and using our choice of $C_{i'\to j',s'}$, this gives (in expectation)

$$f_{j',s'} \approx \textstyle\sum_{i=2}^{j'-1}\sum_{r\in[p]} \Phi_{i\to j',r,s'}(\langle w^*_{i\to j',r,s'}, x_i^\star\rangle) = F^*_{j',s'}(x^\star) \ .$$

Proving concentration (with respect to $k \in [m]$) is a lot more challenging due to the sophisticated correlations across different indices $k$. To achieve this, we replace some of the pairs $w_k, a_k$ with fresh new samples $\widetilde{w}_k, \widetilde{a}_k$ for all $k \in \mathcal{N}$ and apply concentration only with respect to $k \in \mathcal{N}$. Here, $\mathcal{N}$ is a random subset of $[m]$ with cardinality $m^{0.1}$. We show that the network stabilizes (details in Section E) against such re-randomization. Full proof is in Section G.3. □

Finally, one can show $\|W^*\|_F \leq O\big(\frac{p\rho^3 C}{\sqrt{m}}\big)$ (see Claim G.1). Crucially, this Frobenius norm scales in $m^{-1/2}$ so standard SGD analysis shall ensure that our sample complexity does not depend on $m$ (up to log factors).

# 6 Coupling and First-Order Approximation

Consider now the scenario when the random initialization matrix $W$ is perturbed to $W + W'$ with $W'$ being small in spectral norm. Intuitively, this $W'$ will later capture how much SGD has moved away from the random initialization, so it may depend on the randomness of $W, A, B$. To untangle this possibly complicated correlation, all lemmas in this section hold for *all* $W'$ being small.

The first lemma below states that the $j$-th layer output difference $B(h_j + h'_j) - Bh_j$ can be approximated by a *linear* function in $W'$, that is $\sum_{i=1}^{j-1} \mathsf{Back}_{i\to j} D_{i+1} W' h_i$. We remind the reader that this linear function in $W'$ is exactly the same as our notation of $f_{j',s'}$ from (5.2).

**Lemma 6.1** (first-order approximation)**.** *Let $W, A, B$ be at random initialization, $x$ be a fixed normalized input sequence, and $\Delta \in [\varrho^{-100}, \varrho^{100}]$. With probability at least $1 - e^{-\Omega(\rho)}$ over $W, A, B$ the following holds. Given any perturbation matrix $W'$ with $\|W'\|_2 \leq \frac{\Delta}{\sqrt{m}}$, letting*

- *$h_\ell, D_\ell, \mathsf{Back}_{i\to j}$ be defined with respect to $W, A, B, x$, and*
- *$h_\ell + h'_\ell, D_\ell + D'_\ell, \mathsf{Back}_{i\to j} + \mathsf{Back}'_{i\to j}$ be defined with respect to $W + W', A, B, x$*

*then*

$$\forall j' \in [L]: \quad \Big\| Bh'_j - \textstyle\sum_{i=1}^{j-1} \mathsf{Back}_{i\to j} D_{i+1} W' h_i \Big\| \leq O\big(\frac{\rho^7 \Delta^{4/3}}{m^{1/6}}\big) \ .$$

The proof of Lemma 6.1 is similar to the semi-smoothness theorem of [2] and can be found in Section H.1.

The next lemma says that, for this linear function $\sum_{i=1}^{j-1} \mathsf{Back}_{i\to j} D_{i+1} \widetilde{W} h_i$ over $\widetilde{W}$, one can replace $h, D, \mathsf{Back}$ with $h + h', D + D', \mathsf{Back} + \mathsf{Back}'$ without changing much in its output. It is a direct consequence of the adversarial stability properties of RNN from prior work (see Section F).

**Lemma 6.2** (first-order coupling)**.** *Let $W, A, B$ be at random initialization, $x$ be a fixed normalized input sequence, and $\Delta \in [\varrho^{-100}, \varrho^{100}]$. With probability at least $1 - e^{-\Omega(\rho)}$ over $W, A, B$ the following holds. Given any matrix $W'$ with $\|W'\|_2 \leq \frac{\Delta}{\sqrt{m}}$, and any $\widetilde{W}$ with $\|\widetilde{W}\|_2 \leq \frac{\omega}{\sqrt{m}}$, letting*

- *$h_\ell, D_\ell, \mathsf{Back}_{i\to j}$ be defined with respect to $W, A, B, x$, and*
- *$h_i + h'_i, D_\ell + D'_\ell, \mathsf{Back}_{i\to j} + \mathsf{Back}'_{i\to j}$ be defined with respect to $W + W', A, B, x$*

*then*

$$\Big\| \sum_{i' \in [j']} \big(\mathsf{Back}_{i'\to j'} + \mathsf{Back}'_{i'\to j'}\big)(D_{i'} + D'_{i'}) \widetilde{W} (h_{i'-1} + h'_{i'-1}) - \sum_{i' \in [j']} \mathsf{Back}_{i'\to j'} D_{i'} \widetilde{W} h_{i'-1} \Big\| \leq O\big(\frac{\omega \rho^6 \Delta^{1/3}}{m^{1/6}}\big) \ .$$

A direct corollary of Lemma 6.2 is that, for our matrix $W^*$ constructed in Definition 5.3 satisfies the same property of Lemma 5.4 after perturbation. Namely,

**Lemma 6.3** (existence after perturbation)**.** *$W^{*}$ in Definition 5.3 satisfies the following. Let $W, A, B$ be at random initialization, $x$ be a fixed normalized input sequence generated by $x^\star$ in the support of $\mathcal{D}$, and $\Delta \in [\varrho^{-100}, \varrho^{100}]$. With probability at least $1 - e^{-\Omega(\rho)}$ over $W, A, B$ the following holds. Given any matrix $W'$ with $\|W'\|_2 \leq \frac{\Delta}{\sqrt{m}}$, any $3 \leq j' \leq L$, and any $s' \in [d]$:*

$$\sum_{i'=2}^{j'} \mathbf{e}_{s'}^\top \big( \mathsf{Back}_{i' \to j'} + \mathsf{Back}'_{i' \to j'} \big)(D_{i'} + D'_{i'}) W^{*} (h_{i'-1} + h'_{i'-1})$$
$$= \sum_{i=2}^{j'-1} \sum_{r \in [p]} \Phi_{i \to j', r, s'}(\langle w^*_{i \to j', r, s'}, x_i^\star \rangle) \pm \Big( p\rho^{11} \cdot O(\varepsilon_e + \mathfrak{C}_{\mathfrak{s}}(\Phi, 1)\varepsilon_x^{1/3} + Cm^{-0.05}) \Big)$$

*Proof.* Combining Lemma 5.4 and Lemma 6.2 gives the proof. □

# 7 Optimization and Convergence

Our main convergence lemma for SGD on the training objective is as follows.

**Lemma 7.1.** *For every constant $\varepsilon \in \big(0, \frac{1}{p \cdot \mathsf{poly}(\rho) \cdot \mathfrak{C}_{\mathfrak{s}}(\Phi, \sqrt{L})}\big)$, there exists $C' = \mathfrak{C}_\varepsilon(\Phi, \sqrt{L})$ and parameters*

$$\varepsilon_x = \frac{1}{\mathsf{poly}(\varepsilon^{-1}, C')} \quad , \quad \varepsilon_c = \frac{1}{\mathsf{poly}(\varepsilon^{-1}, C')} \quad , \quad C = \mathsf{poly}(\varepsilon^{-1}, C') \quad , \quad \lambda = \Theta\big(\frac{\varepsilon}{L\rho}\big)$$

*so that, as long as $m \geq \mathsf{poly}(\varrho)$ and $N \geq \Omega(\frac{\rho^3 p \mathfrak{C}_{\mathfrak{s}}^2(\Phi,1)}{\varepsilon^2})$, setting learning rate $\eta = \Theta\big(\frac{1}{\varepsilon \rho^2 m}\big)$ and $T = \Theta\big(\frac{p^2 C^2 \mathsf{poly}(\rho)}{\varepsilon^2}\big)$, we have*

$$\mathop{\mathbb{E}}_{sgd}\left[\frac{1}{T}\sum_{t=0}^{T-1} \mathop{\mathbb{E}}_{(x^\star, y^\star) \sim \mathcal{Z}} \Big[ \sum_{j=3}^{L} G\left(\lambda F_j(x^\star, W + W_t), y_j^\star\right) \Big]\right] \leq \mathsf{OPT} + \frac{\varepsilon}{2} \ .$$

*and $\|W_t\|_F \leq \frac{\Delta}{\sqrt{m}}$ for $\Delta = \frac{C^2 p^2 \mathsf{poly}(\rho)}{\varepsilon^2}$.*

*Sketch Proof of Lemma 7.1.* The full proof is in Section I and we sketch the main idea here. Recall the training objective

$$\mathsf{Obj}(W_t) \stackrel{\text{def}}{=} \mathbb{E}_{(x^\star, y^\star) \sim \mathcal{Z}} \left[ \mathsf{Obj}(x^\star, y^\star; W_t) \right]$$
$$\text{where} \quad \mathsf{Obj}(x^\star, y^\star; W_t) \stackrel{\text{def}}{=} \textstyle\sum_{j=3}^{L} G\left(\lambda F_j(x^\star; W + W_t), y_j^\star\right)$$

Let $x$ be a normalized input sequence generated by some $x^\star$ in the support of $\mathcal{D}$. Consider an iteration $t$ where the current weight matrix is $W + W_t$. Let

- $h_\ell, g_\ell, D_\ell, \mathsf{Back}_{i \to j}$ be defined with respect to $W, A, B, x$, and
- $h_\ell + h'_{\ell,t}, g_\ell + g'_{\ell,t}, D_\ell + D'_{\ell,t}, \mathsf{Back}_{i \to j} + \mathsf{Back}'_{i \to j, t}$ be defined with respect to $W + W_t, A, B, x$

Let us define function $R_j(x^\star; \widetilde{W}) \in \mathbb{R}^d$ as

$$R_j(x^\star; \widetilde{W}) \stackrel{\text{def}}{=} \sum_{i=2}^{j} \big( \mathsf{Back}_{i \to j} + \mathsf{Back}'_{i \to j, t} \big)(D_i + D'_{i,t}) \widetilde{W} (h_{i-1} + h'_{i-1,t})$$

which is a *linear* function over $\widetilde{W}$. Let us define a loss function $\widetilde{G}$ as:

$$\widetilde{G}(\widetilde{W}) \stackrel{\text{def}}{=} \mathbb{E}_{(x^\star,y^\star)\sim\mathcal{Z}}\left[\widetilde{G}(x^\star,y^\star;\widetilde{W})\right]$$
$$\text{where} \quad \widetilde{G}(x^\star,y^\star;\widetilde{W}) \stackrel{\text{def}}{=} \textstyle\sum_{j=3}^{L} G\left(\lambda F_j(x^\star;W+W_t)+\lambda R_j(x^\star;\widetilde{W}),y_j^\star\right)$$

It is not hard to verify that

$$\mathsf{Obj}(W_t)=\widetilde{G}(0) \quad \text{and} \quad \nabla\mathsf{Obj}(W_t)=\nabla\widetilde{G}(0) \ .$$

Let $W^{\divideontimes}$ be defined in Definition 5.3. By Lemma 6.3, we know that as long as $\|W_t\|_2$ is small (which we shall ensure towards the end),

$$R_j(x^\star;W^{\divideontimes}) \approx F_j^*(x^\star)$$

Thus, by the 1-Lipschitz continuity of $G$, one can derive that

$$\mathbb{E}_{(x^\star,y^\star)\sim\mathcal{Z}}\left[\textstyle\sum_{j=3}^{L} G(R_j(x^\star,W^{\divideontimes}),y_j^\star)\right] \le \mathsf{OPT}+\tfrac{\varepsilon}{100} \ .$$

By Lemma 6.1 and Lemma 6.2 together, we know that

$$\begin{aligned}F_j(x^\star,W+W_t)-F_j(x^\star,W) &\approx \sum_{i=1}^{j-1}\mathsf{Back}_{i\to j}\,D_{i+1}W_t h_i\\ &\approx \sum_{i=1}^{j-1}\left(\mathsf{Back}_{i\to j}+\mathsf{Back}'_{i\to j,t}\right)\left(D_i+D'_{i,t}\right)W_t(h_{i-1}+h'_{i-1,t}) \approx R_j(x^\star,W_t) \ .\end{aligned}$$

Using the linearity of $R_j$ and the 1-Lipschitz continuity of $G$, we have

$$\begin{aligned}\widetilde{G}\left(\frac{1}{\lambda}W^{\divideontimes}-W_t\right) &= \mathbb{E}_{(x^\star,y^\star)\sim\mathcal{Z}}\left[\textstyle\sum_{j=3}^{L} G\left(\lambda F_j(x^\star,W+W_t)+\lambda R_j\left(x^\star,\tfrac{1}{\lambda}W^{\divideontimes}-W_t\right),y_j^\star\right)\right]\\ &\approx \mathbb{E}_{(x^\star,y^\star)\sim\mathcal{Z}}\left[\textstyle\sum_{j=3}^{L} G\left(\lambda F_j(x^\star,W)+R_j(x^\star,W^{\divideontimes}),y_j^\star\right)\right]\\ &\stackrel{①}{\approx} \mathbb{E}_{(x^\star,y^\star)\sim\mathcal{Z}}\left[\textstyle\sum_{j=3}^{L} G\left(R_j(x^\star,W^{\divideontimes}),y_j^\star\right)\right] \pm \tfrac{\varepsilon}{10}\end{aligned}$$

where ① is by our choice of $\lambda$ which implies $\lambda\|F_j(x^\star;W)\| \le \lambda\cdot O(\rho) \le \frac{\varepsilon}{10L}$ by Lemma B.1h.

Together, we have $\widetilde{G}\left(\frac{1}{\lambda}W^{\divideontimes}-W_t\right) \le \mathsf{OPT}+\frac{\varepsilon}{5}$. Thus, by the convexity of $\widetilde{G}(\widetilde{W})$ (composing convex function with linear function is convex), we know

$$\left\langle\nabla\widetilde{G}(0),\tfrac{1}{\lambda}W^{\divideontimes}-W_t\right\rangle \le \widetilde{G}\left(\tfrac{1}{\lambda}W^{\divideontimes}-W_t\right)-\widetilde{G}(0) \le \mathsf{OPT}+\tfrac{\varepsilon}{5}-\widetilde{G}(0) \ . \tag{7.1}$$

Suppose in this high-level sketch that we apply gradient descent as opposed to SGD. Then, $W_{t+1} = W_t - \eta\nabla\widetilde{G}(0)$ and we have

$$\left\|\lambda^{-1}W^{\divideontimes}-W_{t+1}\right\|_F^2 = \left\|\lambda^{-1}W^{\divideontimes}-W_t\right\|_F^2 + \underbrace{2\eta\langle\lambda^{-1}W^{\divideontimes}-W_t,\nabla\widetilde{G}(0)\rangle}_{\diamondsuit} + \eta^2\left\|\nabla\widetilde{G}(0)\right\|_F^2 \ .$$

Putting (7.1) into this formula, we know that as long as $\widetilde{G}(0) > \mathsf{OPT}+\frac{\varepsilon}{5}$, then $\diamondsuit$ is a very negative term and thus, when $\eta$ is sufficiently small, it guarantees to decrease $\left\|\lambda^{-1}W^{\divideontimes}-W_{t+1}\right\|_F$. This cannot happen for too many iterations, and thus we arrive at a convergence statement. □

# 8 Rademacher Complexity Through Coupling

We have the following simple lemma about the Rademacher complexity of RNNs. It first uses the coupling Lemma 6.1 to reduce the network to a linear function, and then calculates the Rademacher complexity for this linear function class.

**Lemma 8.1** (Rademacher complexity of RNNs). *For every $s \in [d]$, $\ell \in [L]$, we have that*

$$\frac{1}{N} \mathop{\mathbb{E}}_{\xi \in \{\pm 1\}^N} \left[ \sup_{\|W'\|_F \le \Delta/\sqrt{m}} \sum_{q \in [N]} \xi_q F_{j,s}(x_q^\star; W + W') \right] \le O\left( \frac{\rho^7 \Delta^{4/3}}{m^{1/6}} + \frac{\rho^2 \Delta}{\sqrt{N}} \right)$$

*Proof.* By Lemma 6.1, we have that

$$\left| F_{j,s}(x_q^\star; W + W') - \sum_{i=1}^{j-1} \mathbf{e}_s^\top \, \mathsf{Back}_{q,i\to j} \, D_{q,i+1} W' h_{q,i} \right|_2 \le O\big(\frac{\rho^7 \Delta^{4/3}}{m^{1/6}}\big)$$

where we use $\mathsf{Back}_{q,i\to j}, h_{q,i}$ and $D_{q,i}$ to denote that calculated from sample $x_q^\star$. Since this function is *linear* in $W'$, we can write it as

$$\langle W', G_q \rangle \stackrel{\text{def}}{=} \sum_{i=1}^{j-1} \mathbf{e}_s^\top \, \mathsf{Back}_{q,i\to j} \, D_{q,i+1} W' h_{q,i} \enspace .$$

We have $\|G_q\|_F \le O(L\rho\sqrt{m/d})$ from Lemma B.1. We bound the Rademacher complexity of this linear function using Proposition A.3 as follows.

$$\frac{1}{N} \mathop{\mathbb{E}}_{\xi \in \{\pm 1\}^N} \left[ \sup_{\|W'\|_F \le \Delta/\sqrt{m}} \sum_{q \in [N]} \xi_q \langle W', G_q \rangle \right] \le O\left( \frac{\rho^2}{\sqrt{d}} \frac{\Delta}{\sqrt{N}} \right) \enspace . \qquad \square$$

# 9 Proof of Theorem 1

Recall $\rho \stackrel{\text{def}}{=} 100Ld\log m$ and $\varrho \stackrel{\text{def}}{=} \frac{100Ldp\cdot\mathfrak{C}_\mathfrak{s}(\Phi,\sqrt{L\log(1/\varepsilon)})\cdot\log m}{\varepsilon}$.

**Theorem 1** (restated). *For every $\varepsilon \in \big(0, \frac{1}{\mathsf{poly}(\rho)\cdot p\cdot \mathfrak{C}_\mathfrak{s}(\Phi, O(\sqrt{L}))}\big)$, define complexity $C = \mathfrak{C}_\varepsilon(\Phi, \sqrt{L})$ and $\lambda = \frac{\varepsilon}{10L\rho}$, if the number of neurons $m \ge \mathsf{poly}(C, p, L, d, \varepsilon^{-1})$ and the number of samples is $N \ge \mathsf{poly}(C, p, L, d, \varepsilon^{-1})$, then SGD with $\eta = \Theta\big(\frac{1}{\varepsilon\rho^2 m}\big)$ and*

$$T = \Theta\Big(\frac{p^2 C^2 \mathsf{poly}(\rho)}{\varepsilon^2}\Big)$$

*satisfies that, with probability at least $1 - e^{-\Omega(\rho^2)}$ over the random initialization*

$$\mathop{\mathbb{E}}_{sgd}\left[ \frac{1}{T} \sum_{t=0}^{T-1} \mathop{\mathbb{E}}_{(x^\star, y^\star)\sim\mathcal{D}} \Big[ \sum_{j=3}^{L} G\left(F_j(x^\star; W + W_t), y_j^\star\right) \Big] \right] \le \mathsf{OPT} + \varepsilon \enspace . \tag{9.1}$$

*Proof of Theorem 1.* One can first apply Lemma 7.1 to obtain $W_t$ for $t = 0, 1, \dots, T-1$ satisfying (recall (I.1))

$$\|W_t\|_F \le \frac{\Delta}{\sqrt{m}} \quad \text{for} \quad \Delta = \Theta\big(\frac{C^2 \rho^{11} p^2}{\varepsilon^2}\big)$$

We can also apply Lemma B.1h together with Lemma F.1a to derive that for each fixed $(x^\star, y^\star) \sim \mathcal{D}$, with probability at least $1 - e^{-\Omega(\rho^2)}$, it satisfies for every $j = 3, 4, \dots, L$,

$$\|F_j(x^\star; W + W_t)\| \le O(\rho^7 \Delta)$$

and therefore by the 1-Lipschitz continuity of $G(\cdot, y^\star)$,

$$\Big| \sum_{j=3}^{L} G\left(\lambda F_j(x^\star; W + W_t), y_j^\star\right) \Big| \le O(\varepsilon \rho^6 \Delta) \enspace . \tag{9.2}$$

Plugging in the Rademacher complexity Lemma 8.1 together with the choice $b = O(\varepsilon\rho^6\Delta)$ into standard generalization argument (see Corollary A.2), we have with probability at least $1-e^{-\Omega(\rho^2)}$, for all $t$

$$\left| \mathop{\mathbb{E}}_{(x^\star,y^\star)\sim\mathcal{Z}} \left[ \sum_{j=3}^{L} G\left(\lambda F_j(x^\star; W+W_t), y_j^\star\right) \right] - \mathop{\mathbb{E}}_{(x^\star,y^\star)\sim\mathcal{D}} \left[ \sum_{j=3}^{L} G\left(\lambda F_j(x^\star; W+W_t), y_j^\star\right) \right] \right| \tag{9.3}$$

$$\leq \lambda \cdot O\left( \frac{\rho^9 \Delta^{4/3}}{m^{1/6}} + \frac{\rho^4 \Delta}{\sqrt{N}} \right) + O\left( \frac{\rho^2 b}{\sqrt{N}} \right) \leq \frac{\varepsilon}{2}$$

where the additional factor $\lambda$ is because we have scaled $F_j$ with factor $\lambda$. In sum, it suffices to choose $N \geq \Omega\big(\frac{\lambda^2\rho^8\Delta^2}{\varepsilon^2}\big) = \Omega(\rho^6\Delta^2)$ and $N \geq \Omega\big(\frac{\rho^4 b^2}{\varepsilon^2}\big) = \Omega(\mathsf{poly}(\rho)\Delta^2)$. □

*Remark* 9.1. Strictly speaking, there is a subtle issue in the above proof because the loss function $G$ is not absolutely bounded for *all* samples $x^\star, y^\star$ (see (9.2)) so one cannot apply Corollary A.2 directly.[5] We only have the statement that for each sample $(x^\star, y^\star)$, the loss function is bounded by some parameter $b$ *with high probability.* By union bound, this can hold for all the training samples, but possibly not all the *testing* samples. A simple fix here is to apply a truncation (for analysis purpose only) on the loss function $L$ to make it always bounded by $b$. Then, we can apply Corollary A.2: the population risk "$\mathbb{E}_{(x^\star,y^\star)\in\mathcal{D}}[\cdots]$" in (9.3) becomes truncated but the empirical risk "$\mathbb{E}_{(x^\star,y^\star)\in\mathcal{Z}}[\cdots]$" in (9.3) stays unchanged. In other words, the *truncated* population risk must be small according to Corollary A.2. Finally, we can remove this truncation from the population risk, because in the rare event that the loss exceeds $b$, it is at most $\mathsf{poly}(m)\cdot e^{O(L)}$ but the probability for this to happen is only $e^{-\Omega(\rho^2)}$. This is negligible in this expectation $\mathbb{E}_{(x,y)\in\mathcal{D}}[\cdots]$.

[5] In some literature this issue was simply ignored or an absolute bound on $L$ is imposed; however, the only globally absolutely bounded convex function is constant.

# Appendix

## A Rademacher Complexity Review

Let $\mathcal{F}$ be a set of functions $\mathbb{R}^d \to \mathbb{R}$ and $\mathcal{X} = (x_1, \dots, x_N)$ be a finite set of samples. Recall the *empirical Rademacher complexity* with respect to $\mathcal{X}$ of $\mathcal{F}$ is

$$\widehat{\mathfrak{R}}(\mathcal{X}; \mathcal{F}) \stackrel{\text{def}}{=} \mathop{\mathbb{E}}_{\xi \sim \{\pm 1\}^N} \Big[ \sup_{f \in \mathcal{F}} \frac{1}{N} \sum_{i=1}^N \xi_i f(x_i) \Big] \tag{A.1}$$

**Lemma A.1** (Rademacher generalization). *Suppose $\mathcal{X} = (x_1, \dots, x_N)$ where each $x_i$ is generated i.i.d. from a distribution $\mathcal{D}$. If every $f \in \mathcal{F}$ satisfies $|f| \le b$, for every $\delta \in (0,1)$ with probability at least $1-\delta$ over the randomness of $\mathcal{Z}$, it satisfies*

$$\sup_{f \in \mathcal{F}} \left| \mathop{\mathbb{E}}_{x \sim \mathcal{D}}[f(x)] - \frac{1}{N} \sum_{i=1}^N f(x_i) \right| \le 2\widehat{\mathfrak{R}}(\mathcal{Z}; \mathcal{F}) + O\left( \frac{b\sqrt{\log(1/\delta)}}{\sqrt{N}} \right) .$$

**Corollary A.2.** *If $\mathcal{F}_1, \dots, \mathcal{F}_k$ are $k$ classes of functions $\mathbb{R}^d \to \mathbb{R}$ and $L_x \colon \mathbb{R}^k \to [-b, b]$ is a 1-Lipschitz continuous function for any $x \sim \mathcal{D}$, then*

$$\sup_{f_1 \in \mathcal{F}_1, \dots, f_k \in \mathcal{F}_k} \left| \mathop{\mathbb{E}}_{x \sim \mathcal{D}}[L_x(f_1(x), \dots, f_k(x))] - \frac{1}{N} \sum_{i=1}^N L_x(f(x_i)) \right| \le O\Big( \sum_{r=1}^k \widehat{\mathfrak{R}}(\mathcal{Z}; \mathcal{F}_r) \Big) + O\left( \frac{b\sqrt{\log(1/\delta)}}{\sqrt{N}} \right) .$$

*Proof.* Let $\mathcal{F}'$ be the class of functions by composing $L$ with $\mathcal{F}_1, \dots, \mathcal{F}_k$, that is, $\mathcal{F}' = \{L_x \circ (f_1, \dots, f_k) \mid f_1 \in \mathcal{F}_1 \cdots f_k \in \mathcal{F}_k\}$. By the (vector version) of the contraction lemma of Rademacher complexity[6] it satisfies $\widehat{\mathfrak{R}}(\mathcal{Z}; \mathcal{F}') \le O(1) \cdot \sum_{r=1}^k \widehat{\mathfrak{R}}(\mathcal{Z}; \mathcal{F}_r)$. □

We recall the simple calculation of the Rademacher complexity for linear function class.

**Proposition A.3.** *Suppose $\|x\|_2 = 1$ for all $x \in \mathcal{X}$. The class $\mathcal{F} = \{x \mapsto \langle w, x \rangle \mid \|w\|_2 \le B\}$ has Rademacher complexity $\widehat{R}(\mathcal{X}; \mathcal{F}) \le O(\frac{B}{\sqrt{N}})$.*

## B Random Initialization: Basic Properties

We first note some important properties about the random initialization of our RNNs. Some of them have already appeared in [2], and the remaining ones can be easily derived from [2].

Let $W, A, B$ be at random initialization and $x_1, \dots, x_L$ be any fixed normalized input sequence (see Definition 3.2). Recall $h_0 = 0$ and

$$g_\ell = W \cdot h_{\ell-1} + Ax_\ell \in \mathbb{R}^m \qquad h_\ell = \sigma(W \cdot h_{\ell-1} + Ax_\ell) \in \mathbb{R}^m .$$

$$(D_\ell)_{k,k} = \mathbb{1}_{(g_\ell)_k \ge 0} \qquad \mathsf{Back}_{\ell \to a} = BD_aW \cdots D_{\ell+1}W \in \mathbb{R}^{d \times m} .$$

*Remark.* We note two main difference between the setting here and [2]. We focus on a single input sequence $x_1, \dots, x_L$, while in [2] the authors study a set of $n$ fixed input sequences. In addition, we focus on a normalized input sequence so that $\|x_1\| = 1$ and $\|x_2\| = \cdots = \|x_L\| = \varepsilon_x$ have small Euclidean norm; instead in [2], the authors study the case when $x_2, \dots, x_L$ can have Euclidean norm up to 1. This is why we can have $\|h_\ell\| \le 2$ but they have $\|h_\ell\| \le O(\ell)$.

[6] There are slightly different versions of the contraction lemma in the literature. For the scalar case without absolute value, see [21, Section 3.8]; for the scalar case with absolute value, see [6, Theorem 12]; and for the vector case without absolute value, see [22].

**Lemma B.1.** *For every $\varepsilon_x \in [0, 1/L]$, there exists a function $\zeta_n(\varepsilon_x, \ell) = \sqrt{1 + (\ell - 1)\varepsilon_x^2} \subseteq [1, 2]$ such that the following holds. For every normalized input sequence $x_1, \ldots, x_L$ (see Definition 3.2), with probability at least $1 - e^{-\Omega(\rho^2)}$ over $W, A, B$*

*(a)* $\big|\|h_\ell\|_2 - \zeta_n(\varepsilon_x, \ell)\big| \leq O\big(\frac{\rho^2}{\sqrt{m}}\big)$ *for every $\ell \in [L]$* *(hidden layer signal)*

*(b)* $\big|\|g_\ell\|_2 - \sqrt{2}\zeta_n(\varepsilon_x, \ell)\big| \leq O\big(\frac{\rho^2}{\sqrt{m}}\big)$ *for every $\ell \in [L]$* *(hidden layer signal)*

*(c)* $\|Wh_\ell\|_\infty, \|Ax_\ell\|_\infty, \|g_\ell\|_\infty \leq O\big(\frac{\rho}{\sqrt{m}}\big)$ *for every $\ell \in [L]$.* *(infinity norm bound)*

*(d)* $\big\{k \in [m] \,\big|\, |[g_\ell]_k| \leq \frac{s}{\sqrt{m}}\big\} \leq O(sm)$ *for every $\ell \in [L], s \in (\frac{\rho^2}{m}, 1]$.* *(alpha-good property)*

*(e)* $\|D_j W_j \cdots D_i W_i u\| \leq \big(1 + \frac{1}{100L}\big)^{j-i+1}\|u\|$ *for all $1 \leq i \leq j \leq L$ and fixed $u \in \mathbb{R}^m$* *(intermediate bound)*

*(f)* $\big|\mathbf{e}_r^\top \mathsf{Back}_{i\to j}\, \mathbf{e}_k\big| \leq O\big(\frac{\rho}{\sqrt{d}}\big)$ *for every $k \in [m], r \in [d], 1 \leq i \leq j \leq L$* *(backward signal)*

*(g)* $\big\|\mathbf{e}_r^\top \mathsf{Back}_{i\to j}\big\| \geq \Omega\big(\frac{\sqrt{m}}{\sqrt{d}}\big)$ *for every $r \in [d], 1 \leq i \leq j \leq L$* *(backward signal)*

*(h)* $\|Bh_\ell\| \leq O(\rho)$ *for every $\ell \in [L]$* *(output signal)*

*(i)* $\|D_{\ell_2} W \cdots D_{\ell_1} W\|_2 \leq O(L^3)$ *for every $1 \leq \ell_1 \leq \ell_2 \leq L$* *(intermediate spectral norm)*

*(intermediate spectral norm)*

*(j)* $|u^\top W D_{\ell_2} W \cdots D_{\ell_1} W v| \leq O\big(\frac{\sqrt{s}\rho}{\sqrt{m}}\big)\|u\|\|v\|$ *for all $s$-sparse vectors $u, v \in \mathbb{R}^m$ with $s \in [1, \frac{m}{\rho^3}]$ and $1 \leq \ell_1 \leq \ell_2 \leq L$* *(intermediate sparse spectral norm)*

*(k)* $\|(I - U_{\ell-1}U_{\ell-1}^\top)h_\ell\| \geq \Omega(\frac{1}{L^2 \log^3 m})$ *. for every $\ell \in [L]$* *(forward correlation)*

*Above,* $U_\ell \stackrel{\text{def}}{=} \mathsf{GS}\,(h_1, h_1, h_2, \ldots, h_\ell)$.

## B.1 Proof for Lemma B.1

*Proof of Lemma B.1.*

(a) The "forward propagation" part of "basic properties at random initialization" of [2] in fact shows (e.g. their Claim B.4 and B.5 of version 3) that, for every $\delta \in \big[\frac{m}{\rho}, \frac{1}{10L}\big]$, with probability at least $1 - e^{-\Omega(\delta^2 m)}$ over $W$ and $A$, for every $\ell = 2, 3, \ldots, L$

$$(1 - \delta) \leq \|h_1\| \leq (1 + \delta)$$
$$(1 - \delta)\sqrt{\|h_\ell\|^2 + \varepsilon_x^2} \leq \|h_{\ell+1}\| \leq (1 + \delta)\sqrt{\|h_\ell\|^2 + \varepsilon_x^2} \quad \text{for all } \ell = 1, 2, \ldots, L - 1$$

Now, suppose we define $\zeta_n(\ell) = \sqrt{1 + (\ell - 1)\varepsilon_x^2}$ then it is trivial to verify by induction

$$(1 - \delta)^\ell \zeta_n(\ell) \leq \|h_\ell\| \leq (1 + \delta)^\ell \zeta_n(\ell)$$

Finally, letting $\delta = \frac{\rho}{\sqrt{m}}$ finishes the proof.

(b) This is similar to the proof of Lemma B.1a, except noticing the $\sqrt{2}$ factor: $\sqrt{2}(1-\delta)\sqrt{1 + \varepsilon_x^2} \leq \|g_1\| \leq \sqrt{2}(1 + \delta)\sqrt{1 + \varepsilon_x^2}$.

(c) We have $Wh_\ell = WUU^\top h_\ell$ where $U = \mathsf{GS}(h_1, \ldots, h_L)$. Each entry of $WU$ is i.i.d. from $\mathcal{N}(0, \frac{2}{m})$. For any fixed $z$ we have $\|WUz\|_\infty \leq O(\sqrt{\rho}/\sqrt{m})$ with probability at least $1 - e^{-\Omega(\rho^2)}$. Taking $\epsilon$-net over $z$ and using $\|h_\ell\| \leq 2$ gives the desired bound $\|Wh_\ell\|_\infty \leq O(\rho/\sqrt{m})$. As for the infinity norm bound on $g_\ell = Wh_{\ell-1} + Ax_\ell$ it follows from $\|Wh_\ell\|_\infty$ and a trivial bound on $\|Ax_\ell\|_\infty$.

(d) Again we write $g_\ell = Wh_{\ell-1} + Ax_\ell = WUU^\top h_{\ell-1} + Ax_\ell$. For any fixed $z$ with $\|z\| \in [0.5, 3]$, we know $y = WUz + Ax_\ell \sim \mathcal{N}(0, \frac{2\|x_\ell\|^2+2\|z\|^2}{m}\mathbf{I})$ and thus with probability at least $1 - e^{-\Omega(sm)}$

- $y$ has at most $O(sm)$ coordinates $k$ with $|y_k| \leq \frac{s}{2\sqrt{m}}$.

Since $z$ is of dimension at most $L$, we can apply a standard $\epsilon$-net argument over all possible $z$ with $\|z\| \in [0.5, 3]$ (with $\epsilon = O(s/\sqrt{m})$) and then apply union bound. Since $sm \geq \rho^2$, with probability at least $1 - e^{-\Omega(\rho^2)}$, for *all* $z$ in this range, it satisfies

- $y$ has at most $O(sm)$ coordinates $k$ with $|y_k| \leq \frac{s}{\sqrt{m}}$.

Substituting $z = U^\top h_{\ell-1}$ (which we know $\|h_{\ell-1}\| \in [0.5, 3]$ from Lemma B.1a) finishes the proof.

(e) The "intermediate layers" part of "basic properties at random initialization" of [2] in fact shows (e.g. their Claim B.12 of version 3) that, for a fixed unit vector $u$, with probability at least $1 - e^{-\Omega(m/L^2)}$, it satisfies

$$(1 - \frac{1}{100L})^{a-\ell} \leq \|D_a W \cdots D_{\ell+1} W u\| \leq (1 + \frac{1}{100L})^{a-\ell} \ .$$

(f) Recall $\mathsf{Back}_{\ell\to a} = BD_a W \cdots D_{\ell+1} W \in \mathbb{R}^{d\times m}$ where $D_\ell$ is determined by $A, W$ and inputs $x_1, \dots, x_L$. The "intermediate layers" part of "basic properties at random initialization" of [2] in fact shows (e.g. their Claim B.12 of version 3) that, for a fixed unit vector $u$, with probability at least $1 - e^{-\Omega(m/L^2)}$, it satisfies

$$(1 - \frac{1}{100L})^{L} \leq \|D_a W \cdots D_{\ell+1} W u\| \leq (1 + \frac{1}{100L})^{L} \ .$$

Further using the randomness of $B$, we have with probability at least $1 - e^{-\Omega(\rho^2)}$,

$$|\mathbf{e}_r^\top BD_a W \cdots D_{\ell+1} W u| \leq O(\frac{\rho}{\sqrt{d}})(1 + \frac{1}{100L})^{L} \ .$$

This finishes the proof after plugging in $u = \mathbf{e}_k$.

(g) Using the same as above, we have with probability at least 1/2,

$$|\mathbf{e}_r^\top BD_a W \cdots D_{\ell+1} W u| \geq \Omega(\frac{1}{\sqrt{d}})(1 - \frac{1}{100L})^{L} \ .$$

This finishes the proof after plugging in $u = \mathbf{e}_k$ for all $k \in [m]$ and taking Chernoff bound.

(h) Since $B$ is independent of the randomness of $h_\ell$, we have $\|Bh_\ell\| \leq O(\rho)\|h_\ell\|$ with probability at least $1 - e^{-\Omega(\rho^2)}$.

(i) This is similar to the proof of Lemma B.1f but requires a careful $\epsilon$-net argument. It is already included in the "intermediate layers: spectral norm" part of the "Basic Properties at Random Initialization" of [2] (e.g. Lemma B.11 in version 3).

(j) This can be proved in the same way as Lemma B.1f. It is already included in the "intermediate layers: sparse spectral norm" part of the "Basic Properties at Random Initialization" of [2] (e.g. Lemma B.14 in version 3).

(k) See the "forward correlation" part of [2] (e.g. Lemma B.6 of version 3).

$\square$

# C Random Initialization: Backward Correlation

In this section, we derive a new property regarding the random initialization of RNNs which needs very new proof techniques from the prior work. Again, let $W, A, B$ be at random initialization and $x_1, \dots, x_L$ be any fixed normalized input sequence (see Definition 3.2). We use $g_\ell, h_\ell, D_\ell, \mathsf{Back}_{i\to j}$ to denote those calculated with this input sequence and $W, A, B$.

**Lemma C.1** (backward correlation)**.** *For every $\varepsilon_x \in [0, 1/L]$, every fixed normalized input sequence $x_1, \dots, x_L$, with probability at least $1 - e^{-\Omega(\rho^2)}$ over $W, A, B$: for every $1 \le i \le j < j' \le L$:*

$$\left|\left\langle u^\top \mathsf{Back}_{i\to j}, v^\top \mathsf{Back}_{i\to j'} \right\rangle\right| \le O(m^{3/4}\rho^4) \cdot \|u\|\|v\|$$

Since $\|\mathsf{Back}_{i\to j}\|_2$ is on the magnitude of $\sqrt{m}$, the above Lemma C.1 says that the two vectors $u^\top \mathsf{Back}_{i\to j}$ and $v^\top \mathsf{Back}_{i\to j'}$ are very *uncorrelated* whenever $j \ne j'$.

*Remark.* In fact, one can prove the same Lemma C.1 for the un-correlation between $u^\top \mathsf{Back}_{i\to j}$ and $v^\top \mathsf{Back}_{i'\to j'}$ whenever $j - i \ne j' - i'$. We do not need that stronger version in this paper.

## C.1 Proof of Lemma C.1

*Proof of Lemma C.1.* It suffices to prove for fixed unit vector $u, v \in \mathbb{R}^d$ because we can apply $\epsilon$-net argument and union bound in the end. Recall

$$u^\top \mathsf{Back}_{i\to j} = u^\top B D_j W \cdots D_{i+1} W \quad \text{and} \quad v^\top \mathsf{Back}_{i\to j'} = v^\top B D_{j'} W \cdots D_{i+1} W \ .$$

Let $\xi_1, \dots, \xi_m$ be a random orthonormal basis of $\mathbb{R}^m$, and we denote by

$$\Xi_p \stackrel{\text{def}}{=} \left(u^\top B D_j W \cdots D_{i+1} W \xi_p\right) \cdot \left(v^\top B D_{j'} W \cdots D_{i+1} W \xi_p\right)$$

so it suffices to bound the absolute value of $\sum_{p\in[m]} \Xi_p = \left\langle u^\top \mathsf{Back}_{i\to j}, v^\top \mathsf{Back}_{i\to j'} \right\rangle$.

Let us fix $N$ coordinates (without loss of generality the first $N$ coordinates) and calculate $\sum_{p\in[N]} \Xi_p$ only over $[N]$ by induction. Define

$$z_{1,i} = \xi_1, \quad z_{2,i} = \xi_2, \quad \dots \quad z_{N,i} = \xi_N$$

and define $z_{p,\ell} = D_\ell W \cdots D_{i+1} W z_{p,i}$. We have

$$\|z_{p,\ell}\| \le 2 \quad \text{and} \quad \Xi_p = u^\top B z_{p,j} \cdot v^\top B z_{p,j'}$$

where the first is due to Lemma B.1e. Let

$$Z_{p,\ell} = \mathsf{GS}\Big(h_1, \dots, h_\ell,\ z_{1,i}, \dots, z_{N,i},\ z_{2,i+1}, \dots, z_{N,i+1},\ \dots\ z_{1,\ell}, \dots, z_{p-1,\ell}\Big) \ .$$

Each $Z_{p,\ell}$ is column orthonormal and has at most $(N+1)L$ columns.

In the base case $\ell = i$, we have

$$\|Z_{p,i}^\top z_{p,i}\| = \|Z_{p,i}^\top \xi_p\| \stackrel{①}{=} \|U_i^\top \xi_p\| \stackrel{②}{\le} O(\sqrt{L}\rho/\sqrt{m})$$

Above, ① is by the definition $U_i = \mathsf{GS}(h_1, \dots, h_i)$, and ② is because for each fixed unit vector $u$, we have $|\langle u, \xi_p\rangle| \le O(\rho/\sqrt{m})$ with probability at least $1 - e^{-\Omega(\rho^2)}$.

For any $\ell = i, \dots, L-1$, we have

$$\begin{aligned} Z_{p,\ell+1}^\top z_{p,\ell+1} &= Z_{p,\ell+1}^\top D_{\ell+1}\big(W(I - Z_{p,\ell}Z_{p,\ell}^\top) z_{p,\ell} + W Z_{p,\ell} Z_{p,\ell}^\top z_{p,\ell}\big) \\ &= Z_{p,\ell+1}^\top D_{\ell+1} \frac{W(I - Z_{p,\ell}Z_{p,\ell}^\top) z_{p,\ell}}{\|(I - Z_{p,\ell}Z_{p,\ell}^\top) z_{p,\ell}\|} \|(I - Z_{p,\ell}Z_{p,\ell}^\top) z_{p,\ell}\| + Z_{p,\ell+1}^\top D_{\ell+1} W Z_{p,\ell} Z_{p,\ell}^\top z_{p,\ell} \end{aligned}$$

We consider the two terms on the right hand side separately:

- For the first term, we know that $g = \frac{W(I - Z_{p,\ell} Z_{p,\ell}^\top) z_{p,\ell}}{\|(I - Z_{p,\ell} Z_{p,\ell}^\top) z_{p,\ell}\|}$ is distributed according to $\mathcal{N}(0, \frac{2\mathbf{I}}{m})$ and is independent of $Z_{p,\ell+1}$.[7] Therefore, with probability at least $1 - e^{-\Omega(\rho^2)}$

$$\left\| Z_{p,\ell+1}^\top D_{\ell+1} g \|(I - Z_{p,\ell} Z_{p,\ell}^\top) z_{p,\ell}\| \right\| \le O(\frac{\rho\sqrt{NL}}{\sqrt{m}}) \cdot \|(I - Z_{p,\ell} Z_{p,\ell}^\top) z_\ell\| \le O(\frac{\rho\sqrt{NL}}{\sqrt{m}} \|z_\ell\|) \le O(\frac{\sqrt{N}\rho^2}{\sqrt{m}})$$

- For the second term, we have

$$\left\| Z_{p,\ell+1}^\top D_{\ell+1} W Z_{p,\ell} Z_{p,\ell}^\top z_{p,\ell} \right\| \le \|D_{\ell+1} W Z_{p,\ell} Z_{p,\ell}^\top z_{p,\ell}\|$$

  Letting $y = D_{\ell+1} W Z_{p,\ell} Z_{p,\ell}^\top z_{p,\ell}$, then we have

$$\begin{aligned} y &= \mathbb{1}_{W h_\ell + A x_{\ell+1} \ge 0} \cdot W Z_{p,\ell} Z_{p,\ell}^\top z_{p,\ell} \\ &\overset{①}{=} \mathbb{1}_{W Z_{p,\ell} Z_{p,\ell}^\top h_\ell + A x_{\ell+1} \ge 0} \cdot W Z_{p,\ell} Z_{p,\ell}^\top z_{p,\ell} \\ &\overset{②}{=} \mathbb{1}_{M a + A x_{\ell+1} \ge 0} \cdot M b \end{aligned}$$

  where in ① we have used $h_\ell$ is in the column span of $Z_{p,\ell}$, and in ② we have defined $M = W Z_{p,\ell}$, $a = Z_{p,\ell}^\top h_\ell$, and $b = Z_{p,\ell}^\top z_{p,\ell}$. We stress here that the entries of $M$ and $A$ are i.i.d. in $\mathcal{N}(0, \frac{2}{m})$. Now, suppose for a moment that we view $a$ and $b$ as fixed. Then, it is a simple exercise to verify that with probability $1 - \exp(-\Omega(m/L^2))$ (over the randomness of $M, A$),[8]

$$\|\mathbb{1}_{Ma + Ax \ge 0} M b\| \le \|b\|(1 + \frac{1}{100L}) \ .$$

  After taking $\epsilon$-net over all possible $a, b$, we have that for fixed $x \in \mathbb{R}^d$ and all $a, b$:

$$\|\mathbb{1}_{Ma + Ax \ge 0} M b\| \le \|b\|(1 + \frac{1}{50L}) \ .$$

  We can thus plug in the choice $a = Z_{p,\ell}^\top h_\ell$, and $b = Z_{p,\ell}^\top z_{p,\ell}$ (both of which may depend on the randomness of $W$ and $A$). We have

$$\left\| Z_{p,\ell+1}^\top D_{\ell+1} W Z_{p,\ell} Z_{p,\ell}^\top z_{p,\ell} \right\| \le \|y\| \le \|Z_{p,\ell}^\top z_{p,\ell}\|(1 + \frac{1}{50L}) \ .$$

Combining the above two properties and using induction, we have

$$\|Z_{p,\ell+1}^\top z_{p,\ell+1}\| \le O(\frac{\sqrt{N}\rho^3}{\sqrt{m}})$$

[7]Indeed,

- For each $z_{p',\ell+1}$ with $p' < p$ in the column span of $Z_{p,\ell+1}$, we have $z_{p',\ell+1} = D_{\ell+1} W z_{p',\ell}$ so it only depends on $z_{p',\ell}$ and the randomness of $W z_{p',\ell}$, both of which already included in the column span of $W Z_{p,\ell}$. Therefore, $z_{p',\ell+1}$ is independent of $g$ because $g$ has already projected out all the randomness in $W Z_{p,\ell}$.
- For each $z_{p',j}$ with $p' \in [N]$ and $j \le \ell$ in the column span of $Z_{p,\ell+1}$, we have $z_{p',j} = D_j W z_{p',j-1}$ so it only depends on $z_{p',j-1}$ and the randomness of $W z_{p',j-1}$, both of which already included in the column span of $W Z_{p,\ell}$. Therefore, $z_{p',j}$ is independent of $g$ because $g$ has already projected out all the randomness in $W Z_{p,\ell}$.
- For $h_{\ell+1}$ included in the column span of $Z_{p,\ell+1}$, we have $h_{\ell+1} = D_\ell W h_\ell$ so it only depends on $h_\ell$ and the randomness of $W h_\ell$, both of which already included in the column span of $W Z_{p,\ell}$. Therefore, $h_{\ell+1}$ is independent of $g$ because $g$ has already projected out all the randomness in $W Z_{p,\ell}$.

In sum, $g$ must be independent of $Z_{p,\ell+1}$.

[8]See for instance Claim B.13 of version 3 of [2].

We calculate

$$
\begin{aligned}
\sum_{p\in[N]} \Xi_p &= \sum_{p\in[N]} u^\top B z_{p,j} \cdot v^\top B z_{p,j'} \\
&= \sum_{p\in[N]} u^\top B(I - Z_{p,j}Z_{p,j}^\top) z_{p,j} \cdot v^\top B(I - Z_{p,j'}Z_{p,j'}^\top) z_{p,j'} \\
&\quad + \sum_{p\in[N]} u^\top B Z_{p,j}Z_{p,j}^\top z_{p,j} \cdot v^\top B z_{p,j'} + \sum_{p\in[N]} u^\top B(I - Z_{p,j}Z_{p,j}^\top) z_{p,j} \cdot v^\top B Z_{p,j'}Z_{p,j'}^\top z_{p,j'} \\
&\overset{①}{=} \underbrace{\sum_{p\in[N]} u^\top B(I - Z_{p,j}Z_{p,j}^\top) z_{p,j} \cdot v^\top B(I - Z_{p,j'}Z_{p,j'}^\top) z_{p,j'}}_{\clubsuit} \pm O\left(\frac{N^{1.5}\rho^5}{\sqrt{m}}\right)
\end{aligned}
$$

Above, ① is because for every fixed vector $x$, with probability at least $1 - e^{-\Omega(\rho^2)}$ over $B$ it satisfies $|u^\top B x| \leq O(\rho\|x\|)$, and therefore we have (similarly if we replace $u$ with $v$)

$$
|u^\top B(I - Z_{p,j}Z_{p,j}^\top) z_{p,j}| \leq O(\rho) \quad \text{and} \quad |u^\top B Z_{p,j'}Z_{p,j'}^\top z_{p,j'}| \leq O\big(\frac{\sqrt{N}\rho^4}{\sqrt{m}}\big)
$$

To bound $\clubsuit$, we note that the following $2N$ vectors

$$
\left\{(I - Z_{p,j}Z_{p,j}^\top) z_{p,j}\right\}_{p\in[N]} \bigcup \left\{(I - Z_{p,j'}Z_{p,j'}^\top) z_{p,j'}\right\}_{p\in[N]}
$$

are pairwise orthogonal, and therefore, when left-multiplied with matrix $B$, their randomness (over $B$) are independent. This means, $|\clubsuit| \leq O\big(\sqrt{N}\rho^2\big)$ with probability at least $1 - e^{-\Omega(\rho^2)}$ over $B$. Choosing $N = m^{1/2}$ we have

$$
\Big|\sum_{p\in[m^{1/2}]} \Xi_p\Big| \leq O(m^{1/4}\rho^4) \ .
$$

Finally, we can divide all the $m$ coordinates into $\sqrt{m}$ chunks each of size $\sqrt{m}$. Performing the above calculation for each of them gives the desired bound. □

# D Stability: After Dropping $x^\star$

In this section we consider two (normalized) input sequences. The first sequence $x^{(0)}$ is given as

$$
x_1^{(0)} = (0^{d_x}, 1) \quad \text{and} \quad x_\ell^{(0)} = (0^{d_x}, \varepsilon_x) \text{ for } \ell = 2, 3, \ldots, L \tag{D.1}
$$

The second sequence $x$ is generated from an arbitrary $x^\star = (x_2^\star, \ldots, x_{L-1}^\star)$ in the support of $\mathcal{D}$:

$$
x_1 = (0^{d_x}, 1) \quad \text{and} \quad x_\ell = (\varepsilon_x x_\ell^\star, 0) \text{ for } \ell = 2, 3, \ldots, L-1 \quad \text{and} \quad x_L = (0^{d_x}, \varepsilon_x) \tag{D.2}
$$

We study the following two executions of RNNs under input $x^{(0)}$ and $x$ respectively:

$$
\begin{aligned}
&g_0 = h_0 = 0 && g_0^{(0)} = h_0^{(0)} = 0 \\
&g_\ell = W h_{\ell-1} + A x_\ell && g_\ell^{(0)} = W h_{\ell-1}^{(0)} + A x_\ell^{(0)} && \text{for } \ell \in [L] \\
&h_\ell = \sigma(W h_{\ell-1} + A x_\ell) && h_\ell^{(0)} = \sigma(W h_{\ell-1}^{(0)} + A x_\ell^{(0)}) && \text{for } \ell \in [L]
\end{aligned}
$$

We also define diagonal sign matrices $D_\ell \in \{0,1\}^{m\times m}$ and $D_\ell^{(0)} \in \{0,1\}^{m\times m}$ by letting

$$
(D_\ell)_{k,k} = \mathbb{1}_{[g_\ell]_k \geq 0} \text{ and } (D_\ell^{(0)})_{k,k} = \mathbb{1}_{[g_\ell^{(0)}]_k \geq 0}.
$$

Accordingly,

$$\mathsf{Back}_{\ell\to a} = BD_aW\cdots D_{\ell+1}W \in \mathbb{R}^{d\times m}$$
$$\mathsf{Back}^{(0)}_{\ell\to a} = BD^{(0)}_aW\cdots D^{(0)}_{\ell+1}W \in \mathbb{R}^{d\times m}$$

We have the following lemma.

**Lemma D.1.** *For every $\varepsilon_x \in [0, \frac{1}{10L}]$, there exists a function $\zeta_d(\varepsilon_x, \ell) \in \left[\sqrt{(\ell-1)\varepsilon_x^2/2}, \sqrt{2(\ell-1)\varepsilon_x^2}\right]$ satisfying the following. For every fixed $x^\star$ in $\mathcal{D}$, consider two normalized input sequences $x^{(0)}$ and $x$ defined as (D.1) and (D.2). Let $W, A, B$ be at random initialization, let $h^{(0)}_\ell, D^{(0)}_\ell, \mathsf{Back}^{(0)}_{i\to j}$ be defined with respect to $W, A, B, x^{(0)}$, and $h_\ell, D_\ell, \mathsf{Back}_{i\to j}$ be defined with respect to $W, A, B, x$. With probability at least $1 - e^{-\Omega(\rho^2)}$ over $W, A, B$ we have*

*(a)* $\left|\|h^{(0)}_\ell - h_\ell\|^2 - \zeta_d(\varepsilon_x, \ell)^2\right| \le O\left(\frac{\rho}{\sqrt{m}}\right)$ *for every $\ell \in [L]$*

*(b)* $\|D^{(0)}_\ell - D_\ell\|_0 \le O((\sqrt{L}\varepsilon_x)^{2/3} m) \le O(L^{1/3}\varepsilon_x^{2/3} m)$ *for every $\ell \in [L]$*

*(c)* $\left\|u^\top \mathsf{Back}^{(0)}_{i\to j} - u^\top \mathsf{Back}_{i\to j}\right\|_2 \le O\left(\rho^{25/6}\varepsilon_x^{1/3}\sqrt{m}\right)\cdot\|u\|$ *for every $1 \le i \le j \le L$ and $u \in \mathbb{R}^d$*

We emphasize that Lemma D.1a is technically the most involved, and the remaining two properties Lemma D.1b and Lemma D.1c are simple corollaries.

## D.1 Proof of Lemma D.1a

*Proof of Lemma D.1a.* We first look at layer $\ell = 1$. We have $h^{(0)}_1 = \sigma(Ax^{(0)}_1) = \sigma(Ax_1) = h_1$ so can set $\zeta_d(\varepsilon_x, 1) = 0$. As for $\ell \ge 2$, we can write $h_\ell - h^{(0)}_\ell$ as:

$$h_\ell - h^{(0)}_\ell = \phi\left(WUU^\top h_{\ell-1} + Ax_\ell\right) - \phi\left(WUU^\top h^{(0)}_{\ell-1} + Ax^{(0)}_\ell\right)$$

where $U = \mathsf{GS}(h^{(0)}_1, \dots, h^{(0)}_L, h_1, \dots, h_L) \in \mathbb{R}^{m\times 2L}$. Let us write $z = U^\top h_{\ell-1}$ and $z_0 = U^\top h^{(0)}_{\ell-1}$. We know that $WU \in \mathbb{R}^{m\times 2L}$ is a random matrix with entries i.i.d. from $\mathcal{N}(0, \frac{2}{m})$. Of course, $z$ and $z_0$ depend on the randomness of $WU$. Since $h_{\ell-1}$ and $h^{(0)}_{\ell-1}$ are in the column span of $U$, we have

$$\|z - z_0\| = \|h^{(0)}_{\ell-1} - h_{\ell-1}\|, \quad \|z\| = \|h_{\ell-1}\|, \quad \|z_0\| = \|h^{(0)}_{\ell-1}\|$$

Let $\widehat{z}_0 = z_0/\|z_0\|$, then we can write

$$z = \langle \widehat{z}_0, z\rangle \widehat{z}_0 + (I - \widehat{z}_0\widehat{z}_0^\top)z = c_1\widehat{z}_0 + c_2\frac{(I - \widehat{z}_0\widehat{z}_0^\top)z}{\|(I - \widehat{z}_0\widehat{z}_0^\top)z\|}$$

with

$$c_1 = \frac{\langle z_0, z\rangle}{\|z_0\|} = \frac{\|z_0\|^2 + \|z\|^2 - \|z_0 - z\|^2}{2\|z_0\|}$$
$$c_2^2 = \|z\|^2 - (\langle z, \widehat{z}_0\rangle)^2 = \|z\|^2 - \frac{(\|z_0\|^2 + \|z\|^2 - \|z_0 - z\|^2)^2}{4\|z_0\|^2} = \|z\|^2 - c_1^2$$

Now, for every $\ell \ge 2$, suppose for now that $z_0$ and $z$ are fixed (as opposed to depending on the randomness of $WU$). Then, by Proposition D.2 (after appropriate re-scaling), we have

$$\mathop{\mathbb{E}}_{WU,A}\left[\left\|\phi\left(WUz + Ax_\ell\right) - \phi\left(WUz_0 + Ax^{(0)}_\ell\right)\right\|_2^2 \,\Big|\, z, z_0\right] = 2\left(\|z\|_2^2 + \varepsilon_x^2\right)\zeta_c(\beta)$$

$$\text{with} \quad \alpha = \sqrt{\frac{c_1^2}{\|z\|_2^2 + \varepsilon_x^2}} = \sqrt{1 - \beta^2}, \quad \beta = \sqrt{\frac{c_2^2 + \varepsilon_x^2}{\|z\|_2^2 + \varepsilon_x^2}}$$

Recall $\|z_0\| = \zeta_n(\varepsilon_x, \ell-1) \pm O(\frac{\rho^2}{\sqrt{m}})$ and $\|z\| = \zeta_n(\varepsilon_x, \ell-1) \pm O(\frac{\rho^2}{\sqrt{m}})$ from Lemma B.1a and $\zeta_n \in [1,2]$. Therefore, using $\zeta_n$ to denote $\zeta_n(\varepsilon_x, \ell-1)$, we have

$$\begin{aligned}\beta^2 &= \frac{\|z\|^2 - \frac{(\|z_0\|^2+\|z\|^2-\|z_0-z\|^2)^2}{4\|z_0\|^2} + \varepsilon_x^2}{\|z\|_2^2 + \varepsilon_x^2} \\ &= \frac{\|z-z_0\|^2 + \frac{\|z-z_0\|^4}{4\zeta_n^2} + \varepsilon_x^2}{\zeta_n^2 + \varepsilon_x^2} \pm O(\frac{\rho^2}{\sqrt{m}})\end{aligned}$$

Using the concentration, we know with probability at least $1 - e^{-\Omega(\rho^2)}$, for any *fixed* $z, z_0$ satisfying $\|z\|, \|z_0\| \in [0.5, 3]$ and $\|z - z_0\| \le 0.1$,

$$\left\|\phi\left(WUz + Ax_\ell\right) - \phi\left(WUz_0 + Ax_\ell^{(0)}\right)\right\|_2^2 = 2\left(\|z\|_2^2 + \varepsilon_x^2\right)\zeta_c(\beta) \pm O\left(\frac{\rho}{\sqrt{m}}\right) \quad .$$

By applying an $\epsilon$-net argument over all possible $z$ and $z_0$, the above formula turns to hold for all $z$ and $z_0$. Therefore, if we pick the special choice $z = U^\top h_{\ell-1}$ and $z_0 = U^\top h_{\ell-1}^{(0)}$ (which depend on the randomness of $WU$), we have

$$\left\|h_\ell - h_\ell^{(0)}\right\|_2^2 = 2\left(\|z\|_2^2 + \varepsilon_x^2\right)\zeta_c(\beta) \pm O\left(\frac{\rho}{\sqrt{m}}\right) \quad .$$

Using again $\|z\| = \zeta_n \pm O(\frac{\rho^2}{\sqrt{m}})$ and the Lipschitz continuity property of $\zeta_c(\sqrt{x})$ from Proposition D.2, we have

$$\begin{aligned}\left\|h_\ell - h_\ell^{(0)}\right\|_2^2 &= 2\left(\zeta_n^2 + \varepsilon_x^2\right)\zeta_c\left(\sqrt{\frac{\|z-z_0\|^2 + \frac{\|z-z_0\|^4}{4\zeta_n^2} + \varepsilon_x^2}{\zeta_n^2 + \varepsilon_x^2}}\right) \pm O\left(\frac{\rho}{\sqrt{m}}\right) \\ &= 2\left(\zeta_n(\varepsilon_x, \ell-1)^2 + \varepsilon_x^2\right)\zeta_c\left(\sqrt{\frac{\|h_{\ell-1} - h_{\ell-1}^{(0)}\|^2 + \frac{\|h_{\ell-1}-h_{\ell-1}^{(0)}\|^4}{4\zeta_n(\varepsilon_x,\ell-1)^2} + \varepsilon_x^2}{\zeta_n(\varepsilon_x, \ell-1)^2 + \varepsilon_x^2}}\right) \pm O\left(\frac{\rho}{\sqrt{m}}\right) \quad .\end{aligned}$$

For such reason, let us define

$$\zeta_d(\varepsilon_x, \ell)^2 \stackrel{\text{def}}{=} 2\left(\zeta_n(\varepsilon_x, \ell-1)^2 + \varepsilon_x^2\right)\zeta_c\left(\sqrt{\frac{\zeta_d(\varepsilon_x, \ell-1)^2 + \frac{\zeta_d(\varepsilon_x,\ell-1)^4}{4\zeta_n(\varepsilon_x,\ell-1)^2} + \varepsilon_x^2}{\zeta_n(\varepsilon_x, \ell-1)^2 + \varepsilon_x^2}}\right) \quad .$$

We analyze two things:

- Recall $-|\beta|^3 \le \zeta_c(\beta) - \frac{\beta^2}{2} \le \frac{\beta^4}{4}$ from Proposition D.2. This means
$$\left|\zeta_d(\varepsilon_x, \ell)^2 - \left(\zeta_d(\varepsilon_x, \ell-1)^2 + \frac{\zeta_d(\varepsilon_x, \ell-1)^4}{4\zeta_n(\varepsilon_x, \ell-1)^2} + \varepsilon_x^2\right)\right| \le O\left(\zeta_d(\varepsilon_x, \ell-1)^3\right) \quad .$$
Applying induction, we have $\zeta_d(\varepsilon_x, \ell)^2 \in [\frac{1}{2}(\ell-1)\varepsilon_x^2, 2(\ell-1)\varepsilon_x^2]$ since $\varepsilon_x \le \frac{1}{10L}$.
- Suppose $\left|\|h_{\ell-1} - h_{\ell-1}^{(0)}\|^2 - \zeta_d(\varepsilon_x, \ell-1)^2\right| \le \xi_{\ell-1}$, then, by the $\frac{1}{2}$-Lipschitz continuity of $\zeta_c(\sqrt{x})$,

we have

$$
\begin{aligned}
&\left|\|h_\ell - h_\ell^{(0)}\|^2 - \zeta_d(\varepsilon_x,\ell)^2\right| \\
&\le \left| \|h_{\ell-1} - h_{\ell-1}^{(0)}\|^2 + \frac{\|h_{\ell-1} - h_{\ell-1}^{(0)}\|^4}{4\zeta_n(\varepsilon_x,\ell-1)^2} - \zeta_d(\varepsilon_x,\ell-1)^2 - \frac{\zeta_d(\varepsilon_x,\ell-1)^4}{4\zeta_n(\varepsilon_x,\ell-1)^2}\right| + O\left(\frac{\rho}{\sqrt{m}}\right) \quad . \\
&\le \xi_{\ell-1} + O\left(\xi_{\ell-1}\ell\varepsilon_x^2 + \frac{\rho}{\sqrt{m}}\right) \le (1+\frac{1}{2L})\xi_{\ell-1}
\end{aligned}
$$

Thus, we can define $\xi_\ell = (1+\frac{1}{2L})\xi_{\ell-1}$ and applying induction we finish the proof. □

## D.2 Proof of Lemma D.1b and D.1c

*Proof of Lemma D.1b and D.1c.*

(b) Since $\|h_\ell^{(0)} - h_\ell\| \le O(\sqrt{L}\varepsilon_x)$ by Lemma D.1a, we have $\|g_\ell^{(0)} - g_\ell\| \le O(\sqrt{L}\varepsilon_x)$. Thus, combining Lemma B.1d and Fact D.3, we can derive that if $g_\ell$ is affected in Euclidean norm by $\tau$, then with probability at least $1 - e^{-\Omega(\rho^2)}$, the number of sign changes is at most $O(\tau^{2/3}m + \rho^2)$. In other words, we have

$$\|D_\ell^{(0)} - D_\ell\|_0 \le s \stackrel{\text{def}}{=} O((\sqrt{L}\varepsilon_x)^{2/3}m) \le O(L^{1/3}\varepsilon_x^{2/3}m)$$

(c) By definition, letting $D'_\ell = D_\ell^{(0)} - D_\ell$,

$$u^\top \mathsf{Back}_{i\to j}^{(0)} - u^\top \mathsf{Back}_{i\to j} = u^\top B(D_j + D'_j)W \dots (D_{i+1} + D'_{i+1})W - u^\top B D_j W \dots D_{i+1} W$$

One can then expand out all the $2^{j-i} - 1$ terms in the above difference formula. Ignoring the subscripts, we recall three basic properties

- Euclidean norm $\|u^\top B(WD\cdots W)D'\| \le O(\rho\sqrt{s})\|u\| \le O(\rho^{7/6}\varepsilon_x^{1/3}m^{1/2})\|u\|$ from Lemma B.1f.
- Spectral norm $\|D'(WD\cdots W)D'\|_2 \le O(\rho\sqrt{s/m}) \le \frac{1}{100L}$ from Lemma B.1j.
- Spectral norm $\|D'(WD\cdots W)\|_2 \le O(L^3)$ from Lemma B.1i.

Therefore, we have

$$
\begin{aligned}
&\|u^\top B(D_j + D'_j)W \dots (D_{i+1} + D'_{i+1})W - u^\top B D_j W \dots D_{i+1} W\| \\
&\le L \cdot \Big(O(\rho^{7/6}\varepsilon_x^{1/3}m^{1/2})\|u\| \cdot O(L^3)\Big) + L^2 \cdot \Big(O(\rho^{7/6}\varepsilon_x^{1/3}m^{1/2})\|u\| \cdot O(L^3)\Big) \cdot \frac{1}{100L} \\
&\quad + L^3 \cdot \Big(O(\rho^{7/6}\varepsilon_x^{1/3}m^{1/2})\|u\| \cdot O(L^3)\Big) \cdot \frac{1}{(100L)^2} + \dots \\
&\le O(\rho^{25/6}\varepsilon_x^{1/3}m^{1/2})
\end{aligned}
$$

□

## D.3 Mathematical Tools

**Proposition D.2** (Gaussian difference with ReLU). *Let $g_1, g_2$ be two independent standard Gaussian random variable $\mathcal{N}(0,1)$, and let parameters $\beta \in [-\frac{3}{4}, \frac{3}{4}]$ and $\alpha = \sqrt{1-\beta^2} \in [\frac{3}{4}, \frac{5}{4}]$. Define*

$$\zeta_c(\beta) \stackrel{\text{def}}{=} \mathbb{E}\left[|\sigma(g_1) - \sigma(\alpha g_1 + \beta g_2)|_2^2\right]$$

*We have:*

- *For* $\beta \in [-\frac{3}{4}, \frac{3}{4}]$ *we have* $-|\beta|^3 \leq \zeta_c(\beta) - \frac{\beta^2}{2} \leq \frac{\beta^4}{4}$.
- *Over* $x \in [-0.05, 0.05]$, *the function* $\zeta_c(\sqrt{x})$ *is* $\frac{1}{2}$*-Lipschitz continuous over* $x$.

*Proof.* At least for the range of $\alpha \in \left[\frac{3}{4}, \frac{5}{4}\right]$ and $\beta \geq 0$, one can exactly integrate out this squared difference over two Gaussian variables. For instance, the "$\delta$-separateness" part of "properties at random initialization" of [3] (e.g. Claim A.6 of their version 3) has already done this for us:

$$2\zeta_c(\alpha, \beta) = 2\,\mathbb{E}\left[|\sigma(g_1) - \sigma(\alpha g_1 + \beta g_2)|_2^2\right] = \left(\alpha^2 - 2\alpha + 1\right) + \beta^2 + \frac{2}{\pi}\sum_{k=1}^{\infty}(-1)^k \frac{(\alpha + k)\beta^{2k+1}}{(2k+1)\alpha^{2k+1}}$$
$$= \frac{\delta\left(-2\alpha^2 + \alpha - 2\delta^2\right)}{\pi\left(\alpha^2 + \delta^2\right)} + \frac{(1 - 2\alpha)\arctan\left(\frac{\alpha}{\delta}\right)}{\pi} + (\alpha - 1)\alpha + \delta^2 + \frac{1}{2}$$

It is easy to see that, as long as $\beta \leq \alpha$, we always have $\frac{(\alpha+k)\beta^{2k+1}}{(2k+1)\alpha^{2k+1}} \geq \frac{(\alpha+k+1)\beta^{2k+3}}{(2k+3)\alpha^{2k+3}}$. Therefore

$$\left(\alpha^2 - 2\alpha + 1\right) + \beta^2 \geq 2\,\mathop{\mathbb{E}}_{g_1,g_2}\left[(\sigma(g_1) - \sigma(\alpha g_1 + \beta g_2))^2\right] \geq \left(\alpha^2 - 2\alpha + 1\right) + \beta^2 - \frac{2}{\pi}\frac{(\alpha + 1)\beta^3}{3\alpha^3}$$
$$\geq \left(\alpha^2 - 2\alpha + 1\right) + \beta^2 - \beta^3$$

- For the value approximation, we have $1 - \alpha \geq \zeta_c(\alpha, \beta) \geq 1 - \alpha - \beta^3$ from the above formula, and thus plugging $\alpha = \sqrt{1 - \beta^2}$, we have
$$\frac{\beta^2}{2} + \frac{\beta^4}{4} \geq \zeta_c(\sqrt{1 - \beta^2}, \beta) \geq \frac{\beta^2}{2} - \beta^3$$
- For the Lipschitz continuity, we have
$$\zeta_c(\sqrt{1 - x}, \sqrt{x}) = \frac{-2\pi\left(\sqrt{1 - x} - 1\right) + \left(\sqrt{1 - x} - 2\right)\sqrt{x} + \left(2\sqrt{1 - x} - 1\right)\arcsin\left(\sqrt{x}\right)}{2\pi}$$
Taking derivative with respect to $x$, we have
$$\frac{d}{d\beta}\zeta_c(\sqrt{1 - x}, \sqrt{x}) = -\frac{\sqrt{x} + \arcsin\left(\sqrt{x}\right) - \pi}{2\pi\sqrt{1 - x}}$$
It is easy to verify that for all $\beta \in [0, 0.9]$:
$$0.3 \leq \frac{d}{d\beta}\zeta_c(\sqrt{1 - x}, x) \leq 0.5$$
This proves the 0.5-Lipschitz continuity over $x$.

□

**Fact D.3.** *If* $x \in \mathbb{R}^m$ *is a vector with at most* $sm$ *coordinates satisfying* $|x_k| \leq \frac{s}{q}$, *then*

$$|\{k \in [m] \,|\, \mathbb{1}_{x_k \geq 0} \neq \mathbb{1}_{y_k \geq 0}\}| \leq sm + \frac{\|x - y\|^2}{s^2/q^2} \quad .$$

*If* $s$ *can be optimally chosen to minimize the above right hand side,*

$$|\{k \in [m] \,|\, \mathbb{1}_{x_k \geq 0} \neq \mathbb{1}_{y_k \geq 0}\}| \leq O\left(m^{2/3}(q\|x - y\|)^{2/3}\right) \quad .$$

# E Stability: After Re-Randomization

In this section we study a scenario where we re-randomize a fixed set of rows in the random initialization matrices $W$ and $A$. Formally, consider a fixed set $\mathcal{N} \subseteq [m]$ with cardinality $N = |\mathcal{N}|$. Define

- $\widetilde{W} \in \mathbb{R}^{m\times m}$ where $\widetilde{w}_k = w_k$ for $k \in [m] \setminus \mathcal{N}$ but $\widetilde{w}_k \sim \mathcal{N}(0, \frac{2\mathbf{I}}{m})$ is i.i.d. for $k \in \mathcal{N}$;
- $\widetilde{A} \in \mathbb{R}^{m\times(d_x+1)}$ where $\widetilde{a}_k = a_k$ for $k \in [m] \setminus \mathcal{N}$ but $\widetilde{a}_k \sim \mathcal{N}(0, \frac{2\mathbf{I}}{m})$ is i.i.d. for $k \in \mathcal{N}$.

For any fixed normalized input sequence $x_1, \dots, x_L \in \mathbb{R}^{d_x+1}$, we consider the following two executions of RNNs under $W$ and $\widetilde{W}$ respectively:

$$\begin{aligned}
&g_0 = h_0 = 0 && g'_0 = h'_0 = 0 && \\
&g_\ell = W h_{\ell-1} + A x_\ell && g_\ell + g'_\ell = \widetilde{W}(h_{\ell-1} + h'_{\ell-1}) + \widetilde{A} x_\ell && \text{for } \ell \in [L] \\
&h_\ell = \sigma(W h_{\ell-1} + A x_\ell) && h_\ell + h'_\ell = \sigma(\widetilde{W}(h_{\ell-1} + h'_{\ell-1}) + \widetilde{A} x_\ell) && \text{for } \ell \in [L]
\end{aligned}$$

We also denote

$$W' = \widetilde{W} - W \quad \text{and} \quad A' = \widetilde{A} - A$$

and define diagonal sign matrices $D_\ell \in \{0,1\}^{m\times m}$ and $D_\ell + D'_\ell \in \{0,1\}^{m\times m}$ by letting

$$(D_\ell)_{k,k} = \mathbb{1}_{(g_\ell)_k \ge 0} \text{ and } (D_\ell + D'_\ell)_{k,k} = \mathbb{1}_{(g_\ell + g'_\ell)_k \ge 0}.$$

Accordingly,

$$\begin{aligned}
\mathsf{Back}_{\ell\to a} &= B D_a W \cdots D_{\ell+1} W \in \mathbb{R}^{d\times m} \\
\mathsf{Back}_{\ell\to a} + \mathsf{Back}'_{\ell\to a} &= B(D_a + D'_a)(W + W') \cdots (D_{\ell+1} + D'_{\ell+1})(W + W') \in \mathbb{R}^{d\times m}
\end{aligned}$$

**Lemma E.1** (stability after re-randomization)**.** *Let $W, A, B$ be at random initialization. Fix $\mathcal{N} \subseteq [m]$ with $N = |\mathcal{N}| \le m/\rho^{23}$ and suppose we replace $W, A$ with $\widetilde{W}, \widetilde{A}$ by re-generating their rows in $\mathcal{N}$. Fix any normalized input sequence $x_1, \dots, x_L$. Let $h_\ell, g_\ell, D_\ell, \mathsf{Back}_{i\to j}$ be defined with respect to $W, A, x$ and $h_\ell + h'_\ell, g_\ell + g'_\ell, D_\ell + D'_\ell, \mathsf{Back}_{i\to j} + \mathsf{Back}'_{i\to j}$ be defined with respect to $\widetilde{W}, \widetilde{A}, x$. We have, with probability at least $1 - e^{-\Omega(\rho^2)}$ over the randomness of $W, \widetilde{W}, A, \widetilde{A}, B$*

1. $\|g'_\ell\|, \|h'_\ell\| \le O(\rho^5 \sqrt{N/m})$ *for every $\ell \in [L]$*
2. $\|D'_\ell\|_0 \le O(\rho^4 N^{1/3} m^{2/3})$ *for every $\ell \in [L]$*
3. $|\langle w_k, h'_\ell\rangle| \le O(\rho^5 N^{2/3} m^{-2/3})$ *for every $k \in [m], \ell \in [L]$*
4. $\left|u^\top \mathsf{Back}'_{i\to j} \mathbf{e}_k\right| \le O(\rho^7 (N/m)^{1/6}) \cdot \|u\|$ *for every $k \in [m]$, $u \in \mathbb{R}^d$ and $1 \le i \le j \le L$.*

## E.1 Proof for Lemma E.1

*Proof of Lemma E.1.* Throughout the proof we will fix some parameters $s$, $\tau_0$, $\tau_1$, $\tau_{\ell_2}$ but specify their values towards the end. (We will make sure $\tau_1 \ge \tau_0$ and $m \ge s\rho^4$.)

**Bound $\|g'_\ell\|$.** We inductively prove that the difference $g'_\ell$ can be written as the following form,

$$g'_\ell = \beta_0 + W\beta_1 + (W D_{\ell-1} W)\beta_2 + (W D_{\ell-1} W D_{\ell-2} W)\beta_3 + \cdots + (W D_{\ell-1} \cdots W D_2 W)\,\beta_{\ell-1}$$

with

$$\|\beta_i\|_0 \le \begin{cases} N, & i = 0; \\ N + s, & i \ge 1. \end{cases} \quad \text{and} \quad \|\beta_i\| \le \begin{cases} \tau_0, & i = 0; \\ \tau_1, & i \ge 1. \end{cases}$$

Under these assumptions:

- $\|g'_\ell\|_2 \le \tau_{\ell_2} \stackrel{\text{def}}{=} \Theta(\rho^4 \tau_1)$ using Lemma B.1i.
- $\|D'_\ell\|_0 \le s \stackrel{\text{def}}{=} \Theta(\rho^3 \tau_1^{2/3} m)$.

  This because combining Lemma B.1d and Fact D.3, we can derive that if $g_\ell$ is affected in Euclidean norm by $\tau$, then with probability at least $1 - e^{-\Omega(\rho^2)}$, the number of sign changes is at most $O(\tau^{2/3} m + \rho^2)$. In other words, $\|D'_\ell\|_0 \le O((\tau_{\ell_2})^{2/3} m) \le O(\rho^3 \tau_1^{2/3} m)$.

To prove the induction for $\ell+1$, we write down the recursive formula

$$g'_{\ell+1} = (W+W')(D_\ell + D'_\ell)(g_\ell + g'_\ell) - W D_\ell g_\ell + A' x_{\ell+1} \ .$$

There are many terms in this difference, and we treat them separately below.

- $A' x_{\ell+1}$. We have $\|A' x_{\ell+1}\| \leq \sqrt{N} \max_{k \in \mathcal{N}} |\langle \widetilde{a}_k - a_k, x_{\ell+1}\rangle| \leq O(\frac{\rho\sqrt{N}}{\sqrt{m}})$.

  Indeed, for each $k$, we have $|\langle \widetilde{a}_k, x_{\ell+1}\rangle| \leq O(\frac{\rho}{\sqrt{m}})$ and $|\langle a_k, x_{\ell+1}\rangle| \leq O(\frac{\rho}{\sqrt{m}})$ both with probability at least $1 - e^{-\Omega(\rho^2)}$.

- $W' D_\ell g_\ell = W' h_\ell$. We have $\|W' h_\ell\| \leq \sqrt{N} \max_{k \in \mathcal{N}} |\langle \widetilde{w}_k - w_k, h_\ell \rangle| \leq O(\frac{\rho\sqrt{N}}{\sqrt{m}})$.

  Indeed, each $|\langle \widetilde{w}_k, h_\ell\rangle| \leq O(\frac{\rho}{\sqrt{m}}) \cdot \|h_\ell\|$ with probability at least $1 - e^{-\Omega(\rho^2)}$, and we have $|\langle w_k, h_\ell\rangle| \leq O(\frac{\rho}{\sqrt{m}})$ from Lemma B.1c.

- $W D'_\ell (g_\ell + g'_\ell)$. We have $\|D'_\ell(g_\ell + g'_\ell)\| \leq 2\|D'_\ell g'_\ell\| \leq 2\tau_0 + O(\frac{\sqrt{s}\rho^2\tau_1}{\sqrt{m}})$. Indeed,

  - We have $\|D'_\ell(g_\ell + g'_\ell)\| \leq 2\|D'_\ell g'_\ell\|$. For each coordinate $k \in [m]$ to have sign change $\left(D'_\ell\right)_{k,k} \neq 0$, it must be because the signs of $[g_\ell]_k$ and $[g_\ell + g'_\ell]_k$ are opposite. This implies $|[g_\ell]_k| \leq |[g'_\ell]_k|$ and therefore $\|D'_\ell(g_\ell + g'_\ell)\| \leq 2\|D'_\ell g'_\ell\|$.
  - We have $\|D'_\ell g'_\ell\| \leq \tau_0 + O(\frac{\sqrt{s}\rho^2\tau_1}{\sqrt{m}})$. Recall from $|u^\top W D_\ell W \cdots D_a W v| \leq O\big(\frac{\sqrt{s}\rho}{\sqrt{m}}\big)\|u\|\|v\|$ for all $s$-sparse vectors $u$ and $v$ from Lemma B.1j. Therefore, $\|D'_\ell W D_{\ell-1} W \cdots W \beta_i\| \leq O(\frac{\sqrt{s}\rho\tau_1}{\sqrt{m}})$ for $i \geq 1$. Also, we have $\|D'_\ell \beta_0\| \leq \tau_0$.

- $W' D_\ell g'_\ell = (\widetilde{W}_{\mathcal{N}} - W_{\mathcal{N}}) D_\ell g'_\ell$ satisfies $\|W' D_\ell g'_\ell\| \leq O(\frac{\sqrt{s}\rho^2}{\sqrt{m}}\tau_1)$. Indeed,

  - $W_{\mathcal{N}} D_\ell g'_\ell = D_{\mathcal{N}} W D_\ell g'_\ell$ where $D_{\mathcal{N}}$ is the diagonal matrix satisfying $(D_{\mathcal{N}})_{k,k} = \mathbb{1}_{k \in [\mathcal{N}]}$. It satisfies $\|D_{\mathcal{N}} W D_\ell g'_\ell\| \leq O(\frac{\sqrt{s}\rho^2}{\sqrt{m}}\tau_1)$ using again Lemma B.1j and $N \leq s$.
  - $\widetilde{W}_{\mathcal{N}} D_\ell g'_\ell$ also satisfies $\|\widetilde{W}_{\mathcal{N}} D_\ell g'_\ell\| \leq O(\frac{\sqrt{s}\rho^2}{\sqrt{m}}\tau_1)$ for similar reasons.

- $W' D'_\ell (g_\ell + g'_\ell)$ satisfies $\|W' D'_\ell (g_\ell + g'_\ell)\| \leq O(\frac{\sqrt{s}\rho\tau_1}{\sqrt{m}})$.

  Indeed, $\|W' D'_\ell(g_\ell + g'_\ell)\| \leq \|W' D'_\ell\|_2 \cdot \|D'_\ell(g_\ell + g'_\ell)\| \leq O(\frac{\sqrt{s}\rho}{\sqrt{m}}) \cdot \big(2\tau_0 + O(\frac{\sqrt{s}\rho^2\tau_1}{\sqrt{m}})\big) \leq O(\frac{\sqrt{s}\rho\tau_1}{\sqrt{m}})$.
- $W D_\ell g'_\ell = W D_\ell \beta_0 + W D_\ell W \beta_1 + \cdots$ by induction

In sum, we have

$$g'_{\ell+1} = \underbrace{(W' D_\ell g_\ell + W' D_\ell g'_\ell + W' D'_\ell (g_\ell + g'_\ell) + A' x_{\ell+1})}_{\beta'_0} + W \underbrace{(D'_\ell(g_\ell + g'_\ell) + D_\ell \beta_0)}_{\beta'_1} + W D_\ell W \underbrace{\beta_1}_{\beta'_2} + \cdots$$

We have that $\|\beta'_0\|_0 \leq N$, $\|\beta'_0\| \leq O(\frac{\rho\sqrt{N}}{\sqrt{m}} + \frac{\rho^2\sqrt{s}}{\sqrt{m}}\tau_1)$, $\|\beta'_1\|_0 \leq s + N$, and $\|\beta'_1\|_2 \leq 3\tau_0 + O(\frac{\rho^2\sqrt{s}}{\sqrt{m}}\tau_1)$.

Finally, we choose $\tau_0 = \Theta(\frac{\rho\sqrt{N}}{\sqrt{m}} + \frac{\rho^2\sqrt{s}}{\sqrt{m}}\tau_1)$ and $\tau_1 = \Theta(\frac{\rho\sqrt{N}}{\sqrt{m}})$ to satisfy (using $N \leq m/\rho^{23}$)

$$\tau_1 \geq \Omega(\tau_0 + \frac{\rho^2\sqrt{s}}{\sqrt{m}}\tau_1) \quad \text{and} \quad m \geq s\rho^4$$

Since these parameters imply $\tau_{\ell_2} = \Theta(\rho^4\tau_1) = \Theta(\rho^5\sqrt{N}/\sqrt{m})$ we finish the proof that $\|g'_\ell\| \leq O(\rho^5\sqrt{N/m})$.

**Bound $\|h'_\ell\|$.** Since $h_\ell = D_\ell g_\ell$ we have $h'_\ell = D'_\ell(g_\ell + g'_\ell) + D_\ell g'_\ell$. We already have Euclidean norm bound on $\|g'_\ell\|$ and on $D'_\ell(g_\ell + g'_\ell)$ from the previous proof.

**Bound $\|D'_\ell\|_0$.** Since $s = \Theta(\rho^3 \tau_1^{2/3} m) \le O(\rho^4 N^{1/3} m^{2/3})$ we have the desired bound.

**Bound $\langle w_k, h'_\ell \rangle$.** We have $\langle w_k, h'_\ell \rangle = \mathbf{e}_k^\top W D'_\ell(g_\ell + g'_\ell) + \mathbf{e}_k^\top W D_\ell g'_\ell$. Using similar proof as before:

- $\|\mathbf{e}_k^\top W D'_\ell(g_\ell + g'_\ell)\| \le \|\mathbf{e}_k^\top W D'_\ell\|_2 \cdot \|D'_\ell(g_\ell + g'_\ell)\| \le O(\frac{\sqrt{s}\rho\tau_1}{\sqrt{m}})$.
- $\|\mathbf{e}_k^\top W D_\ell g'_\ell\| \le O(\frac{\sqrt{s}\rho^2\tau_1}{\sqrt{m}})$

**Bound backward.** Ignoring indices in $\ell$, we can write

$$u^\top \mathsf{Back}'_{i\to j}\, \mathbf{e}_k = u^\top B(D+D')(W+W')\cdots(D+D')(W+W')\mathbf{e}_k - u^\top BDW\cdots DW\mathbf{e}_k \ .$$

Let us compare the difference

$$(D+D')(W+W')\cdots(D+D')(W+W')\mathbf{e}_k - DW\cdots DW\mathbf{e}_k$$

and compute its Euclidean norm. One can in fact expand out all the (exponentially many) difference terms and bound them separately.

(1) If $D'$ shows up once and $W'$ never shows up, then the term is

$$(DW)^a D'(WD)^b W\mathbf{e}_k \ .$$

We have $\|(DW)^a\|_2 \le O(L^3)$ by Lemma B.1i and $\|D'(WD)^b W\mathbf{e}_k\| \le O(\rho\sqrt{s/m})$ by Lemma B.1j. Therefore, its absolute value is at most $O(\rho^4\sqrt{s/m})$, and there are at most $L$ such terms.

(2) If $W' = D_{\mathcal{N}}(\widetilde{W} - W)$ shows up once and $D'$ never shows up, then the term is

$$(DW)^a DW'(DW)^b \mathbf{e}_k \ .$$

We have $\|(DW)^a D\| \le O(L^3)$ by Lemma B.1i and $\|D_{\mathcal{N}} W'(DW)^b\mathbf{e}_k\| \le O(\rho\sqrt{s/m})$ by Lemma B.1j. (We also have $\|D_{\mathcal{N}}\widetilde{W}(DW)^b\mathbf{e}_k\| \le O(\rho\sqrt{s/m})$ but this is much easier to prove because $\widetilde{W}$ is fresh new random.) Therefore, its absolute value is at most $O(\rho^4\sqrt{s/m})$, and there are at most $L$ such terms.

(3) If the total number of times $D'$ and $W'$ show up is 2, then the occurance of $D'$ and $W'$ divides the difference term into three consecutive parts. As before, the norm of the first and the last parts are at most $O(L^3)$ and $O(\rho\sqrt{s/m})$ respectively, so it suffices to bound the *matrix spectral norm* of the middle part. There are four possibilities for this middle part:

- $D'WD\cdots WD'$.
- $D'WD\cdots WD_{\mathcal{N}}$.
- $D_{\mathcal{N}}(\widetilde{W} - W)DW\cdots WD'$.
- $D_{\mathcal{N}}(\widetilde{W} - W)DW\cdots WD_{\mathcal{N}}$.

All of such matrices have spectral norm at most $O(\rho\sqrt{s/m}) \le \frac{1}{100L}$ by Lemma B.1j because $D'$ and $D_{\mathcal{N}}$ are both $s$-sparse. Therefore, although there are at most $(2L)^2$ such difference terms, each of them is at most $\frac{1}{100L}\cdot O(\rho^4\sqrt{s/m})$ in magnitude. Therefore, their total contribution is negligible when comparing to cases (1) and (2).

(4) If the total number of times $D'$ and $W'$ show up is 3, then there are at most $(2L)^3$ such terms and each of them is at most $\frac{1}{(100L)^2}\cdot O(\rho^4\sqrt{s/m})$ in magnitude.

(5) And so on.

In sum, we conclude that

$$\big\|(D+D')(W+W')\cdots(D+D')(W+W')\mathbf{e}_k - DW\cdots DW\mathbf{e}_k\big\| \le L\cdot O(\rho^4\sqrt{s/m}) \le O(\rho^5\sqrt{s/m})$$

and finally using the randomness of $B$ we have with probability at least $1-e^{-\Omega(\rho^2)}$

$$|u^\top \mathsf{Back}'_{i\to j}\,\mathbf{e}_k| \le O(\rho)\cdot O(\rho^5\sqrt{s/m}) \enspace .$$

□

# F Stability: After Adversarial Perturbation

In this section we study a scenario where the random initialization matrix $W$ is perturbed to $W+W'$ with $W'$ being small in spectral norm. Intuitively, this $W'$ will later capture how much SGD has moved away from the random initialization, so it may depend on the randomness of $W, A, B$. To untangle this possibly complicated correlation, we consider stability with respect to *all* $W'$ being small. The following lemma has appeared already in the "Stability After Adversarial Perturbation" section of [2].

**Lemma F.1.** *Let $W, A, B$ be at random initialization, $x$ be a fixed normalized input sequence, and $\Delta\in[\varrho^{-100},\varrho^{100}]$. With probability at least $1-e^{-\Omega(\rho)}$ over the randomness of $W, A, B$, the following holds. Given any perturbation matrix $W'$ with $\|W'\|_2\le\frac{\Delta}{\sqrt{m}}$, letting*

- *$h_\ell, D_\ell, \mathsf{Back}_{i\to j}$ be defined with respect to $W, A, B, x$, and*
- *$h_\ell+h'_\ell, D_\ell+D'_\ell, \mathsf{Back}_{i\to j}+\mathsf{Back}'_{i\to j}$ be defined with respect to $W+W', A, B, x$*

*then*

*(a)* $\|h'_i\|\le O(\rho^6\Delta/\sqrt{m})$ *for every* $i\in[L]$ *(forward stability)*

*(b)* $\|D'_i\|_0\le O(\rho^4\Delta^{2/3}m^{2/3})$ *for every* $i\in[L]$ *(sign change)*

*(c)* $\|\mathsf{Back}'_{i\to j}\|_2\le O\big(\Delta^{1/3}\rho^6 m^{1/3}\big)$ *for every* $1\le i\le j\le L$ *(backward stability)*

Specifically, Lemma F.1a and F.1b can be found in Lemma C.2 and Lemma F.1c can be found in Lemma C.9 of [2, ver.3].

# G Proof for Section 5

## G.1 Missing Proof of Lemma 5.1

*Proof of Lemma 5.1.* Recall the "indicator to function" lemma of [1, Lemma 6.2 of ver.5] which says the following.[9] For any smooth function $\Psi$, there exists two-dimensional function $H^\Psi\colon\mathbb{R}^2\to[-\mathfrak{C}_{\varepsilon_e}(\Psi,1),\mathfrak{C}_{\varepsilon_e}(\Psi,1)]$ such that, for every $v^*, y\in\mathbb{R}^d$ with $\|v^*\|_2=\|y\|_2=1$:

$$\left|\mathop{\mathbb{E}}_{v\sim\mathcal{N}(0,\mathbf{I}),b_0\sim\mathcal{N}(0,1)}\left[\mathbb{1}_{\langle v,y\rangle+b_0\ge 0}H^\Psi(\langle v,v^*\rangle,b_0)\right]-\Psi(\langle v^*,y\rangle)\right|\le\varepsilon_e \enspace . \tag{G.1}$$

We now revise (G.1) in two ways without changing much of its original proof.

- First, the above $b_0\sim\mathcal{N}(0,1)$ is quite an arbitrary choice in their proof, and can be replaced with any other $b_0\sim\mathcal{N}(0,\tau^2)$ for constant $\tau\in(0,1]$. They have constructed $H^\Psi$ by first expanding $\Psi$ into its Taylor expansions, and then approximating each term $x^i$ with $h_i(z)$— probabilists' Hermite polynomial of degree $i$— with $z = \langle v, v^* \rangle$. Only the coefficient in front of each $h_i(z)$, namely $c_i'$ in their Eq. (117) depends on the choice of $\tau$, and $c_i'$ decreases from its original value as $\tau$ decreases from 1. Therefore, their final construction of $H^\Psi$ will only have a smaller magnitude in these coefficients so (G.1) remains unchanged if we choose $\tau = \frac{1}{\sqrt{3+4\sigma^2}}$.

[9] For concreteness, here is a link to their arXiv version 5: `https://arxiv.org/pdf/1811.04918v5.pdf`.

- Second, although not explicitly stated, it satisfies $\left|\mathbb{E}[H^\Psi(\langle v, v^* \rangle, b_0)] - \Psi(0)\right| \leq \varepsilon_e/4$. See for instance [1, page 42-43 of ver.5], their $H^\Psi$ is a linear combination of Hermite polynomials $h_i(z)$ with $z = \langle v, v^* \rangle$ which is a standard Gaussian. Since $\mathbb{E}_{g \sim \mathcal{N}(0,1)}[h_i(g)] = 0$ for all $i \geq 1$, this is zero for all terms except the degree-0 constant that is $2\Psi(0)$. The reason for the error $\varepsilon_e$ term is because they have additionally truncated each $h_i(z)$ when $|z|$ is very large. This truncation (see their Claim B.2) creates at most $\varepsilon_e/4$ error.

Taking into account the above two observations, we can restate [1, Lemma 6.2 of ver.5] as follows. For any smooth function $\Psi$, there exists $H^\Psi \colon \mathbb{R}^2 \to [-\mathfrak{C}_{\varepsilon_e}(\Psi, 1), \mathfrak{C}_{\varepsilon_e}(\Psi, 1)]$ such that, for every $v^*, y \in \mathbb{R}^d$ with $\|v^*\|_2 = \|y\|_2 = 1$:

$$\left| \mathop{\mathbb{E}}_{v \sim \mathcal{N}(0, \mathbf{I}), b_0 \sim \mathcal{N}(0, \frac{1}{3+4\sigma^2})} \left[ \mathbb{1}_{\langle v, y \rangle + b_0 \geq 0} H^\Psi(\langle v, v^* \rangle, b_0) \right] - \Psi(\langle v^*, y \rangle) \right| \leq \varepsilon_e \quad . \tag{G.2}$$

$$\left| \mathop{\mathbb{E}}_{v \sim \mathcal{N}(0, \mathbf{I}), b_0 \sim \mathcal{N}(0, \frac{1}{3+4\sigma^2})} \left[ H^\Psi(\langle v, v^* \rangle, b_0) \right] - 2\Psi(0) \right| \leq \frac{\varepsilon_e}{4} \quad . \tag{G.3}$$

**Parameter Choices.** Having restated (G.2) and (G.3) from the prior work, let us choose parameters to apply them. Let us separate out the last coordinate for these vectors. Suppose

$$a = (a_\triangleleft, a_\triangleright) \quad , \quad x^\star = (x^\star_\triangleleft, \frac{1}{2}) \quad , \quad w^* = (w^*_\triangleleft, 0)$$

where $a_\triangleleft, x^\star_\triangleleft, w^*_\triangleleft \in \mathbb{R}^{d_x - 1}$ and $a_\triangleright \in \mathbb{R}$. Recall $\|x^\star_\triangleleft\| = \sqrt{1 - (1/2)^2} = \sqrt{3/4}$ and $\|w^*_\triangleleft\| = 1$. Let us choose

$$v = (a_\triangleleft, \frac{n}{\sigma}) \quad , \quad y = \frac{(x^\star_\triangleleft, \sigma)}{\sqrt{\sigma^2 + 3/4}} \quad , \quad b_0 = \frac{a_\triangleright}{\sqrt{3 + 4\sigma^2}} \quad , \quad v^* = (w^*_\triangleleft, 0) \quad , \quad \Psi(z) = \Phi(\frac{z}{\sqrt{\sigma^2 + 3/4}})$$

One can carefully verify the following:

- $v \sim \mathcal{N}(0, \mathbf{I})$, $\|y\| = 1$, $b_0 \sim \mathcal{N}(0, \frac{1}{3+4\sigma^2})$
- $\langle v, y \rangle + b_0 = \frac{1}{\sqrt{\sigma^2 + 3/4}}\big(\langle a_\triangleleft, x^\star_\triangleleft \rangle + n + \frac{a_\triangleright}{2}\big) = \frac{1}{\sqrt{\sigma^2 + 3/4}}\big(\langle a, x^\star \rangle + n\big)$
- $\Phi(\langle v^*, y \rangle) = \Psi(\sqrt{\sigma^2 + 3/4}\langle v^*, y \rangle) = \Phi(\langle w^*_\triangleleft, x^\star_\triangleleft \rangle) = \Phi(\langle w^*, x^\star \rangle)$
- $\langle v, v^* \rangle = \langle w^*_\triangleleft, a_\triangleleft \rangle$

Therefore, choosing $H(a) \stackrel{\text{def}}{=} H^\Psi(\langle w^*_\triangleleft, a_\triangleleft \rangle, \frac{a_\triangleright}{\sqrt{3+4\sigma^2}}) = H^\Psi(\langle v, v^* \rangle, b_0)$, plugging these choices into (G.2) and (G.3), we have

$$\left| \mathop{\mathbb{E}}_{a \sim \mathcal{N}(0, \mathbf{I}), n \sim \mathcal{N}(0, \sigma^2)} \left[ \mathbb{1}_{\langle a, x^\star \rangle + n \geq 0} H(a) \right] - \Phi(\langle w^*, x^\star \rangle) \right| \leq \varepsilon_e$$

$$\left| \mathop{\mathbb{E}}_{a \sim \mathcal{N}(0, \mathbf{I}), n \sim \mathcal{N}(0, \sigma^2)} \left[ H(a) \right] - 2\Phi(0) \right| \leq \frac{\varepsilon_e}{4}$$

The first statement above finishes the proof of Lemma 5.1a. We note that $\mathfrak{C}_{\varepsilon_e}(\Psi, 1) = \mathfrak{C}_{\varepsilon_e}(\Phi, \sigma)$ because $\Phi$ is re-scaled from $\Psi$ by $\sqrt{\sigma^2 + 3/4} \leq O(\sigma)$.

**Off Target.** For Lemma 5.1b, we can derive the following

$$\begin{aligned}
\mathbb{E}_{a,n}\left[\mathbb{1}_{\langle a,x^\star\rangle+\gamma n\geq 0} H(a)\right] &\overset{①}{=} \mathbb{E}_{a,n}\left[\mathbb{1}_{\langle a,x^\star\rangle+\gamma n\geq 0} H(a)\,\Big|\, |\langle a,x^\star\rangle|\leq\sqrt{\log(\gamma\sigma)}\right]\pm O\big(\frac{C'}{\gamma\sigma}\big)\\
&= \mathbb{E}_{a}\left[\mathbf{Pr}\left[\langle a,x^\star\rangle+\gamma n\geq 0\right] H(a)\,\Big|\, |\langle a,x^\star\rangle|\leq\sqrt{\log(\gamma\sigma)}\right]\pm O\big(\frac{C'}{\gamma\sigma}\big)\\
&= \mathbb{E}_{a}\left[\Big(\frac{1}{2}\pm O\big(\frac{\sqrt{\log(\gamma\sigma)}}{\gamma\sigma}\big)\Big) H(a)\,\Big|\, |\langle a,x^\star\rangle|\leq\sqrt{\log(\gamma\sigma)}\right]\pm O\big(\frac{C'}{\gamma\sigma}\big)\\
&\overset{②}{=} \frac{1}{2}\mathbb{E}_{a}\left[H(a)\,\Big|\, |\langle a,x^\star\rangle|\leq\sqrt{\log(\gamma\sigma)}\right]\pm O\big(\frac{C'\sqrt{\log(\gamma\sigma)}}{\gamma\sigma}\big)\\
&= \frac{1}{2}\mathbb{E}_{a}[H(a)]\pm O\big(\frac{C'\sqrt{\log(\gamma\sigma)}}{\gamma\sigma}\big)\\
&= \Phi(0)\pm\big(\frac{\varepsilon_e}{4}+O\big(\frac{C'\sqrt{\log(\gamma\sigma)}}{\gamma\sigma}\big)\big)\ .
\end{aligned}$$

Above, ① is because $|\langle a,x^\star\rangle|>\sqrt{\log(\gamma\sigma)}$ with probability at most $O\big(\frac{1}{\gamma\sigma}\big)$ and $|H|\leq C'$; ② uses $|H|\leq C'$. □

## G.2 Missing Proof of Lemma 5.2

*Proof of Lemma 5.2.* Let us denote

$$\widehat{h}=h^{(0)}_{i-1}/\|h^{(0)}_{i-1}\|\ ,$$

and we have

$$\big\|(I-\widehat{h}\widehat{h}^\top)h_{i-1}\big\|^2=\|h_{i-1}\|^2-\frac{\langle h^{(0)}_{i-1},h_{i-1}\rangle^2}{\|h^{(0)}_{i-1}\|^2}=\|h_{i-1}\|^2-\frac{\big(\|h^{(0)}_{i-1}\|^2+\|h_{i-1}\|^2-\|h^{(0)}_{i-1}-h_{i-1}\|^2\big)^2}{4\|h^{(0)}_{i-1}\|^2}\ .$$

Using Lemma B.1a and Lemma D.1a, we can write (abbreviating by $\zeta_n=\zeta_n(\varepsilon_x,i-1)$ and $\zeta_d=\zeta_d(\varepsilon_x,i-1)$),

$$\begin{aligned}
\big\|(I-\widehat{h}\widehat{h}^\top)h_{i-1}\big\|^2 &= \big(\zeta_n^2\pm O(\frac{\rho^2}{\sqrt{m}})\big)-\frac{\big(2\big(\zeta_n^2\pm O(\frac{\rho^2}{\sqrt{m}})\big)-\big(\zeta_d^2\pm O(\frac{\rho}{\sqrt{m}})\big)\big)^2}{4\big(\zeta_n^2\pm O(\frac{\rho^2}{\sqrt{m}})\big)}\\
&= \zeta_n^2-\frac{\big(2\zeta_n^2-\zeta_d^2\pm O(\frac{\rho^2}{\sqrt{m}})\big)^2}{4\zeta_n^2\pm O(\frac{\rho^2}{\sqrt{m}})}\pm O(\frac{\rho^2}{\sqrt{m}})\\
&= \zeta_n^2-\frac{\big(2\zeta_n^2-\zeta_d^2\big)^2\pm O(\frac{\rho^2}{\sqrt{m}})}{4\zeta_n^2\pm O(\frac{\rho^2}{\sqrt{m}})}\pm O(\frac{\rho^2}{\sqrt{m}})\\
&= \zeta_d^2-\frac{\zeta_d^4}{4\zeta_n^2}\pm O(\frac{\rho^2}{\sqrt{m}})=\tau\pm O(\frac{\rho^2}{\sqrt{m}}) \qquad\text{(G.4)}
\end{aligned}$$

where we have defined

$$\tau\overset{\text{def}}{=}\sqrt{\zeta_d(\varepsilon_x,i-1)^2-\frac{\zeta_d(\varepsilon_x,i-1)^4}{4\zeta_n(\varepsilon_x,i-1)^2}}\ .$$

Note that by Lemma B.1a and Lemma D.1a we have

$$\tau=[0.9\sqrt{i-1}\varepsilon_x,1.1\sqrt{i-1}\varepsilon_x]\ .$$

Let us apply Lemma 5.1 with

$$\sigma = \frac{\tau}{\varepsilon_x} \le 1.1\sqrt{L} \quad , \quad \Phi = \Phi_{i\to j,r,s}$$

to get the $H$ with $|H| \le C'$, and let us define

$$H_{i\to j,r,s} = \frac{1}{\varepsilon_c'} H \text{ where } \varepsilon_c' = \mathbf{Pr}_{\widetilde{w}_k}\big[|\langle \widetilde{w}_k, h_{i-1}^{(0)}\rangle| \le \frac{\varepsilon_c}{\sqrt{m}}\big].$$

A standard property of Gaussian random variable shows that:

$$\varepsilon_c' = \Theta(\varepsilon_c)$$

**On Target.** Let us now focus on

$$\widetilde{g}_i \stackrel{\text{def}}{=} \widetilde{W} h_{i-1} + \widetilde{A} x_i \ .$$

We can decompose $h_{i-1}$ into the projection on $h_{i-1}^{(0)}$ and on the perpendicular space of $h_{i-1}^{(0)}$. We can write $(\widetilde{g}_i)_k$ as:

$$(\widetilde{g}_i)_k = \langle \widetilde{w}_k, h_{i-1}^{(0)}\rangle \cdot \frac{\langle h_{i-1}^{(0)}, h_{i-1}\rangle}{\|h_{i-1}^{(0)}\|^2} + \Big\langle \widetilde{w}_k, (I - \widehat{h}\widehat{h}^\top) h_{i-1}\Big\rangle + \langle \widetilde{a}_k, x_i\rangle$$

By the fact that $\|h_{i-1}^{(0)}\|, \|h_{i-1}\| \in [0.9, 2]$ from Lemma B.1a, we know that $\big|\frac{\langle h_{i-1}^{(0)}, h_{i-1}\rangle}{\|h_{i-1}^{(0)}\|^2}\big| \le 3$. As a result, conditioning on $|\langle \widetilde{w}_k, h_{i-1}^{(0)}\rangle| \le \frac{\varepsilon_c}{\sqrt{m}}$, we have

$$\Big|\langle \widetilde{w}_k, h_{i-1}\rangle - \Big\langle \widetilde{w}_k, (I - \widehat{h}\widehat{h}^\top) h_{i-1}\Big\rangle\Big| \le 2\frac{\varepsilon_c}{\sqrt{m}} \tag{G.5}$$

Since $\langle \widetilde{a}_k, x_i\rangle \sim \mathcal{N}\big(0, \frac{2\varepsilon_x^2}{m}\big)$ is independent of the randomness of $\widetilde{w}_k$, equation (G.5) implies

$$\begin{aligned}
&\mathbf{Pr}_{\widetilde{a}_k}\left[\mathbb{1}_{\langle \widetilde{w}_k, h_{i-1}\rangle + \langle \widetilde{a}_k, x_i\rangle \ge 0} \ne \mathbb{1}_{\langle \widetilde{w}_k, (I - \widehat{h}\widehat{h}^\top) h_{i-1}\rangle + \langle \widetilde{a}_k, x_i\rangle \ge 0} \;\middle|\; |\langle \widetilde{w}_k, h_{i-1}^{(0)}\rangle| \le \frac{\varepsilon_c}{\sqrt{m}}, \widetilde{w}_k\right] \\
&\stackrel{①}{\le} \mathbf{Pr}_{\widetilde{a}_k}\left[|\langle \widetilde{a}_k, x_i\rangle| \in \left[\xi, \xi + \Big|\langle \widetilde{w}_k, h_{i-1}\rangle - \Big\langle \widetilde{w}_k, (I - \widehat{h}\widehat{h}^\top) h_{i-1}\Big\rangle\Big|\right] \;\middle|\; |\langle \widetilde{w}_k, h_{i-1}^{(0)}\rangle| \le \frac{\varepsilon_c}{\sqrt{m}}, \widetilde{w}_k\right] \\
&\stackrel{②}{\le} \mathbf{Pr}_{\widetilde{a}_k}\left[|\langle \widetilde{a}_k, x_i\rangle| \in \left[\xi, \xi + \frac{2\varepsilon_c}{\sqrt{m}}\right]\right] \\
&\stackrel{③}{\le} \frac{2\varepsilon_c}{\varepsilon_x} \ .
\end{aligned} \tag{G.6}$$

Above, inequality ① is because $\mathbb{1}_{a+g\ge 0} \ne \mathbb{1}_{b+g\ge 0}$ implies there exists some $\xi \in \mathbb{R}$ such that $g \in [\xi, \xi + |b-a|]$; inequality ② is because of (G.5); and inequality ③ is because $\langle \widetilde{a}_k, x_i\rangle \sim \mathcal{N}(0, \frac{2\|x_i\|^2}{m})$

and $\|x_i\| \geq \varepsilon_x$. Using this, we can derive that

$$
\begin{aligned}
&\mathop{\mathbb{E}}_{\widetilde{w}_k,\widetilde{a}_k}\left[\mathbb{1}_{|\langle \widetilde{w}_k, h_{i-1}^{(0)}\rangle| \leq \frac{\varepsilon_c}{\sqrt{m}}}\mathbb{1}_{\langle \widetilde{w}_k, h_{i-1}\rangle + \langle \widetilde{a}_k, x_i\rangle \geq 0} H_{i\to j,r,s}(\widetilde{a}_k)\right] \\
&= \mathop{\mathbb{E}}_{\widetilde{w}_k,\widetilde{a}_k}\left[\mathbb{1}_{\langle \widetilde{w}_k, h_{i-1}\rangle + \langle \widetilde{a}_k, x_i\rangle \geq 0} H_{i\to j,r,s}(\widetilde{a}_k) \;\middle|\; |\langle \widetilde{w}_k, h_{i-1}^{(0)}\rangle| \leq \frac{\varepsilon_c}{\sqrt{m}}\right] \mathbf{Pr}\left[|\langle \widetilde{w}_k, h_{i-1}^{(0)}\rangle| \leq \frac{\varepsilon_c}{\sqrt{m}}\right] \\
&= \varepsilon_c' \mathop{\mathbb{E}}_{\widetilde{w}_k,\widetilde{a}_k}\left[\mathbb{1}_{\langle \widetilde{w}_k, h_{i-1}\rangle + \langle \widetilde{a}_k, x_i\rangle \geq 0} H_{i\to j,r,s}(\widetilde{a}_k) \;\middle|\; |\langle \widetilde{w}_k, h_{i-1}^{(0)}\rangle| \leq \frac{\varepsilon_c}{\sqrt{m}}\right] \\
&\overset{①}{=} \varepsilon_c' \mathop{\mathbb{E}}_{\widetilde{w}_k,\widetilde{a}_k}\left[\mathbb{1}_{\langle \widetilde{w}_k, (I-\widehat{h}\widehat{h}^\top)h_{i-1}\rangle + \langle \widetilde{a}_k, x_i\rangle \geq 0} H_{i\to j,r,s}(\widetilde{a}_k) \;\middle|\; |\langle \widetilde{w}_k, h_{i-1}^{(0)}\rangle| \leq \frac{\varepsilon_c}{\sqrt{m}}\right] \pm \frac{2\varepsilon_c C'}{\varepsilon_x} \\
&= \varepsilon_c' \mathop{\mathbb{E}}_{\widetilde{w}_k,\widetilde{a}_k}\left[\mathbb{1}_{\left\langle \widetilde{w}_k, (I-\widehat{h}\widehat{h}^\top)h_{i-1}\right\rangle + \langle \widetilde{a}_k, x_i\rangle \geq 0} H_{i\to j,r,s}(\widetilde{a}_k)\right] \pm \frac{2\varepsilon_c C'}{\varepsilon_x} \\
&= \mathop{\mathbb{E}}_{\widetilde{w}_k,\widetilde{a}_k}\left[\mathbb{1}_{\left\langle \widetilde{w}_k, (I-\widehat{h}\widehat{h}^\top)h_{i-1}\right\rangle + \langle \widetilde{a}_k, x_i\rangle \geq 0} H(\widetilde{a}_k)\right] \pm \frac{2\varepsilon_c C'}{\varepsilon_x} \ . \qquad \text{(G.7)}
\end{aligned}
$$

Above, ① uses (G.6) and $|H| \leq C'$.

Next, recall from (G.4) that $|\|(I - \widehat{h}\widehat{h}^\top)h_{i-1}\| - \tau| = O(\frac{\rho^2}{\sqrt{m}})$. As a result, with probability at least $1 - e^{-\Omega(\rho^2)}$ over $\widetilde{w}_k$,

$$
\begin{aligned}
\left|\left\langle \widetilde{w}_k, (I - \widehat{h}\widehat{h}^\top)h_{i-1}\right\rangle - \left\langle \widetilde{w}_k, \tau\frac{(I - \widehat{h}\widehat{h}^\top)h_{i-1}}{\|(I - \widehat{h}\widehat{h}^\top)h_{i-1}\|}\right\rangle\right| &\leq \left|\left\langle \widetilde{w}_k, \frac{(I - \widehat{h}\widehat{h}^\top)h_{i-1}}{\|(I - \widehat{h}\widehat{h}^\top)h_{i-1}\|}\right\rangle\right| \cdot O(\frac{\rho^2}{\sqrt{m}}) \\
&\overset{①}{\leq} O(\frac{\rho^3}{m}) \ . \qquad \text{(G.8)}
\end{aligned}
$$

Above, ① is because $\left\langle \widetilde{w}_k, \frac{(I-\widehat{h}\widehat{h}^\top)h_{i-1}}{\|(I-\widehat{h}\widehat{h}^\top)h_{i-1}\|}\right\rangle \sim \mathcal{N}\big(0, \frac{2}{m}\big)$, so with probability at least $1 - e^{-\Omega(\rho^2)}$ over $\widetilde{w}_k$ it is at most $O(\rho/\sqrt{m})$. Using (G.8), together with a similar argument to (G.6), we have

$$
\mathop{\mathbf{Pr}}_{\widetilde{a}_k}\left[\mathbb{1}_{\left\langle \widetilde{w}_k, (I-\widehat{h}\widehat{h}^\top)h_{i-1}\right\rangle + \langle \widetilde{a}_k, x_i\rangle} \neq \mathbb{1}_{\left\langle \widetilde{w}_k, \tau\frac{(I-\widehat{h}\widehat{h}^\top)h_{i-1}}{\|(I-\widehat{h}\widehat{h}^\top)h_{i-1}\|}\right\rangle + \langle \widetilde{a}_k, x_i\rangle} \;\middle|\; \widetilde{w}_k \text{ satisfies (G.8)}\right] = O(\frac{\rho^3}{\varepsilon_x\sqrt{m}}) \ . \qquad \text{(G.9)}
$$

Thus, defining $n_k \overset{\text{def}}{=} \tau\frac{(I-\widehat{h}\widehat{h}^\top)h_{i-1}}{\|(I-\widehat{h}\widehat{h}^\top)h_{i-1}\|}$, we have

$$
\begin{aligned}
\mathop{\mathbb{E}}_{\widetilde{w}_k,\widetilde{a}_k}\left[\mathbb{1}_{\left\langle \widetilde{w}_k, (I-\widehat{h}\widehat{h}^\top)h_{i-1}\right\rangle + \langle \widetilde{a}_k, x_i\rangle \geq 0} H(\widetilde{a}_k)\right] &\overset{①}{=} \mathop{\mathbb{E}}_{n_k,\widetilde{a}_k}\left[\mathbb{1}_{n_k + \langle \widetilde{a}_k, x_i\rangle \geq 0} H(\widetilde{a}_k)\right] \pm O(\frac{\rho^3 C'}{\varepsilon_x\sqrt{m}}) \\
&\overset{②}{=} \Phi_{i\to j,r,s}(\langle w^*_{i\to j,r,s}, x_i^\star\rangle) \pm \varepsilon_e \pm O(\frac{\rho^3 C'}{\varepsilon_x\sqrt{m}}) \ . \qquad \text{(G.10)}
\end{aligned}
$$

Above, ① is because of (G.9) and the definition of $n_k$; and ② is because of Lemma 5.1 with $x_i = (\varepsilon_x x_i^\star, 0)$ and $\sigma = \frac{\tau}{\varepsilon_x} = O(\sqrt{L})$ and re-scaling. Putting together (G.7) and (G.10) finish the proof for the on target part.

**Off Layer.** Let us next focus on

$$\widetilde{g}_{i'} \overset{\text{def}}{=} \widetilde{W}h_{i'-1} + \widetilde{A}x_{i'} \ .$$

We again decompose $h_{i'-1}$ into the projection on $h_{i-1}^{(0)}$ and on the perpendicular space of $h_{i-1}^{(0)}$. We can write $(\widetilde{g}_{i'})_k$ as:

$$(\widetilde{g}_{i'})_k = \langle \widetilde{w}_k, h_{i-1}^{(0)} \rangle \cdot \frac{\langle h_{i-1}^{(0)}, h_{i'-1} \rangle}{\|h_{i-1}^{(0)}\|^2} + \Big\langle \widetilde{w}_k, (I - \widehat{h}\widehat{h}^\top) h_{i'-1} \Big\rangle + \langle \widetilde{a}_k, x_{i'} \rangle \enspace .$$

By the fact that $\|h_{i-1}^{(0)}\|, \|h_{i'-1}\| \in [0.9, 2]$ from Lemma B.1a, we again know $\big|\frac{\langle h_{i-1}^{(0)}, h_{i'-1}\rangle}{\|h_{i-1}^{(0)}\|^2}\big| \leq 3$. Therefore, the same derivation of (G.7) implies

$$\begin{aligned} &\mathbb{E}_{\widetilde{w}_k, \widetilde{a}_k} \Big[ \mathbb{1}_{|\langle \widetilde{w}_k, h_{i-1}^{(0)} \rangle| \leq \frac{\varepsilon_c}{\sqrt{m}}} \mathbb{1}_{\langle \widetilde{w}_k, h_{i'-1}\rangle + \langle \widetilde{a}_k, x_{i'}\rangle \geq 0} H_{i\to j,r,s}(\widetilde{a}_k) \Big] \\ &= \mathbb{E}_{\widetilde{w}_k, \widetilde{a}_k} \Big[ \mathbb{1}_{\langle \widetilde{w}_k, (I - \widehat{h}\widehat{h}^\top) h_{i'-1} \rangle + \langle \widetilde{a}_k, x_{i'}\rangle \geq 0} H(\widetilde{a}_k) \Big] \pm \frac{2\varepsilon_c C'}{\varepsilon_x} \enspace . \end{aligned} \tag{G.11}$$

Using Lemma 5.1b, we have for every $\gamma > 1$,

$$\begin{aligned} &\mathbb{E}_{\widetilde{w}_k, \widetilde{a}_k} \Big[ \mathbb{1}_{\langle \widetilde{w}_k, (I - \widehat{h}\widehat{h}^\top) h_{i'-1} \rangle + \langle \widetilde{a}_k, x_{i'}\rangle \geq 0} H(\widetilde{a}_k) \,\Big|\, \|(I - \widehat{h}\widehat{h}^\top) h_{i'-1}\| = \gamma\sigma\varepsilon_x \Big] \\ &= \mathbb{E}_{n_k \sim \mathcal{N}(0, \frac{2\sigma^2\varepsilon_x^2}{m}), \langle \widetilde{a}_k, x_{i'}\rangle \sim \mathcal{N}(0, \frac{2\varepsilon_x^2}{m})} \big[ \mathbb{1}_{\gamma n_k + \langle \widetilde{a}_k, x_{i'}\rangle \geq 0} H(\widetilde{a}_k) \big] \\ &\overset{①}{=} 0 \pm \varepsilon_e \pm O\big(\frac{C' \log(\gamma\sigma)}{\gamma\sigma}\big) \end{aligned} \tag{G.12}$$

Above, ① is by applying Lemma 5.1b after re-scaling and $\Phi_{i\to j,r,s}(0) = 0$.

Using Lemma B.1k, we know for $i' > i$ with high probability $\|(I - \widehat{h}\widehat{h}^\top) h_{i'-1}^{(0)}\| \geq \Omega\big(\frac{1}{L^2 \log^3 m}\big)$. (One can argue similarly with the help of Lemma B.1 for $i' < i$.) Thus, by the closeness property $\|h_{i'-1} - h_{i'-1}^{(0)}\| \leq O(\sqrt{L}\varepsilon_x)$ from Lemma D.1a, we also have with probability at least $1 - e^{-\Omega(\rho^2)}$ over the randomness of $W, A$,

$$\|(I - \widehat{h}\widehat{h}^\top) h_{i'-1}\| \geq \Omega\big(\frac{1}{L^2 \log^3 m}\big) \enspace .$$

Putting this into (G.12), we have with probability at least $1 - e^{-\Omega(\rho^2)}$,

$$\mathbb{E}_{\widetilde{w}_k, \widetilde{a}_k} \Big[ \mathbb{1}_{\langle \widetilde{w}_k, (I - \widehat{h}\widehat{h}^\top) h_{i'-1} \rangle + \langle \widetilde{a}_k, x_{i'}\rangle \geq 0} H(\widetilde{a}_k) \Big] = \pm\varepsilon_e \pm O\big(C'\varepsilon_x \rho^3 \log(\varepsilon_x \rho)\big) = \pm 2\varepsilon_e \tag{G.13}$$

Combining this with (G.11), and using our choice of $\varepsilon_c$ finishes the proof of for the off target part. □

### G.3 Missing Proof of Lemma 5.4

*Proof of Lemma 5.4.* Let us consider a random subset $\mathcal{N} \subset [m]$ of size $|\mathcal{N}| = N$, and replace the rows $\{w_k, a_k\}_{k\in\mathcal{N}}$ of $W$ and $A$ with freshly new i.i.d. samples $\widetilde{w}_k, \widetilde{a}_k \sim \mathcal{N}(0, \frac{2}{m}\mathbf{I})$. We call these new matrices $\widetilde{W}$ and $\widetilde{A}$. For any fixed $x^\star$ in the support of $\mathcal{D}$ and normalized input sequence $x$ defined accordingly, we use $h_\ell, g_\ell, D_\ell, \mathsf{Back}_{i\to j}$ to denote those generated by $W, A$, and $\widetilde{h}_\ell, \widetilde{g}_\ell, \widetilde{D}_\ell, \widetilde{\mathsf{Back}}_{i\to j}$ to denote those generated by $\widetilde{W}, \widetilde{A}$. Recall from Lemma E.1 we have with probability at least

$1 - e^{-\Omega(\rho^2)}$ over the randomness of $W, A, \widetilde{W}, \widetilde{A}$,

$$\|g_\ell - \widetilde{g}_\ell\|, \|h_\ell - \widetilde{h}_\ell\| \leq O(\rho^5 \sqrt{N/m}) \quad \text{for every } \ell \in [L] \tag{G.14}$$

$$|\langle w_k, h_\ell - \widetilde{h}_\ell \rangle| \leq O(\rho^5 (N/m)^{2/3}) \quad \text{for every } k \in [m], \ell \in [L] \tag{G.15}$$

$$\left| \mathbf{e}_s^\top [\mathsf{Back}_{i' \to j'}]_k - \mathbf{e}_s^\top [\widetilde{\mathsf{Back}}_{i' \to j'}]_k \right| \leq O(\rho^7 (N/m)^{1/6}) \quad \forall k \in [m],\ s \in [d] \text{ and } 1 \leq i \leq j \leq L. \tag{G.16}$$

Let us choose $s = O(\rho^5 N^{2/3}/m^{1/6})$, and by Lemma B.1d, we have with probability $1 - e^{-\Omega(\rho^2)}$ over the randomness of $W, A$,

$$\left\{ k \in [m] \;\middle|\; |[g_{i'}]_k| \leq \frac{s}{\sqrt{m}} \right\} \leq O(sm + \rho^2) \ .$$

Using the randomness of $\mathcal{N}$, we have with probability $1 - e^{-\Omega(\rho^2)}$ over the randomness of $W, A, \mathcal{N}$,

$$\left\{ k \in \mathcal{N} \;\middle|\; |[g_{i'}]_k| \leq \frac{s}{\sqrt{m}} \right\} \leq O(sN + \rho^2) \ .$$

Let us define $\mathcal{N}_1$ be the above subset. Since $[g_{i'}]_k = \langle w_k, h_{i'-1} \rangle + \langle a_k, x_{i'} \rangle$ and since $|\langle w_k, h_{i'-1} \rangle - \langle w_k, \widetilde{h}_{i'-1} \rangle| \leq \frac{s}{\sqrt{m}}$ from (G.15), we have

$$k \in \mathcal{N} \setminus \mathcal{N}_1 \implies \mathbb{1}_{\langle w_k, h_{i'-1} \rangle + \langle a_k, x_{i'} \rangle \geq 0} = \mathbb{1}_{\langle w_k, \widetilde{h}_{i'-1} \rangle + \langle a_k, x_{i'} \rangle \geq 0}$$

For a similar reason, using $\|\widetilde{h}_{i-1}^{(0)}\| \in [0.5, 3]$ from Lemma B.1a and the independence between $w_k$ and $h_{i-1}^{(0)}$, we know with probability $1 - e^{-\Omega(\rho^2)}$ over the randomness of $W, A, \widetilde{W}, \widetilde{A}, \mathcal{N}$,

$$\left\{ k \in \mathcal{N} \;\middle|\; \left| |\langle w_k, \widetilde{h}_{i-1}^{(0)} \rangle| - \frac{\varepsilon_c}{\sqrt{m}} \right| \leq \frac{s}{\sqrt{m}} \right\} \leq O(sN + \rho^2) \ .$$

Let us define $\mathcal{N}_2$ be the above subset. Using $|\langle w_k, h_{i-1}^{(0)} - \widetilde{h}_{i-1}^{(0)} \rangle| \leq \frac{s}{\sqrt{m}}$ from (G.15) again, we have

$$k \in \mathcal{N} \setminus \mathcal{N}_2 \implies \mathbb{1}_{|\langle w_k, h_{i-1}^{(0)} \rangle| \leq \frac{\varepsilon_c}{\sqrt{m}}} = \mathbb{1}_{|\langle w_k, \widetilde{h}_{i-1}^{(0)} \rangle| \leq \frac{\varepsilon_c}{\sqrt{m}}}$$

Together, we have $|\mathcal{N}_1 \cup \mathcal{N}_2| \leq O(sN + \rho^2) = O(\rho^5 N^{5/3}/m^{1/6} + \rho^2)$. Let us choose

$$N \leq O(m^{0.1}/\rho^2)$$

so we have

$$|\mathcal{N}_1 \cup \mathcal{N}_2| \leq O(\rho^2)$$

Let us now fix $i, i', j, j', s, s', r$. Summing over $k \in \mathcal{N} \setminus \mathcal{N}_1 \cup \mathcal{N}_2$, we have with probability $1 - e^{-\Omega(\rho^2)}$,

$$\begin{aligned}
&\sum_{k \in \mathcal{N} \setminus \mathcal{N}_1 \cup \mathcal{N}_2} \left[ \mathbf{e}_{s'}^\top \mathsf{Back}_{i' \to j'} \right]_k \left[ \mathbf{e}_s^\top \mathsf{Back}_{i \to j}^{(0)} \right]_k \mathbb{1}_{|\langle w_k, h_{i-1}^{(0)} \rangle| \leq \frac{\varepsilon_c}{\sqrt{m}}} \mathbb{1}_{\langle w_k, h_{i'-1} \rangle + \langle a_k, x_{i'} \rangle \geq 0} H_{i \to j, r, s}(a_k) \langle h_{i'-1}, h_{i-1}^{(0)} \rangle \\
= &\sum_{k \in \mathcal{N} \setminus \mathcal{N}_1 \cup \mathcal{N}_2} \left[ \mathbf{e}_{s'}^\top \mathsf{Back}_{i' \to j'} \right]_k \left[ \mathbf{e}_s^\top \mathsf{Back}_{i \to j}^{(0)} \right]_k \mathbb{1}_{|\langle w_k, \widetilde{h}_{i-1}^{(0)} \rangle| \leq \frac{\varepsilon_c}{\sqrt{m}}} \mathbb{1}_{\langle w_k, \widetilde{h}_{i'-1} \rangle + \langle a_k, x_{i'} \rangle \geq 0} H_{i \to j, r, s}(a_k) \langle h_{i'-1}, h_{i-1}^{(0)} \rangle \\
\overset{①}{=} &\sum_{k \in \mathcal{N} \setminus \mathcal{N}_1 \cup \mathcal{N}_2} \left[ \mathbf{e}_{s'}^\top \widetilde{\mathsf{Back}}_{i' \to j'} \right]_k \left[ \mathbf{e}_s^\top \widetilde{\mathsf{Back}}_{i \to j}^{(0)} \right]_k \mathbb{1}_{|\langle w_k, \widetilde{h}_{i-1}^{(0)} \rangle| \leq \frac{\varepsilon_c}{\sqrt{m}}} \mathbb{1}_{\langle w_k, \widetilde{h}_{i'-1} \rangle + \langle a_k, x_{i'} \rangle \geq 0} H_{i \to j, r, s}(a_k) \langle \widetilde{h}_{i'-1}, \widetilde{h}_{i-1}^{(0)} \rangle \\
&\pm O(CN\rho^8 (N/m)^{1/6})
\end{aligned}$$

Above, ① uses $\left| \left[ \mathbf{e}_{s'}^\top \mathsf{Back}_{i' \to j'} \right]_k \right|, \left| \left[ \mathbf{e}_{s'}^\top \mathsf{Back}_{i' \to j'}^{(0)} \right]_k \right| \leq O(\rho)$ and $\|\widetilde{h}_{i'-1}\|_2, \|\widetilde{h}_{i'-1}^{(0)}\|_2 \leq 2$ from Lemma B.1, as well as (G.16) and (G.14), and the range $|H_{i \to j, r, s}| \leq C \overset{\text{def}}{=} \frac{4(C')^2}{\varepsilon_e \varepsilon_x}$ from Lemma 5.2. Using these

bounds, we can also sum up over $k \in \mathcal{N}_1 \cup \mathcal{N}_2$:

$$\sum_{k\in\mathcal{N}_1\cup\mathcal{N}_2} \left[\mathbf{e}_{s'}^\top \mathsf{Back}_{i'\to j'}\right]_k \left[\mathbf{e}_s^\top \mathsf{Back}_{i\to j}^{(0)}\right]_k \mathbb{1}_{|\langle w_k, h_{i-1}^{(0)}\rangle| \le \frac{\varepsilon_c}{\sqrt{m}}} \mathbb{1}_{\langle w_k, h_{i'-1}\rangle + \langle a_k, x_{i'}\rangle \ge 0} H_{i\to j,r,s}(a_k) \langle h_{i'-1}, h_{i-1}^{(0)}\rangle$$
$$= \pm O(C\rho^4)$$

Thus, putting together the two summations we have

$$\underbrace{\sum_{k\in\mathcal{N}} \left[\mathbf{e}_{s'}^\top \mathsf{Back}_{i'\to j'}\right]_k \left[\mathbf{e}_s^\top \mathsf{Back}_{i\to j}^{(0)}\right]_k \mathbb{1}_{|\langle w_k, h_{i-1}^{(0)}\rangle| \le \frac{\varepsilon_c}{\sqrt{m}}} \mathbb{1}_{\langle w_k, h_{i'-1}\rangle + \langle a_k, x_{i'}\rangle \ge 0} H_{i\to j,r,s}(a_k) \langle h_{i'-1}, h_{i-1}^{(0)}\rangle}_{G_{i,i',j,j',r,s,s'}}$$
$$= \underbrace{\sum_{k\in\mathcal{N}} \left[\mathbf{e}_{s'}^\top \widetilde{\mathsf{Back}}_{i'\to j'}\right]_k \left[\mathbf{e}_s^\top \widetilde{\mathsf{Back}}_{i\to j}^{(0)}\right]_k \mathbb{1}_{|\langle w_k, \widetilde{h}_{i-1}^{(0)}\rangle| \le \frac{\varepsilon_c}{\sqrt{m}}} \mathbb{1}_{\langle w_k, \widetilde{h}_{i'-1}\rangle + \langle a_k, x_{i'}\rangle \ge 0} H_{i\to j,r,s}(a_k) \langle \widetilde{h}_{i'-1}, \widetilde{h}_{i-1}^{(0)}\rangle}_{\widetilde{G}_{i,i',j,j',r,s,s'}}$$
$$\pm O(CN\rho^8 (N/m)^{1/6} + C\rho^4)$$

Now, we consider fixed $\widetilde{W}, \widetilde{A}, \mathcal{N}$ and only use the randomness of $\{w_k, a_k\}_{k\in\mathcal{N}}$ to analyze $\widetilde{G}_{i,i',j,j',r,s,s'}$. It is a summation of $N$ independent random variables. By Chernoff bound, with probability $1 - e^{-\Omega(\rho^2)}$,

$$|\widetilde{G}_{i,i',j,j',r,s,s'} - \mathbb{E}_{\{w_k,a_k\}_{k\in\mathcal{N}}}[\widetilde{G}_{i,i',j,j',r,s,s'}]| \le O(C\rho^2\sqrt{N})$$

By Lemma 5.2 we know that the expectation is given by:

$$\mathbb{E}[G_{i,i',j,j',r,s,s'}] = \Big( \underbrace{\sum_{k\in\mathcal{N}} \left[\mathbf{e}_{s'}^\top \widetilde{\mathsf{Back}}_{i'\to j'}\right]_k \left[\mathbf{e}_s^\top \widetilde{\mathsf{Back}}_{i\to j}^{(0)}\right]_k \langle \widetilde{h}_{i'-1}, \widetilde{h}_{i-1}^{(0)}\rangle}_{\widetilde{Z}_{i,i',j,j',s,s'}} \Big) \cdot \left(\Phi_{i\to j,r,s}(\langle w^*_{i\to j,r,s}, x_i^\star\rangle) \cdot \mathbb{1}_{i=i'} \pm \varepsilon_e\right)$$

Let us define

$$Z_{i,i',j,j',s,s'} \stackrel{\text{def}}{=} \sum_{k\in\mathcal{N}} \left[\mathbf{e}_{s'}^\top \mathsf{Back}_{i'\to j'}\right]_k \left[\mathbf{e}_s^\top \mathsf{Back}_{i\to j}^{(0)}\right]_k \langle h_{i'-1}, h_{i-1}^{(0)}\rangle$$

Using (G.14) and (G.16) again, we can bound

$$\left|\widetilde{Z}_{i,i',j,j',s,s'} - Z_{i,i',j,j',s,s'}\right| \le O\left(N\rho^8 (N/m)^{1/6}\right) \quad .$$

Therefore, so far we have calculated that with probability at least $1 - e^{-\Omega(\rho^2)}$ over the randomness of $W, A, \mathcal{N}$:

$$G_{i,i',j,j',r,s,s'} = Z_{i,i',j,j',s,s'} \cdot \Phi_{i\to j,r,s}(\langle w^*_{i\to j,r,s}, x_i^\star\rangle) \cdot \mathbb{1}_{i=i'} \pm O(\varepsilon_e \rho^2 N + CN\rho^8 (N/m)^{1/6} + C\rho^4 + C\rho^2\sqrt{N})$$

Summing it up, and using $C_{i\to j,s} \ge \Omega(\frac{1}{d})$ from Claim G.1b, we can write

$$f_{j',s'} = \sum_{i'=2}^{j'} \sum_{i=2}^{L-1} \sum_{j=i+1}^{L} \sum_{r\in[p],s\in[d]} \frac{1}{mC_{i\to j',s}} \frac{m}{N} \mathbb{E}_{\mathcal{N}}[G_{i,i',j,j',r,s,s'}]$$
$$= \sum_{i=2}^{j'} \sum_{j=i+1}^{L} \sum_{r\in[p],s\in[d]} \frac{1}{C_{i\to j',s}} \frac{1}{N} \mathbb{E}_{\mathcal{N}}[Z_{i,i,j,j',s,s'}] \cdot \Phi_{i\to j,r,s}(\langle w^*_{i\to j,r,s}, x_i^\star\rangle) \pm error$$
$$= \sum_{i=2}^{j'} \sum_{j=i+1}^{L} \sum_{r\in[p],s\in[d]} \frac{1}{mC_{i\to j',s}} \left\langle \mathbf{e}_{s'}^\top \mathsf{Back}_{i\to j'}, \mathbf{e}_s^\top \mathsf{Back}_{i\to j}^{(0)} \right\rangle \langle h_{i-1}, h_{i-1}^{(0)}\rangle \cdot \Phi_{i\to j,r,s}(\langle w^*_{i\to j,r,s}, x_i^\star\rangle) \pm error$$

with $error = \frac{p}{m} \cdot O(\varepsilon_e \rho^5 m + Cm\rho^{11}(N/m)^{1/6} + C\frac{m}{N}\rho^7 + C\rho^5\frac{m}{\sqrt{N}})$.

Next, using (1) $\left\|\mathbf{e}_{s'}^\top \mathsf{Back}_{i\to j'}^{(0)} - \mathbf{e}_{s'}^\top \mathsf{Back}_{i\to j'}\right\|_2 \leq O\big(\rho^5 \varepsilon_x^{1/3}\sqrt{m}\big)$ from Lemma D.1c, (2) $\|h_{i-1}^{(0)} - h_{i-1}\| \leq O(\sqrt{L}\varepsilon_x)$ from Lemma D.1a, (3) $\|\mathsf{Back}_{i\to j}\|_2 \leq O(\rho\sqrt{m})$ from Lemma B.1f, and (4) notation $C'' \stackrel{\text{def}}{=} \mathfrak{C}_{\mathfrak{s}}(\Phi, 1)$, we can write

$$f_{j',s'} = \sum_{i=2}^{j'} \sum_{j=i+1}^{L} \sum_{r\in[p], s\in[d]} \frac{1}{mC_{i\to j',s}} \left\langle \mathbf{e}_{s'}^\top \mathsf{Back}_{i\to j'}^{(0)}, \mathbf{e}_s^\top \mathsf{Back}_{i\to j}^{(0)} \right\rangle \langle h_{i-1}^{(0)}, h_{i-1}^{(0)}\rangle \cdot \Phi_{i\to j,r,s}(\langle w_{i\to j,r,s}^*, x_i^\star\rangle)$$
$$\pm \frac{p}{m} \cdot O(C''(\rho^5\varepsilon_x + \rho^8\varepsilon_x^{1/3})m + \varepsilon_e\rho^5 m + Cm\rho^{11}(N/m)^{1/6} + C\frac{m}{N}\rho^7 + C\rho^5\frac{m}{\sqrt{N}})$$

Using Lemma C.1, we further have

$$f_{j',s'} = \sum_{i=2}^{j'-1} \sum_{r\in[p], s\in[d]} \frac{1}{mC_{i\to j',s}} \left\langle \mathbf{e}_{s'}^\top \mathsf{Back}_{i\to j'}^{(0)}, \mathbf{e}_s^\top \mathsf{Back}_{i\to j'}^{(0)} \right\rangle \langle h_{i-1}^{(0)}, h_{i-1}^{(0)}\rangle \cdot \Phi_{i\to j',r,s}(\langle w_{i\to j',r,s}^*, x_i^\star\rangle)$$
$$\pm \frac{p}{m} \cdot O(C''\rho^6 m^{3/4} + C''(\rho^5\varepsilon_x + \rho^8\varepsilon_x^{1/3})m + \varepsilon_e\rho^5 m + Cm\rho^{11}(N/m)^{1/6} + C\frac{m}{N}\rho^7 + C\rho^5\frac{m}{\sqrt{N}})$$

When $s \neq s'$, using the randomness of $B$, it is easy to see with probability at least $1 - e^{-\Omega(\rho^2)}$

$$\left\langle \mathbf{e}_{s'}^\top \mathsf{Back}_{i\to j'}^{(0)}, \mathbf{e}_s^\top \mathsf{Back}_{i\to j'}^{(0)} \right\rangle = \pm O\left(\frac{\rho}{\sqrt{m}} \cdot \|\mathsf{Back}_{i\to j'}^{(0)}\|_2^2\right) = \pm O(\rho^8\sqrt{m})$$

where the last equality is due to Lemma B.1i and $\|B\|_2 \leq O(\rho\sqrt{m})$. Therefore, we can write

$$\begin{aligned} f_{j',s'} &= \sum_{i=2}^{j'-1} \sum_{r\in[p]} \frac{1}{mC_{j',i,s'}} \left\|\mathbf{e}_{s'}^\top \mathsf{Back}_{i\to j'}^{(0)}\right\|^2 \|h_{i-1}^{(0)}\|^2 \cdot \Phi_{i\to j',r,s}(\langle w_{i\to j',r,s}^*, x_i^\star\rangle) \pm error \\ &= \sum_{i=2}^{j'-1} \sum_{r\in[p]} \Phi_{i\to j',r,s}(\langle w_{i\to j',r,s}^*, x_i^\star\rangle) \pm error \end{aligned}$$

with

$$\begin{aligned} error &= \frac{p}{m} \cdot O\big(C''\rho^{10}\sqrt{m} + C''\rho^6 m^{3/4} + C''(\rho^5\varepsilon_x + \rho^8\varepsilon_x^{1/3})m \\ &\qquad + \varepsilon_e\rho^5 m + Cm\rho^{11}(N/m)^{1/6} + C\frac{m}{N}\rho^7 + C\rho^5\frac{m}{\sqrt{N}}\big) \\ &\leq \frac{p}{m} \cdot O(C''\rho^8\varepsilon_x^{1/3} m + \varepsilon_e\rho^5 m + Cm\rho^{11}(N/m)^{1/6} + C\frac{m}{N}\rho^7 + C\rho^5\frac{m}{\sqrt{N}}) \\ &\stackrel{①}{\leq} p\rho^{11} \cdot O(C''\varepsilon_x^{1/3} + \varepsilon_e + Cm^{-0.05}) \end{aligned}$$

Above, ① uses our choice $N = \Theta(m^{0.1}/\rho^2)$.

□

## G.4 Norm Bound on $W^*$

**Claim G.1.** *With probability at least $1 - e^{-\Omega(\rho^2)}$ over the random initialization $W, A, B$, we have*

*(a)* $\|W^*\|_{2,\infty} \leq O\big(\frac{p\rho^3 C}{m}\big)$ *so* $\|W^*\|_F \leq O\big(\frac{p\rho^3 C}{\sqrt{m}}\big)$.

*(b)* $\Omega(\frac{1}{d}) \leq C_{i\to j,s} \leq O(\frac{\rho^2}{d})$ *for every* $1 \leq i \leq j \leq L$ *and* $s \in [d]$.

*where the notion $C$ comes from Definition 5.3.*

*Proof of Claim G.1.* For the norm bound on $W^*$, note that for any unit vector $z$, we have

$$|\langle W_k^*, z\rangle| \leq \sum_{i,j} \sum_{r\in[p],s\in[d]} \frac{1}{mC_{i\to j',s}} \left|\left[\mathbf{e}_s^\top \mathsf{Back}_{i\to j}^{(0)}\right]_k\right| \cdot |H_{i\to j,r,s}(a_k)|\, |\langle h_{i-1}^{(0)}, z\rangle|$$

$$\leq \sum_{i,j} \sum_{r\in[p],s\in[d]} O(\frac{d}{m}) \cdot O(\frac{\rho}{\sqrt{d}}) \cdot C \cdot O(1) \leq O(\frac{p\rho^3 C}{m})$$

and as a result $\|W^*\|_F \leq O\big(\frac{p\rho^3 C}{\sqrt{m}}\big)$ . Lemma B.1a and Lemma B.1g together imply $C_{i\to j,s} \geq \Omega(\frac{1}{d})$. Lemma B.1a and Lemma B.1f together imply $C_{i\to j,s} \leq O(\frac{\rho^2}{d})$. □

# H Proof for Section 6

## H.1 Proof of Lemma 6.1

*Proof of Lemma 6.1.* This lemma is similar to the semi-smoothness theorem of [2]. Using the 1-Lipschitz continuity of ReLU, it is not hard to derive that there exists diagonal matrices $D''_\ell$ where $[D''_\ell]_{k,k} \in [-1,1]$ and is non zero *only when* $[D'_\ell]_{k,k} \neq 0$ satisfying

$$B(h_j + h'_j) - Bh_j = \sum_{i=1}^{j-1} B(D_j + D''_j)W \cdots W(D_{i+1} + D''_{i+1})W'(h_i + h'_i) \ . \tag{H.1}$$

(This is for instanced proved in [2, Claim G.2 of ver.3].) Note that $\|D''_\ell\|_0 \leq \|D'_\ell\|_0 \leq O(\rho^4 \Delta^{2/3} m^{2/3})$ by Lemma F.1b.

Now, for each term on its right hand side, we compare it to $BD_jW \cdots WD_{i+1}W'(h_i + h'_i)$. We can expand the difference into at most $2^L$ terms, each of the form (ignoring subscripts)

$$(BDW \cdots DW)D''(WD \cdots DW)D'' \cdots D''(WD \cdots WD)W'h$$

We have (note we can write $D'' = D'D''D'$ because each diagonal entry of $D'_\ell$ is in $\{-1, 0, 1\}$.

- $\|BDW \cdots DWD'\|_2 \leq O(\rho\sqrt{\|D'\|_0}) \leq O(\rho^3 \Delta^{1/3} m^{1/3})$ by Lemma B.1f
- $\|D'WD \cdots DWD'\|_2 \leq O(\rho\frac{\sqrt{\|D'\|_0}}{\sqrt{m}}) \leq \frac{1}{100L}$ by Lemma B.1j.
- $\|(WD \cdots WD)W'(h + h')\| \leq \|WD \cdots WD\|_2 \|W'\|_2 \|h + h'\| \leq O\big(\frac{L^3\Delta}{\sqrt{m}}\big)$ by Lemma B.1i (and $\|h + h'\| \leq O(1)$ using Lemma B.1a with Lemma F.1a).

Together, we have

$$\begin{aligned}
&\big\|B(D_j + D''_j)W \cdots W(D_{i+1} + D''_{i+1})W'(h_i + h'_i) - BD_jW \cdots WD_{i+1}W'(h_i + h'_i)\big\| \\
&\leq L \cdot O(\rho^3\Delta^{1/3}m^{1/3}) \cdot O(\frac{L^3\Delta}{\sqrt{m}}) + \binom{L}{2} \cdot \frac{1}{100L} \cdot O(\rho^3\Delta^{1/3}m^{1/3}) \cdot O(\frac{L^3\Delta}{\sqrt{m}}) \\
&\quad + \binom{L}{3} \cdot \frac{1}{(100L)^2} \cdot O(\rho^3\Delta^{1/3}m^{1/3}) \cdot O(\frac{L^3\Delta}{\sqrt{m}}) + \cdots \\
&\leq O(\frac{\rho^7\Delta^{4/3}}{m^{1/6}}) \ .
\end{aligned} \tag{H.2}$$

Finally, we also have

$$\big\|BD_jW \cdots WD_{i+1}W'h'_i\big\| \leq O(\rho\sqrt{m}) \cdot \|W'\|_2 \cdot \|h'_i\| \leq O\big(\frac{\rho^7\Delta^2}{\sqrt{m}}\big) \ . \tag{H.3}$$

Putting (H.2) and (H.3) back to (H.1) finishes the proof. □

### H.2 Proof of Lemma 6.2

*Proof of Lemma 6.2.* We write

$$\sum_{i'\in[j']}\big(\mathsf{Back}_{i'\to j'}+\mathsf{Back}'_{i'\to j'}\big)(D_{i'}+D'_{i'})\widetilde{W}(h_{i'-1}+h'_{i'-1})$$
$$=\sum_{i'\in[j']}\underbrace{\mathsf{Back}'_{i'\to j'}(D_{i'}+D'_{i'})\widetilde{W}(h_{i'-1}+h'_{i'-1})}_{\clubsuit_{i'}}+\sum_{i'\in[j']}\underbrace{\mathsf{Back}_{i'\to j'}(D_{i'}+D'_{i'})\widetilde{W}h'_{i'-1}}_{\spadesuit_{i'}}$$
$$+\sum_{i'\in[j']}\underbrace{\mathsf{Back}_{i'\to j'}D'_{i'}\widetilde{W}h_{i'-1}}_{\diamondsuit_{i'}}+\sum_{i'\in[j']}\mathsf{Back}_{i'\to j'}D_{i'}\widetilde{W}h_{i'-1}$$

We analyze the three error terms separately.

- $\|\clubsuit_{i'}\|\leq\|\mathsf{Back}'_{i'\to j'}\|_2\cdot\|\widetilde{W}\|_2\cdot O(1)\leq O\big(\frac{\omega\rho^6\Delta^{1/3}}{m^{1/6}}\big)$ using Lemma F.1c.
- $\|\spadesuit_{i'}\|\leq\|\mathsf{Back}_{i'\to j'}\|_2\cdot\|\widetilde{W}\|_2\cdot\|h'_{i'}\|\leq O\big(\frac{\omega\rho^7\Delta}{\sqrt{m}}\big)$ using Lemma B.1f and Lemma F.1a.
- $\|\diamondsuit_{i'}\|\leq\|\mathsf{Back}_{i'\to j'}D'_{i'}\|_2\cdot\|\widetilde{W}\|_2\cdot O(1)\leq O\big(\frac{\omega\rho^2\Delta^{1/3}}{m^{1/6}}\big)$ using Lemma B.1f and the bound $\|D'_{i'}\|_0$ from Lemma F.1b. □

# I Proof for Lemma 7.1

*Proof of Lemma 7.1.* To begin with, recall

$$\mathbb{E}_{(x^\star,y^\star)\sim\mathcal{D}}\left[\sum_{j=3}^{L}G(F^*_j(x^\star),y^\star_j)\right]\leq\mathsf{OPT}\ .$$

Since it satisfies $\|F^*_j(x^\star)\|\leq O(\sqrt{pLd}\,\mathfrak{C}_\mathfrak{s}(\Phi,1))$ for all $x^\star$, by the 1-Lipschitz continuity of $G(\cdot,y^\star)$, we also have $|G(F^*_j(x^\star),y^\star_j)|\leq O(\sqrt{pLd}\,\mathfrak{C}_\mathfrak{s}(\Phi,1))$. Therefore, by Chernoff bound, as long as $N\geq\Omega(\frac{\rho^3\cdot p\cdot\mathfrak{C}^2_\mathfrak{s}(\Phi,1)}{\varepsilon^2})$ we also have

$$\mathbb{E}_{(x^\star,y^\star)\sim\mathcal{Z}}\left[\sum_{j=3}^{L}G(F^*_j(x^\star),y^\star_j)\right]\leq\mathsf{OPT}+\tfrac{\varepsilon}{200}\ .$$

Let us recall the training objective

$$\mathsf{Obj}(W_t)\stackrel{\text{def}}{=}\mathop{\mathbb{E}}_{(x^\star,y^\star)\sim\mathcal{Z}}\big[\mathsf{Obj}(x^\star,y^\star;W_t)\big]$$
$$\text{where}\quad\mathsf{Obj}(x^\star,y^\star;W_t)\stackrel{\text{def}}{=}\sum_{j=3}^{L}G\left(\lambda F_j(x^\star;W+W_t),y^\star_j\right)$$

Let $x$ be a normalized input sequence generated by some $x^\star$ in the support of $\mathcal{D}$. Consider an iteration $t$ where the current weight matrix is $W+W_t$. Let

- $h_\ell,g_\ell,D_\ell,\mathsf{Back}_{i\to j}$ be defined with respect to $W,A,B,x$, and
- $h_\ell+h'_{\ell,t},g_\ell+g'_{\ell,t},D_\ell+D'_{\ell,t},\mathsf{Back}_{i\to j}+\mathsf{Back}'_{i\to j,t}$ be defined with respect to $W+W_t,A,B,x$

Let us define function $R_j(x^\star;\widetilde{W})\in\mathbb{R}^d$ as

$$R_j(x^\star;\widetilde{W})\stackrel{\text{def}}{=}\sum_{i=2}^{j}\big(\mathsf{Back}_{i\to j}+\mathsf{Back}'_{i\to j,t}\big)(D_i+D'_{i,t})\widetilde{W}(h_{i-1}+h'_{i-1,t})$$

which is a *linear* function over $\widetilde{W}$. Let us define a loss function $\widetilde{G}$ as:

$$\widetilde{G}(\widetilde{W}) \stackrel{\text{def}}{=} \mathbb{E}_{(x^\star,y^\star)\sim\mathcal{Z}}\left[\widetilde{G}(x^\star,y^\star;\widetilde{W})\right]$$

$$\text{where}\quad \widetilde{G}(x^\star,y^\star;\widetilde{W}) \stackrel{\text{def}}{=} \sum_{j=3}^{L} G\left(\lambda F_j(x^\star;W+W_t)+\lambda R_j(x^\star;\widetilde{W}),y_j^\star\right)$$

It is not hard to verify that

$$\mathsf{Obj}(W_t)=\widetilde{G}(0) \quad\text{and}\quad \nabla\mathsf{Obj}(W_t)=\nabla\widetilde{G}(0) \ .$$

Let $W^*$ be defined in Definition 5.3. By Lemma 6.3, we know that as long as $\|W_t\|_2 \le \frac{\Delta}{\sqrt{m}}$ (for some parameter $\Delta\in[\varrho^{-100},\varrho^{100}]$ we shall choose at the end),

$$\left\|R_j(x^\star;W^*)-F_j^*(x^\star)\right\| \le p\rho^{12}\cdot O(\varepsilon_e+\mathfrak{C}_{\mathfrak{s}}(\Phi,1)\varepsilon_x^{1/3}+Cm^{-0.05}) \le \frac{\varepsilon}{100L}$$

where the last inequality is by choosing $\varepsilon_e=\Theta(\frac{\varepsilon}{p\rho^{13}})$ and $\varepsilon_x=\frac{1}{\mathsf{poly}(\rho,p,\varepsilon^{-1},C')}$ and sufficiently large $m$. Taking union bound over all samples in $\mathcal{Z}$, by the 1-Lipschitz continuity of $G$, we have

$$\mathbb{E}_{(x^\star,y^\star)\sim\mathcal{Z}}\left[\sum_{j=3}^{L}G(R_j(x^\star,W^*),y_j^\star)\right] \le \mathsf{OPT}+\frac{\varepsilon}{50} \ .$$

By Lemma 6.1 and Lemma 6.2 we know that

$$\begin{aligned}
F_j(x^\star,W+W_t)-F_j(x^\star,W) &= \sum_{i=1}^{j-1}\mathsf{Back}_{i\to j}\,D_{i+1}W_t h_i \pm O\big(\frac{\rho^7\Delta^{4/3}}{m^{1/6}}\big)\\
&= \sum_{i=1}^{j-1}\big(\mathsf{Back}_{i\to j}+\mathsf{Back}'_{i\to j,t}\big)\big(D_i+D'_{i,t}\big)W_t(h_{i-1}+h'_{i-1,t}) \pm O\big(\frac{\rho^7\Delta^{4/3}}{m^{1/6}}\big)\\
&= R_j(x^\star,W_t)\pm O\big(\frac{\rho^7\Delta^{4/3}}{m^{1/6}}\big) \ .
\end{aligned}$$

Using 1-Lipschitz continuity of $G$, we have

$$\begin{aligned}
\widetilde{G}\left(\frac{1}{\lambda}W^*-W_t\right) &= \mathbb{E}_{(x^\star,y^\star)\sim\mathcal{Z}}\left[\sum_{j=3}^{L}G\left(\lambda F_j(x^\star,W+W_t)+\lambda R_j\left(x^\star,\frac{1}{\lambda}W^*-W_t\right),y_j^\star\right)\right]\\
&= \mathbb{E}_{(x^\star,y^\star)\sim\mathcal{Z}}\left[\sum_{j=3}^{L}G\left(\lambda F_j(x^\star,W)+R_j(x^\star,W^*),y_j^\star\right)\right]\pm O\big(\frac{\rho^7\Delta^{4/3}}{m^{1/6}}\big)\\
&\overset{①}{=} \mathbb{E}_{(x^\star,y^\star)\sim\mathcal{Z}}\left[\sum_{j=3}^{L}G\left(R_j(x^\star,W^*),y_j^\star\right)\right]\pm\frac{\varepsilon}{10}\pm O\big(\frac{\rho^7\Delta^{4/3}}{m^{1/6}}\big)
\end{aligned}$$

where ① is by our choice of $\lambda$ which implies $\lambda\|F_j(x^\star;W)\| \le \lambda\cdot O(\rho)\le\frac{\varepsilon}{10L}$ by Lemma B.1h.

Together, we have

$$\widetilde{G}\left(\frac{1}{\lambda}W^*-W_t\right)\le\mathsf{OPT}+\frac{\varepsilon}{5}$$

Thus, by the convexity of $\widetilde{G}(\widetilde{W})$ (composing convex function with linear function is convex), we

know

$$\left\langle \nabla \widetilde{G}(0), \frac{1}{\lambda} W^* - W_t \right\rangle \leq \widetilde{G}\big(\frac{1}{\lambda} W^* - W_t\big) - \widetilde{G}(0) \ .$$

Finally, recall that SGD takes a stochastic gradient so

$$W_{t+1} = W_t - \eta \widetilde{\nabla}_t \quad \text{where} \quad \widetilde{\nabla}_t \stackrel{\text{def}}{=} \nabla \mathsf{Obj}\left((x^\star)^{(t)}, (y^\star)^{(t)}; W_t\right) = \nabla \widetilde{G}\left((x^\star)^{(t)}, (y^\star)^{(t)}; 0\right)$$

and we have

$$\left\|\lambda^{-1} W^* - W_{t+1}\right\|_F^2 = \left\|\lambda^{-1} W^* - W_t\right\|_F^2 + 2\eta \langle \lambda^{-1} W^* - W_t, \widetilde{\nabla}_t \rangle + \eta^2 \left\|\widetilde{\nabla}_t\right\|_F^2 \ .$$

On one hand, we have (for any $(x^\star, y^\star) \sim \mathcal{D}$)

$$\left\|\nabla \widetilde{G}\left(x^\star, y^\star; 0\right)\right\|_F \stackrel{①}{\leq} \sum_{j=3}^{L} \lambda \left\|\nabla R_j(x^\star; 0)\right\|_F \stackrel{②}{\leq} L^2 \lambda O(\rho\sqrt{m}) \leq O(\varepsilon\rho\sqrt{m}) \ .$$

Above, ① uses the 1-Lipschitz continuity of $G$, and ② uses Lemma B.1 and the choice of $\lambda \leq \frac{\varepsilon}{L\rho}$.

On the other hand, we have $\mathbb{E}_{sgd}[\widetilde{\nabla}_t] = \nabla \widetilde{G}(0)$. Together, we know

$$\begin{aligned}
\mathbb{E}_{sgd}\left[ \left\|\lambda^{-1} W^* - W_{t+1}\right\|_F^2 \right] &\leq \left\|\lambda^{-1} W^* - W_t\right\|_F^2 + 2\eta\left(\mathsf{OPT} + \frac{\varepsilon}{5} - \widetilde{G}(0)\right) + O\left(\varepsilon^2\eta^2\rho^2 m\right) \\
&= \left\|\lambda^{-1} W^* - W_t\right\|_F^2 + 2\eta\left(\mathsf{OPT} + \frac{\varepsilon}{5} - \mathsf{Obj}(W_t)\right) + O\left(\varepsilon^2\eta^2\rho^2 m\right)
\end{aligned}$$

Telescoping over all $t = 0, 1, \ldots, T-1$, we have

$$\begin{aligned}
\frac{1}{T} \mathbb{E}_{sgd}[\mathsf{Obj}(W_t)] &\leq \mathsf{OPT} + \frac{\varepsilon}{5} + \eta \cdot O\left(\varepsilon^2\rho^2 m\right) + \frac{\|\lambda^{-1} W^*\|_F^2}{\eta T} \\
&\stackrel{①}{\leq} \mathsf{OPT} + \frac{\varepsilon}{4} + \frac{\rho^6 m \|W^*\|_F^2}{\varepsilon T} \\
&\stackrel{②}{\leq} \mathsf{OPT} + \frac{\varepsilon}{4} + \frac{p^2\rho^{12} C^2}{\varepsilon T}
\end{aligned}$$

Above, ① uses our choice $\eta = \Theta\big(\frac{1}{\varepsilon\rho^2 m}\big)$ and $\lambda$; and ② uses the bound on $\|W^*\|_F$ from Claim G.1a. It thus suffices to choose $T = \Theta\big(\frac{p^2\rho^{12}C^2}{\varepsilon^2}\big)$ to satisfy $\frac{1}{T}\mathbb{E}\left[\mathsf{Obj}(W_t)\right] \leq \mathsf{OPT} + \frac{\varepsilon}{2}$.

As for the norm bound, we have

$$\|W_t\|_F \leq \eta \sum_{t=0}^{T-1} \|\widetilde{\nabla}_t\|_F \leq \eta T \cdot O(\varepsilon\rho\sqrt{m}) = O\big(\frac{C^2\rho^{11}p^2}{\varepsilon^2} \cdot \frac{1}{\sqrt{m}}\big) \tag{I.1}$$

so we can choose $\Delta = \Theta\big(\frac{C^2\rho^{11}p^2}{\varepsilon^2}\big)$.

Finally, we can replace the notation $\mathfrak{C}_{\varepsilon_e}$ with $\mathfrak{C}_\varepsilon$ because $\log(1/\varepsilon_e) = O(\log(1/\varepsilon))$. □